%% file: main.tex
\documentclass{ucetd}

\usepackage[T1]{fontenc}
\usepackage{subcaption,graphicx}
\usepackage{natbib}
\usepackage{mathtools}  
\usepackage{amssymb}    
\usepackage{amsthm}

\newcommand{\thesistitle}{Towards Interpreting Language Models: A Case Study in Multi-Hop Reasoning}
\newcommand{\thesisauthor}{Mansi Sakarvadia}
\department{Computer Science}
\division{Physical Sciences}
\degree{Master of Science}
\date{April 5, 2024}

\usepackage{csquotes}
\usepackage{amsmath, amssymb, amsfonts}
\usepackage{booktabs}
\usepackage{comment}
\usepackage{graphicx}
\usepackage{multirow}
\usepackage{subcaption}
\usepackage{enumerate}
\usepackage{xfrac}
\usepackage[dvipsnames]{xcolor}
\usepackage[linesnumbered,ruled,vlined]{algorithm2e}

\SetCommentSty{mycommfont}
\usepackage{float}
\usepackage{color}
\usepackage{pifont}

\title{\thesistitle}
\author{\thesisauthor}

\dedication{
I would like to thank my advisors Dr. Ian Foster and Dr. Kyle Chard for their support, encouragement, and guidance in writing this research. They gave me the freedom to explore and pursue creative ideas, and for that I am very grateful.

I would like to acknowledge my amazing collaborators: Aswathy Ajith, Arham Khan, Daniel Grzenda, Dr. Nathaniel Hudson, and Dr. Andr\'e Bauer. I learn so much from each of you.

My sincere thanks to all of my fellow lab mates at Globus Labs. They have made my graduate journey a lot of fun.

Most importantly, I would like to thank my family for their unwavering support and encouragement during my graduate studies and my lovely friends who cheer me on and celebrate my accomplishments every step of the way.
}

\epigraph{

\enquote{Nothing in life is to be feared, it is only to be understood.} 

Dr. Marie Curie

}

\usepackage{doi}
\usepackage{xurl}
\hypersetup{bookmarksnumbered,
            unicode,
            linktoc=all,
            pdftitle={\thesistitle},
            pdfauthor={\thesisauthor},
            pdfsubject={},                                
            pdfborder={0 0 0}}
\makeatletter
\let\ORG@hyper@linkstart\hyper@linkstart
\protected\def\hyper@linkstart#1#2{%
  \lowercase{\ORG@hyper@linkstart{#1}{#2}}}
\makeatother

\input{meta/custom_commands}

\input{meta/notation}

\begin{document}
\maketitle

\makecopyright
\makededication
\makeepigraph

\tableofcontents
\listoffigures 
\listoftables

\acknowledgments
\label{sec:ack}
\input{_acknowledgements}

\abstract
Answering multi-hop reasoning questions requires retrieving and synthesizing information from diverse sources.
Language models (LMs) struggle to perform such reasoning consistently. We propose an approach to pinpoint and rectify multi-hop reasoning failures through targeted \emph{memory injections} on LM attention heads. First, we analyze the per-layer activations of GPT-2 models in response to single- and multi-hop prompts. We then propose a mechanism that allows users to inject relevant prompt-specific information, which we refer to as \enquote{memories,} at critical LM locations during inference. By thus enabling the LM to incorporate additional relevant information during inference, we enhance the quality of multi-hop prompt completions. We empirically show that a simple, efficient, and targeted memory injection into a key attention layer often increases the probability of the desired next token in multi-hop tasks, by up to 424\%. 
We observe that small subsets of attention heads can significantly impact the model prediction during multi-hop reasoning. To more faithfully interpret these heads, 
we develop \texttt{Attention Lens}: an open source tool that translates the outputs of attention heads into vocabulary tokens via learned transformations called \emph{lenses}. We demonstrate the use of lenses to reveal how a model arrives at its answer and use them to localize sources of model failures such as in the case of biased and malicious language generation. 


\mainmatter

\chapter{Introduction}
\input{introduction/introduction}

\chapter{Related Work}
\input{related_works/related_works}

\chapter{Memory Injections}

\section{Introduction}
\label{sec:intro}
\input{chapter1/1_introduction}

\section{Background \& Notation}
\label{sec:background}
\input{chapter1/2_background}

\section{Experimental Overview}
\label{sec:experiments}
\input{chapter1/3_experiments}

\section{Proposed Methods}
\label{sec:algorithm}
\input{chapter1/4_algorithm}

\section{Results and Discussion}
\label{sec:results}
\input{chapter1/5_results}

\section{Additional Memory Encoding Styles}
\label{sec:more_experiments}
\input{chapter1/more_experiments}

\section{Related Work}
\label{sec:related_work}
\input{chapter1/related_work}

\section{Conclusions and Future Directions}
\label{sec:conclusions}
\input{chapter1/conclusions}

\section{Broader Impacts \& Ethics}
\label{sec:ethics}
\input{chapter1/ethics}


\chapter{Attention Lens}

\section{Introduction}
\label{AL:intro}
\input{AttentionLens/sections/1_intro}

\label{AL:background}
\input{AttentionLens/sections/2_background}

\section{Training Lenses}
\label{AL:design}
\input{AttentionLens/sections/3_training}

\section{Attention Lens Applications}
\label{AL:use}
\input{AttentionLens/sections/4_use_cases}

\section{Evaluating Lenses}
\label{AL:eval}
\input{AttentionLens/sections/5_evaluation}

\section{Conclusion}
\label{AL:conclusion}
\input{AttentionLens/sections/6_conclusion}

\section*{Limitations}
\input{AttentionLens/sections/_limitations}

\chapter{Summary \& Future Works}
\input{Summary/summary}

\makebibliography
\nocite{*}

\end{document}

%% file: meta/custom_commands.tex
\newcommand{\mansi}[1]{{\color{ForestGreen} \textbf{Mansi:} \enquote{#1}}}

\newcommand{\gptsmall}{GPT2-Small}
\newcommand{\gptlarge}{GPT2-Large}
\newcommand{\gptxl}{GPT2-XL}
\newcommand{\gptj}{GPT-J}
\newcommand{\gptneosmall}{GPT-Neo (125M)}
\newcommand{\gptneolarge}{GPT-Neo (1.3B)}
\newcommand{\gptneoxl}{GPT-Neo (2.7B)}

\newcommand{\easydata}{\emph{Hand}}
\newcommand{\harddata}{\emph{2WMH}}

\newcommand{\embed}{\emph{Embed}}
\newcommand{\unembed}{\emph{Unembed}}
\newcommand{\layerwise}{\emph{Layer-wise}}

\newlength{\encerr}

\newcommand{\encodingerror}{%
  \hbox{%
    $\blacklozenge$%
    \settowidth{\encerr}{$\blacklozenge$}%
    \hspace{-0.77\encerr}%
    {\color{white}\textsmaller[2]{?}}%
   }%
  \xspace%
 }

%% file: meta/notation.tex
\DeclareMathOperator*{\softmax}{softmax}

\DeclareMathOperator*{\argmin}{arg\,min}

\newcommand{\cmark}{\ding{51}}
\newcommand{\xmark}{\ding{55}}

%% file: _acknowledgements.tex
This material is based upon work supported by the U.S. Department of
Energy, Office of Science, Office of Advanced Scientific Computing Research, Department of
Energy Computational Science Graduate Fellowship under Award Number DE-SC0023112. 
This work is also supported in part by the U.S.\ Department of Energy under Contract DE-AC02-06CH11357.

%% file: introduction/introduction.tex
Despite recent widespread adoption of neural Language Models (LMs) \citep{vaswani2017attention, brown_LanguageModelsAre_2020} in chat-based applications \citep{openai2022chatgpt}, the mechanisms by which LMs acquire knowledge during training and recall knowledge to form predictions at inference time are not well understood. This complicates the safe deployment of LMs in consumer and scientific pipelines \citep[inter alia]{gaudin2023algorithms,yun2023fine, hardalov2018towards, jablonka202314} as LM behavior can be unpredictable. For example,  LMs are capable of exhibiting harmful behaviors including displaying bias, regurgitating private information, hallucinating, producing offensive language, and producing malicious outputs due to adversarial training \citep{nadeem2020stereoset, winograd2023loose, zhang2023siren, bender2021dangers, kandpal2023backdoor}. Mitigating these harmful behaviors is limited by our lack of understanding of how LMs work. To ensure success and safety of LM-based applications, better interpretability techniques must be developed to understand how models develop behaviors. In this work, we focus on developing better interpretability techniques to understand how models recall knowledge during inference.

We study the case of LMs attempting to perform multi-hop reasoning. Multi-hop reasoning is the task of answering a prompt that contains references to an entity that is never explicitly named (see Fig.~\ref{subfig:multi_hop_demo}). Many modern LMs struggle to consistently perform multi-hop reasoning \citep{arkoudas2023gpt,guo2023indeed,blair2023can}. We develop a method to localize multi-hop reasoning failures to specific attention heads within a model, inspect what terms an attention heads is outputting via a tool called \texttt{Attention Lens}, and an efficiently enhance multi-hop reasoning abilities during inference via our technique called \enquote{memory injections}. Our interpretability-driven techniques can be easily adapted to localize additional sets of LM behavior within model weights, are computationally efficient, and overcome the limitations of existing model behavior corrective techniques.

Popular techniques to correct model behavior cannot guarantee improved performance on a target task without negatively affecting the model's performance on unrelated tasks. This is because researchers do not have the tools to reliably pinpoint the source of model failure within the weight space, so they attempt to apply general and broad corrective techniques to entire model architectures. Examples of popular corrective techniques include fine-tuning \citep{dodge2020fine}, parameter-efficient fine-tuning \citep{peft}, human-feedback reinforcement learning \citep{christiano2017deep}, retraining \citep{wu2020deltagrad}, model editing \citep{wang2023knowledge, zhang2024comprehensive}, and unlearning \citep{bourtoule2021machine}. Many of these techniques can have unintended consequences. For example, fine-tuning can distort features learned during pre-training, induce catastrophic forgetting, and compromise model safety \citep{kumar2022finetuning, kirkpatrick2017overcoming, kemker2018measuring, qi2023finetuning}. 
Model editing can have unintended effects on other knowledge originally embedded in the model’s weights \citep{cohen2023evaluating}. 
Machine unlearning often does not provide guarantees about what a model knows; \cite{shi2024detecting, patil2023sensitive} showed that sensitive information could still be recovered from an edited model, even if unlearning strategies were applied to the model to remove the sensitive information. Through our work, we argue that the effectiveness of these corrective techniques could be enhanced via a more robust understanding of \textit{how} knowledge is embedded in models' weights. For example, a more faithful interpretation of a model's weight space could enable the application of model corrective techniques to subsets of model weights rather than entire model architectures which may alleviate some of the unintended consequences of current techniques.

Corrective techniques are also out of reach for many individuals and organizations due to high computational costs. Model sizes are rapidly increasing \citep{hestness2017deep, hoffmann2022training} which means that applying any gradient-based procedure (e.g., training, fine-tuning) to models at billion parameter scales with internet-scale datasets requires vast amounts of computational resources. For example, the 70 billion parameter LLama2 model was trained at Meta's Research Super Cluster \citep{Lee_Sengupta} and their internal production clusters; Llama2 70B's pre-training took $1,720,320$ GPU hours on Nvidia A100-80GB GPUs \citep{touvron2023llama}. Few organization have these types of resources to pre-train models. After training LLama2 70B, \cite{touvron2023llama} reports that \enquote{Llama 2-Chat is the result of several months of research and iterative applications of alignment techniques, including both instruction tuning and RLHF, requiring significant computational and annotation resources.} Ultimately, this report alludes to the reality that assessing and correcting model behavior at scale (with current tools) can be prohibitively expensive for individuals and smaller organizations. Better interpretability techniques could alleviate some of the computational cost in the model assessment and alignment workflows. For example, techniques that allow ML practitioners to quickly diagnose sources of model failure in weight space and apply targeted remedies to subsets of weights have lower computational resource requirements compared to corrective techniques that must be applied iteratively to full model architectures.


To summarize, better LM interpretability techniques could enable strides in many open problems:
\begin{enumerate}
    \item Limited effectiveness of current model behavior corrective techniques.
    \item High computational cost of evaluating and correcting models.
\end{enumerate}

To address these problems, we need to develop interpretability tools that allow us to further localize sources of model behavior within model architectures. There is evidence that much model behavior is localizable \citep{frankle2018lottery,  goldowsky2023localizing, wardat2021deeplocalize, maini2023can}. 
Behavior localization will enable the application of corrective techniques to a model in a more exact and targeted manner, thus alleviating both high computational costs and harmful side effects associated with current corrective techniques such as full model fine-tuning. Ultimately, better localization techniques will greatly enhance model transparency and boost understanding of how and why models succeed and fail in various scenarios.

A central challenge to our ability to localize model behavior is a lack of understanding as to how a model arrives at its final prediction: \textit{How are humans supposed to interpret the computations in intermediate model layers?} 
Additionally, we have a limited arsenal of model corrective techniques and could benefit from more tools in our tool box; especially gradient-free methods as these would be more accessible due to lower computational cost.
    
In this work, we address these shortcomings by making the following contributions:
    \begin{enumerate}
        \item We develop a gradient-free method, \enquote{memory injections}, to enhance model behavior at inference time in a human understandable format via intervening on hidden activations.
        \item We show how memory injections can be used to localize model failure on multi-hop reasoning tasks, and even correct model performance without modifying weights, thus eliminating concerns of hurting model behavior in unrelated tasks.
        \item We develop a method to interpret the outputs of attention heads in human-understandable formats.
        \item We develop a software framework, \texttt{Attention Lens}, to support training of probes into individual attention heads to better characterize their role during inference.
    \end{enumerate}

%% file: related_works/related_works.tex
We review interpretability tools such as probing, activation engineering, model editing, circuit discovery, and knowledge extraction. We also review recent advances in the study of language model reasoning capabilities and retrieval augmented generation.

\section{Probing Models}
\label{RW:probing}
\textbf{Probing} is a class of interpretability methods that attempt the decode the contents/functions encoded by neural network weights. Probing does this by directly mapping subsets of weight activations into human-understandable domains. Since activations are directly related to model inputs, probes allow researchers to causally draw connections between model inputs and probe outputs. Therefore, researchers may be able to use probes to localize sources of model behavior to a specific subset of model weights.
    
Probes can be designed flexibly to suite the task at hand and are typically optimized using gradient based techniques like stochastic gradient descent \citep{ruder2016overview}. There are two main axes of freedom in probe design: 
\begin{itemize}
        \item \textbf{Architecture: } Probe architectures are informed by the types of insight a researcher requires from a probe. For example, probes can be designed at varying level's of model architecture (e.g. weight-level, layer-level, attention head-level). Additionally, probes can be linear or non-linear. Certain model behaviors may be linearly decodable \citep{alain2016understanding} while others may need non-linear probes to decode \citep{white2021non}. Some works have even found that the manner in which a decoding task is defined can allow a probe to transition from non-linear \citep{li2022emergent} to linear \citep{nanda2023emergent}.
        \item \textbf{Training dataset: } since probes are trained to perform a mapping between a model's activation and a desired domain, the training data a probe sees will govern its behavior. This training data must elicit all of the behaviors the researcher is attempting to study.
\end{itemize}

There are many use cases for trained model probes. \cite{ettinger-etal-2016-probing} introduces using classifier probes to understand semantic information in sentence representations. Probes can be trained to decode a hidden model representation into vocabulary, often with the goal of attempting to understand how each model layer informs how the model arrives at a final token prediction \citep{logitlens, tuned_lens, pal2023future, katz2024backward}. \cite{li2022emergent} trained probes to understand a model's internal board representation when predicting the next best move in the board game Othello. \cite{kim2019probing} explored the effect of pre-training on the model's learned representations of \enquote{function words} via trained probes. \cite{aina2021language} used probes to quantify model uncertainty in its completions of ambiguous prompts.

A limitation of probes as a diagnostic tool is that it is not obvious if the probes are are correlational or causal tools for attempting to understand a model's internal representations \citep{belinkov2022probing}. To combat this limitation, researchers can further validate the efficacy of probes by using the probes to guide activation engineering and observe if downstream model performance is affected as they would expect. For example, \cite{li2022emergent} shows that by using probes to guide the editing of Othello-GPT's representation of the board state, they could alter the model's final next move prediction as expected; this further validated that the probes were faithfully decoding information as it was known to the model. See section~\ref{RW:activation_engineering} for additional techniques about how to engineer activations.

\section{Activation Engineering}
\label{RW:activation_engineering}

\textbf{Activation Engineering} is a class of interpretability method that allows researchers to decode the functions of model components, by modifying their respective output activation values and observing the downstream effect. Like probing~\ref{RW:probing}, the goal of activation engineering can be to allow researchers to attribute model behavior back to specific model components. Additionally, activation engineering can also be used to directly influence model behavior at inference time. 

Some examples of applications of activation engineering are:
\begin{itemize}
    \item \cite{vig2020investigating} introduced \enquote{causal mediation analysis}: a method to understand which components are a model are responsible for specific behaviors in language modeling; they apply causal mediation analysis to investigate which components of a model are responsible for gender bias.
    \item  \cite{sun2021react} demonstrates that activations of neural networks can be used to identify in-distribution and out-of-distribution model inputs in vision tasks. \cite{djurisic2022extremely} builds on this concept by both pruning and modifying late layer model activations for out-of-distribution detection. 
    \item \cite{ROME} used \enquote{causal mediation analysis} \citep{vig2020investigating} as a method for localizing knowledge within model weights by comparing the activations of model forward passes over two different inputs.
    \item \cite{turner2023activation} proposes a method to add vectors that encode human-understandable semantic information directly to the activations of LMs to steer their outputs.
    \item \cite{fort2023scaling} showed how to adversarially engineer activations to have harmful downstream effects on LM prompt completions.
\end{itemize}

\section{Model Editing}
\label{RW:model_editing}

\textbf{Model editing} aims to change specific facts, associations, or information embedded in an LM outside of the constraints of traditional model training. Model editing requires the ability to localize learned information within subsets of the weight space and employs efficient and targeted methods to change this information while mitigating its effects of other information also embedded in the weight space. Model editing can be used to remove or alter private information, incorrect information, outdated information, biased information, and harmful information stored within model weights \citep{wu2023depn,yan2024potential,chen2023purr,wang2024earth}. Model editing can enable machine learning models to more exactly reflect human knowledge, without the massive overhead cost of typical model pre-training/fine-tuning.
\cite{zhu2020modifying} proposes an approach to modify specific learned facts encoded withing a LM's weights, while preserving model performance on other previously learned knowledge via a constrained optimization problem. 
\cite{dai-KnowledgeNeurons-2022} developed attribution methods to decipher which neurons are responsible for specific facts within languages models and developed methods to manipulate these neurons to edit a given fact. \cite{DeCao2021EditingFK, mitchell_FastModelEditing_2022} both propose hypernetwork based approaches to edit facts within models. Hypernetworks are additional networks that are trained to predict which weights are responsible for a given fact and how to modify the weights of a given neural network to better represent the desired knowledge. \cite{ROME} proposed  Rank-One Model Editing (ROME): by interpreting multi-layer perceptrons as key-values stores, ROME is able to replace specific keys-value pairs to override old or establish new knowledge associations in the model.

\section{Circuit Discovery}
\label{RW:circuits}

\textbf{Circuits} are sparse subsets of neural network weights that are responsible for (sub)sets of model behavior. Interpreting neural networks as circuits is useful as it allows researchers to localize sources of model behavior and may even help them better understand how stochastic training processes compress knowledge and skill into weights.

It was not always obvious that it was tractable to attribute model behavior to specific model components as these models are increasing massive (e.g. million, billion, trillion parameter scales). However, much research has shown that neural networks are very sparse: a small subset of weights are often responsible for much of a model's behaviors. For example, \cite{frankle2018lottery} showed that neural networks contain \enquote{winning lottery tickets}: sparse sub-networks within trained NNs that are nearly as performant as the original dense network. Follow up work, by \cite{amid2022learning} further developed methods to extract performant sub-networks from randomly initialized dense NNs. Results about the sparsity of models held across neural network architectures \citep{han2017ese, chen2020lottery, behnke2020losing}. 

Further work attempted to take this work a step further by designing methods to attribute more specific model behaviors to individual model components. \cite{elhage2021mathematical} conducted a detailed analysis of the types of circuits that appear in zero, one, and two layer transformers. \cite{chintam-etal-2023-identifying} identified components of transformer-based LMs for gender bias. \cite{nanda_ProgressMeasuresGrokking_2022} reverse engineers the algorithm implemented by a one layer transformer for modular addition. While the types of behaviors exhibited by a given model can be large and diverse, the workflow to discover circuits share many similarities. \cite{conmy2023automated} outlines a typical circuit discovery workflow for many ML interpretability pipelines and proposes a framework automate the workflow.

\section{Knowledge Extraction}
\label{RW:knowledge_extraction}

\textbf{Knowledge extraction} in language modeling is the practice of discovering what information is embedded in a LM's weights. The practice of knowledge extraction has grown in popularity as it became better known that LMs can be treated as successful knowledge stores. For example, \cite{roberts2020much} showed that by fine-tuning an LM on question answering tasks, the LM was able to successfully perform question-answering in a closed-book setting. This finding implied that models were good at storing knowledge during training and retrieving knowledge during inference time. In the context of language modeling, knowledge extraction is useful because it illuminates what a model knows well and what information it might be lacking. This enables ML developers to stage the appropriate interventions to improve model performance on desired tasks (e.g., further fine-tuning, knowledge editing). A simple method to extract model knowledge is to prompt the model and observe the outputs. In a closed-loop model prompting scenario the model would have to rely on its internal knowledge store in order to appropriately respond to a prompt. Therefore, based on the model outputs for any given prompt, the prompter would be able to infer what information the model is storing in its weights. \cite{petroni2019language,jiang2020can} both design prompting strategies to elucidate what knowledge is contained in LMs. The immediate shortcoming with prompting a model to elicit information is that models will only output information that it deems relevant to the prompt. Therefore, vanilla prompting knowledge extraction strategies may fail to uncover a model's full breadth of knowledge. It is often challenging to come up with a comprehensive prompting scheme to enable a model to exercise its entire knowledge store. 
To combat this, researchers have also devised more rigorous knowledge extraction techniques:
    \begin{itemize}
        \item \cite{cohen2023crawling} proposes a strategy to extract a knowledge-graph (KG) of facts from a LM: given a seed entity, they \enquote{crawl} the KG via prompts that are designed for both precision and recall.
        \item \cite{zhong2021factual} demonstrated that by training probes, rather than using discrete prompts, to illicit knowledge from LMs, they were able to tighten the lower bound on knowledge extraction benchmarks like LAMA \citep{petroni2019language}.
        \item \cite{elazar-MeasuringAndImprovingConsistency-2021} proposed a novel framework to assess if facts known to a LM are generalizable. By using a paraphrasing technique, they show that models are often inconsistent in reporting facts thus implying they do not contain generalizable facts.
    \end{itemize}

\subsection{Memorization}
\label{RW:memorization}
\textbf{Memorization} is an undesirable phenomena observed in LMs: models can be prompted to output their training data verbatim \citep{feldman2020neural}. Eliciting memorized data from a LM can be viewed as a subset of general purpose knowledge extraction tasks. However, while many forms of knowledge extraction are for benign purposes, such as gauging and improving a model's knowledge base, memorized data extraction can have harmful consequences. Memorized data can contain sensitive and/or private data that should not be recoverable by a model prompter. Dataset extraction attacks aim to prompt a model in a manner such that the model regurgitates its training data. \cite{carlini2021extracting} proposed one of the first training data extraction attacks from LMs.  \cite{nasr2023scalable} designed a black-box model prompting scheme to extract training data from LMs. Many works have attempted to better understand the causes of memorization. \cite{kandpal2022deduplicating} finds that deduplicating text may result in models memorizing less training data. A recent work by \cite{carlini2023quantifying} finds that there are 3 main reasons for memorization: 1) larger model scale, 2) data duplication, 3) larger input context length and attempts to quantify how much of a model's pre-trained data is memorized.

\section{Language Model Reasoning}
\label{RW:reasoning}
\cite{huang2022towards} defines reasoning as \enquote{a cognitive process that involves using evidence, arguments, and logic to arrive at conclusions or make judgments.} Reasoning has been studied as an aspect of human behavior in fields like psychology \citep{wason1972psychology} and philosophy \citep{passmore1961philosophical}. With the recent advances in conversation-based language modeling \citep[inter alia]{brown_LanguageModelsAre_2020, chowdhery2023palm, chung2022scaling, openai2022chatgpt}, researchers have begun to investigate the possibility of reasoning skills emerging in models. LMs have been showed to exhibit emergent behaviors, including the ability to \enquote{reason}, as their architecture sizes increase \citep{wei2022emergent}. Reasoning is measured in LMs by evaluating them on end task performance. Examples reasoning task include:
    \begin{itemize}
        \item Arithmetic reasoning: the ability to apply mathematical concepts to solve problems. Examples of arithmetic reasoning benchmarks are GSM8k \citep{cobbe2021training}, Math \citep{hendrycks2021measuring}, MathQA \citep{amini2019mathqa}, SVAMP \citep{patel2021nlp}, ASDiv \citep{miao2021diverse}, AQuA \citep{ling2017program}, and MAWPS \citep{roy2016solving}.
        \item Common Sense reasoning: the ability to use commonly known knowledge to make decisions in unknown situations. Examples of commonsense reasoning benchmarks are CSQA \citep{talmor2018commonsenseqa}, StrategyQA \citep{geva2021did}, and ARC \citep{clark2018think}.
        \item Multi-hop reasoning: the ability to synthesize related facts for answer questions with answers require many dependencies. Examples of multi-hop reasoning benchmarks include 2WikiMultiHopQA \citep{2wmh}, and HotpotQA \citep{yang2018hotpotqa}.
    \end{itemize}
To elicit reasoning abilities from pre-training LMs, much work demonstrated notable performance gains via new prompting strategies. For example, \cite{wei_chain-thought_2023} demonstrated that using a chain-of-thought prompting paradigm greatly improved reasoning abilities in LM. Follow up work from \cite{wang2023selfconsistency} introduced the importance of self-consistency in chain-of-though prompting scenarios. Following these works, many works have innovated on the \enquote{x-of-thought} prompting paradigm \citep{yao2023tree, besta2023graph, sel2023algorithm}. As interest in eliciting reasoning abilities from LMs grew, so did interest in understanding how LMs conducted reasoning. Researchers have tried to explain how models seem to \enquote{reason}. \cite{geva_FeedForwardKeyValue_2021} finds that feed-forward layers in LMs act as knowledge stores which can be queried by the model when certain input prompts require additional knowledge. \cite{geva_dissecting_2023} reverse engineers how transformers are able to recall facts. \cite{hou2023towards} posits that models \enquote{reason} by building internal tree-like representations of multi-hop reasoning processes.

\section{Retrieval Augmented Generation}
\label{RW:RAG}

\textbf{Retrieval Augmented Generation} (RAG) is a method of supplementing LMs with external sources of information as they respond to prompts. \cite{lewis2020retrieval} studied RAG in the context of improving a LM's question answering (QA) ability: to enhance a LM's QA ability, the authors trained a neural retriever model that was able to traverse a vector database of Wikipedia articles and select the appropriate article to supply the the LM in conjunction with its input prompt. The authors demonstrated that LM's with RAG were able to outperform vanilla LM's in open domain QA tasks. In addition to enhanced QA ability, RAG boasts many benefits. \cite{ovadia2023fine} demonstrated that RAG outperformed conventional fine-tuning of model weights when encountering both knowledge seen during training and new knowledge. This meant that RAG was better at introducing new knowledge to LMs. For example, if a model was trained in year 2021 and it is desirable for this model to be able to answer questions about news events in 2022, it would be beneficial to use RAG (rather than vanilla fine-tuning) to introduce 2022 news articles to the model to enable it to answer questions about it. 
A recent survey by \cite{gao2023retrieval} reported that:
    \begin{itemize}
        \item RAG improved model interpretability, as model responses can be attributed to specific data sources.
        \item RAG models inherently may have a greater breadth of knowledge, due to the external knowledge database being vast. Vanilla LMs are constrained by the fact that all of their knowledge must be able to be compressed into their weight space during training. 
        \item Model inference using RAG would increase latency (due to the retrieval step) and thus may be constrained by computational resources. However, RAG does not have the same fine-tuning computational costs that vanilla LM's do, therefore it is beneficial to do a case-by-case analysis when considering cost of RAG.
    \end{itemize}
        
In addition to QA, many works have explored unique applications for RAG in language modeling. \cite{liu2020retrieval} demonstrated the use of RAG in the context of code summarization. \cite{chen2022re} developed a pipeline to augment a text-to-image model with a multi-modal database (image, text) pairs to enhance image generation capabilities. \cite{komeili2021internet} augmented dialogue based LM with the ability to do internet search queries and showed superior dialogue performance.

RAG is a promising technology with which to augment language modeling abilities, and many opportunities for innovation exist: \textit{How do we retrieve the most useful information? How do we best encode this information before supplying it to an LM?}

%% file: chapter1/1_introduction.tex
\begin{figure}[t]
    \centering
    \subfloat[Multi-hop prompt.]
    {
	   \includegraphics[width=0.875\linewidth]{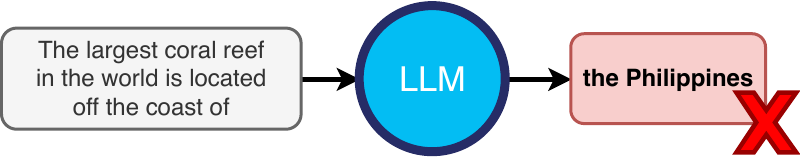} 
    } 

    \subfloat[Multi-hop prompt broken into $2$ single-hop prompts.]
    {
	   \includegraphics[width=0.875\linewidth]{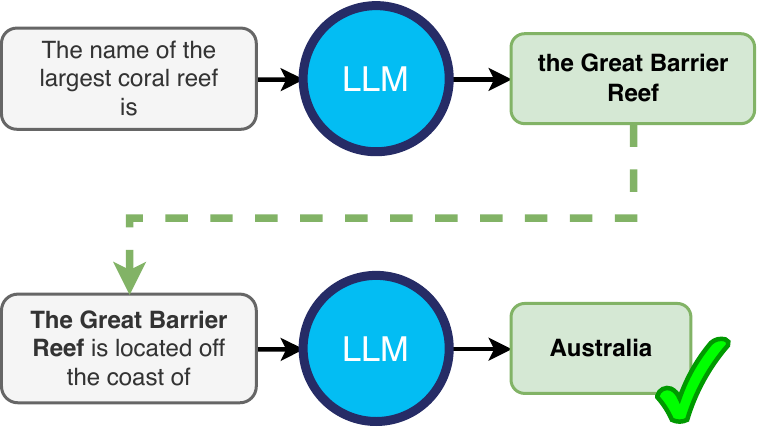} 
    } 
    \caption{A multi-hop prompt vs.\ two analogous single-hop prompts. The outputs are from \gptsmall{}.}
    \label{subfig:multi_hop_demo}
\end{figure}

Transformer-based \emph{Large Language Models} (LMs) \citep{vaswani2017attention, brown_LanguageModelsAre_2020} have shown exceptional promise for basic knowledge retrieval and language generation; however, they often lack the ability to perform basic reasoning tasks \citep{arkoudas2023gpt,guo2023indeed,blair2023can}. In this work, we focus on the simple task of answering multi-hop prompts (i.e., prompts in which the subject is not stated explicitly), which humans handle easily but with which LMs often struggle  (see Fig.~\ref{subfig:multi_hop_demo}). 

Researchers have attempted to rectify multi-hop reasoning failures by using various prompting methods such as \emph{Chain-of-Thought}~(CoT), \emph{Tree-of-Thought}~(ToT), and \emph{Graph-of-Thought}~(GoT) reasoning \citep{wei_chain-thought_2023,wang2023selfconsistency, tot_long2023large,tot_xie2023decomposition, yao2023tree,besta2023graph}. However, these approaches often put the burden on users to know how to elicit desired responses---and, in the hands of non-expert users, can lead to unreliable prompt completions. Researchers have also proposed model editing \citep{ROME, meng2022mass, mquake, li2023pmet} approaches that may hard-code distant relationships directly into model weights, rather than enhancing the model's abilities to recall and then link simpler relationships. These approaches can be computationally expensive and have unintended effects on other knowledge originally embedded in the model's weights \citep{cohen2023evaluating}. 

Our approach to this problem is based on the hypothesis that LMs often fail to recall relevant \textit{memories} when attempting to answer a prompt that requires multiple \enquote{hops} of reasoning, rather than lacking knowledge of the \textit{memories} altogether. For example, when attempting to complete the multi-hop prompt, \enquote{The largest coral reef system in the world is located off the coast of\ldots,} we hypothesize that the model does not correctly recall that \enquote{the largest coral reef system in the world} is \enquote{the Great Barrier Reef} before predicting the next token in the sequence. Yet the model can accurately complete both the corresponding single-hop prompt \enquote{The Great Barrier Reef is located of the coast of\ldots,} and, when prompted,  \enquote{the largest coral reef} as \enquote{the Great Barrier Reef.} Clearly, this information was encoded in the model during training but is not incorporated when answering questions that reference the prompt's subject indirectly. In this case, therefore, we  define the missing \emph{memory} to be \enquote{the Great Barrier Reef.} 

To study our hypothesis, we first attempt to reverse engineer a key mechanism by which transformer-based LMs conduct reasoning. Specifically, we find that in transformer-based models it is attention heads, rather than multi-layer perceptrons, that are responsible for retrieving \emph{memories} critical to successful model predictions; our finding is further substantiated by similar findings by \citet{li2023pmet,geva_dissecting_2023,dar2022analyzing}. We then study instances in which this mechanism fails in multi-hop reasoning tasks and find that this mechanism is likely the source of incorrect, insufficient, or irrelevant \textit{memory} retrievals \textbf{(Contribution 1)}---for an example, see Fig.~\ref{fig:transformer diagram}.

We then propose a lightweight \emph{memory injection} method that can be employed to correct a multi-hop reasoning failure during inference \textbf{(Contribution~2)}. As an example: by employing our method to inject the \emph{memory} of \enquote{The Great Barrier Reef} into the multi-hop prompt \enquote{The largest coral reef system in the world is located off the coast of\ldots} during inference, we increase the probability of the next token \enquote{Australia} by $189\%$; refer to Fig.~\ref{fig:injection_mechanism} for details. 

For our analyses, we hand-crafted a dataset for interpretabilty purposes \textbf{(Contribution 3)} and make use of a larger programmatically-generated dataset---refer Table~\ref{tab:data} for more information.

Finally  we conduct additional experiments \textbf{(Contribution 4)} to:
\begin{enumerate}
    \item Identify the ideal layer and magnitude for the \textit{memory injection}.
    \item Demonstrate the significance of curating prompt-specific \textit{memories} for injection.
    \item Analyze if \textit{memories} drawn from different parts of speech---namely, nouns, adjectives, adverbs, conjunctions, verbs---behave differently during \textit{memory injection}.
\end{enumerate}

%% file: chapter1/2_background.tex
We define single- vs.\ multi-hop prompts and provide a formal definition of the transformer model.

\subsection{Multi-hop vs. single-hop prompts}

We refer to a prompt as \textit{single-hop} if the subject of the relation is stated explicitly in the prompt, and \textit{multi-hop} otherwise. Multi-hop prompts refer to their subject in a way that requires an additional \enquote{hop} or inference step. For example, consider the single-hop prompt, \enquote{\emph{George Washington} fought in the\ldots} with a correct answer being \enquote{Revolutionary War.} In the analogous multi-hop prompt, \enquote{\emph{The first president of the United States} fought in the\ldots,} a preliminary inference step is needed to identity of the first US president before predicting the next token. For additional examples of single- and mutli-hop prompts, see Table~\ref{tab:prompt_examples} in the appendix.

\subsection{Transformer Architecture}
We introduce a common notation for the components of the transformer-based language model architectures that are the focus of our analyses. 
Specifically, we focus on auto-regressive, decoder-only models. 
We adopt much of our notation from \citet{elhage2021mathematical} and \citet{geva_dissecting_2023}.

\subsubsection{Embedding Inputs}
\label{sec:embed_inputs}
An input text is parsed into $N$ distinct tokens $t_0, \cdots, t_N$. Each token $t_i$ is then embedded as $x_{i}^{0} \in \mathbb{R}^d $ via an embedding matrix $ W_{E}  \in  \mathbb{R}^{|V| \times d}$, where $V$ is the vocabulary and $d$ is the hidden dimension.

\subsubsection{Residual Stream}
\label{sec:resid_stream}
Following the embedding layer, all tokenized embeddings $x_i^0$ are passed through a series of \emph{residual blocks}. The outputs of each \emph{residual block} are added back into the model's \emph{residual stream} denoted by $R^{\ell}\;(\forall \ell\in\{1,\cdots,L\}
)$ where $L$ is the number of layers in the LM.

We define the \emph{residual stream} at layer $\ell$ as:
\begin{equation}
    R^{\ell} = [x^{\ell}_0, \cdots, x^{\ell}_N], 
\end{equation}

\noindent
where $x^{\ell}_i$ is the representation of token $i$ at layer~$\ell$. The \emph{residual stream} is updated by its respective \textit{residual block} $r^{\ell}$:

\begin{equation}
    R^{\ell+1} = R^{\ell} + r^{\ell+1},
\end{equation}

\noindent
and the output of a \textit{residual block} $r^{\ell}$ is:
\begin{equation}
     r^{\ell} = a^{\ell}+ m^{\ell},
\end{equation}

\noindent
where $a^{\ell}$ is the output of the \emph{Multi-Headed Self Attention}~(MHSA) layer and $m^{\ell}$ is the output of the \emph{Multi-Layer Perceptron}~(MLP). We define MHSA and MLP in the following sections.

\subsubsection{Multi-Headed Self Attention (MHSA)}
Each MHSA layer $\ell$ is defined via four parameter matrices $W^{\ell}_Q, W^{\ell}_K, W^{\ell}_V, W^{\ell}_O \in \mathbb{R}^{d \times d}\;(\forall \ell\in\{1,\cdots,L\})$ and the hyperparameter $H$ denotes the number of attention heads.
Following \citet{elhage2021mathematical} and \citet{geva_dissecting_2023}, we can further dissect our parameter matrices to better observe the relationship between unique sets of parameters and individual attention heads: 
$W_Q^{l,j}, W_K^{\ell,j}, W_V^{\ell,j} \in \mathbb{R}^{d \times \frac{d}{H}}$ and $W_O^{\ell,j} \in \mathbb{R}^{\frac{d}{H} \times d}$ for $j \in [1,H]$. Now, we can define the output of each MHSA $a^{\ell}$ as the sum of all attention head outputs,

\begin{equation}
    a^{\ell} = \sum_{j=1}^H h^{\ell,j},
\label{eq:mhsa_output}
\end{equation}

\noindent
where $h^{\ell,j}$ is the output of the $j^{th}$ head in layer $\ell$:

\begin{equation}
    h^{\ell,j} = A^{\ell, j}
    \big(R^{\ell-1} W_V^{\ell, j}\big) W_O^{\ell, j}.
\label{eq:attention_output}
\end{equation}

\begin{equation}
\small
    A^{\ell, j}= \softmax 
    \Bigg(
        \frac
        {\big(R^{\ell-1}W_Q^{\ell,j}\big) \big(R^{\ell-1}W_K^{\ell,j}\big)^T}
        {\sqrt{\sfrac{d}{H}}} 
         \odot M
    \Bigg)
\label{eq:attention}
\end{equation}

\noindent 
where the $\softmax(\cdot)$ is performed as a row-wise operation, $\odot$ is the Hadamard product, and $M \in \{0, 1\}^{N \times N}$ is an auto-regressive attention mask where masked token positions are set to $0$. 

\subsubsection{Multi-Layer Perceptron (MLP)}
Each MLP is defined via two parameter matrices $W^{\ell}_F, W^{\ell}_I \in \mathbb{R}^{d \times d_p}$ with inner-dimension $d_p$ and a nonlinear activation function, $\sigma$. 

\begin{equation}
    m^{\ell} = W^{\ell}_F\; 
    \sigma
    \Big(
        W^{\ell}_I \big(a^{\ell} + R^{\ell-1}\big)
    \Big)
\end{equation}

\subsubsection{Unembedding Predictions into Logits}
After the final \textit{residual block}, all token positions $x_i^{-1}$ will be projected back into the vocabulary domain via the \emph{unembedding matrix} $W_{U} \in  \mathbb{R}^{d \times |V|}$. The output of the last token position is the next token prediction of the model.

%% file: chapter1/3_experiments.tex
Our central aim is to better understand how the outputs of the attention heads affect model performance with respect to predicting the correct next token in prompts requiring single-hop reasoning versus in prompts requiring multi-hop reasoning. 

\subsection{Dataset Descriptions}

We employ three datasets in this work. Two, used to assess model prompt completion accuracy, are our own high-quality manually curated dataset of single and multi-hop pairs and a programmatically generated dataset of prompt pairs. The third comprises lists of words from common parts of speech, which we use to study how the effectiveness of our intervention varies with the part of speech of injected tokens.

\subsubsection{Programmatically Generated Dataset}
The 2WikiMultiHop~dataset~\citep{2wmh} contains pairs of knowledge triples\\
$\{(s_1, r_1, s_2)_1, (s_2, r_2, s_3)_2\}$, each with two subjects $s$ and a relationship $r$.
We used these knowledge triples, plus a set of predefined templates, to generate a set of pairs of single- and multiple-hop questions, \harddata{}\/: see Tables~\ref{tab:data} and~\ref{tab:prompt_examples}. 

For example, let
$ s_1$ =  \enquote{Lilli's Marriage,} $r_1 = $\enquote{director,} $s_2$ = \enquote{Jaap Speyer,} $r_2 =$ \enquote{country of citizenship,} $s_3$ = \enquote{Dutch.} Then for
\textbf{single-hop}, the template: \enquote{The $r_2$ of $s_2$ is \ldots $s_3$}, the prompt yields the prompt 
\enquote{The country of citizenship of Jaap Speyer is \ldots [Dutch]}; for 
\textbf{multi-hop}, the template \enquote{The {$r_2$} of the $r_1$ of $s_1$ is \ldots $s_3$} yields then the prompt:
\enquote{The country of citizenship of the director of Lilli's Marriage is \ldots [Dutch].}

\subsubsection{Human-Generated Dataset}
As evidenced by the example presented above, the \harddata{} dataset, while scalable, contains many grammatical flaws. Therefore, we construct an additional dataset for multi-hop reasoning with a focus on grammatical and factual correctness presented below.
We hand-crafted 106 (single-hop, multiple-hop) prompt pairs, each in the same form as those in \harddata{}: e.g.,
\textbf{single-hop}:
\enquote{St. Peter's Basilica is in the city of\ldots [Rome]} and 
\textbf{multi-hop:}
\enquote{The biggest church in the world is in the city of\ldots [Rome]}.
Each prompt pair was also evaluated by two external reviewers for factual and grammatical accuracy.
We hereafter refer to this dataset as \easydata{}; see Tables~\ref{tab:data} and~\ref{tab:prompt_examples}.

\input{tables/dataset_meta_data}
\input{tables/prompt_samples}

\subsubsection{Part of Speech Dataset}

We used a subset of the Corpus of Contemporary American English \citep{davies2011word} which compiles word frequencies \citep{davies2010corpus} to generate lists of (i) the most common words from various parts of speech: 824 adjectives, 331 adverbs, 40 conjunctions, 2635 nouns, 969 verbs, and (ii) the 5050 most common words overall (``top 5050'').

\subsection{Model Description}
We work with two pretrained GPT2 models \citep{radford_LMsOpenAI_2019}. 
\textbf{\gptsmall{}} has 12 layers, 12 attention heads per attention layer, and $\sim$160M parameters.
\textbf{\gptlarge{}} has 36 layers, 20 attention heads per attention layer, and $\sim$840M parameters.
Both have a vocabulary of $\sim$50K tokens.

\subsection{Tools \& System Setup}
We use the \textit{Transformer Lens} Python package \citep{transformer_lens} to cache, inspect, and construct interventions on model inference passes. 
We ran experiments on a single A100 GPU with $40$~GB RAM.
Experimental code, dependency information, and datasets are available on GitHub.\footnote{\url{https://github.com/msakarvadia/memory_injections}}

%% file: tables/dataset_meta_data.tex
\begin{table}[hbt!]
\scriptsize
\centering
\begin{tabular}{cccllllll}
\toprule
    \multicolumn{3}{}{} & \multicolumn{3}{c}{Single-hop} & \multicolumn{3}{c}{Multi-hop}
\\
\cmidrule(lr){4-6}
\cmidrule(lr){7-9}
    Data & Size & Model & Answer prob. & Surprisal & Prompt len. & Answer prob. & Surprisal & Prompt len.
\\
\midrule 
   \easydata{} & 106 & \gptsmall{} & 0.157 & 4.21 & 9.66 & 0.087 & 4.91 & 12.99 
\\

    \easydata{} & 106 & \gptlarge{} & 0.28 & 2.90 & 9.66 & 0.157 & 3.97 & 12.99 
\\

    \harddata{} & 1000 & \gptsmall{} & 0.0007 & 9.80 & 10.44 & 0.00086 & 9.64 & 14.00 
\\

    \harddata{} & 1000 & \gptlarge{} & 0.0023 & 8.71 & 10.44 & 0.002 & 8.57 & 14.00 
\\
\bottomrule
\end{tabular}
\caption{Properties of the datasets used in our work. \emph{Size}: Number of prompts. \emph{Answer prob.}: Average model probability model for expected next token. \emph{Surprisal}: Average model surprisal value for expected next token ($\mathit{surprisal} \triangleq -\log(p)$ where $p$ is a probability). \emph{Prompt len.}: Average tokenized length of prompt.}
\label{tab:data}
\end{table}

%% file: tables/prompt_samples.tex
\begin{table*}[!htbp]
\centering
\small
\begin{tabular}{p{0.08\linewidth}p{0.4\linewidth}p{0.45\linewidth}}
\toprule
    Dataset & Single-Hop Prompt & Multi-Hop Prompt \\
\midrule
\multirow{9}{*}{\easydata{}}
    & George Washington fought in the \ldots [Revolutionary War] 
    & The first president of the United States fought in the \ldots [Revolutionary War] 
\\ \cmidrule{2-3}
    & Burj Khalifa is located in the city of \ldots [Dubai]
    & The tallest building in the world is located in the city of \ldots [Dubai]
\\ \cmidrule{2-3}
    & Nelson Mandela brought an end to \ldots [Apartheid]
    & The first president of South Africa brought an end to \ldots [Apartheid] 
\\ \cmidrule{2-3}
    & John F Kennedy was assassinated by a person named \ldots [Lee Harvey Oswald] 
    & The 35th president of the United States was assassinated by a person named \ldots [Lee Harvey Oswald]
\\ \cmidrule{2-3}
    & The father of Hermes is \ldots [Zeus] 
    & The father of the Greek messenger god is \ldots [Zeus] 
\\
\midrule
\multirow{10}{*}{\harddata{}}
    & The place of birth of Dušan Hanák is \ldots [Bratislava] 
    & The place of birth of the director of I Love, You Love is \ldots [Bratislava] 
\\ \cmidrule{2-3}
    & The employer of Éric Rohmer is \ldots [Cahiers du cinéma] 
    & The employer of the director of Triple Agent is \ldots [Cahiers du cinéma] 
\\ \cmidrule{2-3}
    & The employer of Chip Gubera is \ldots [University of Missouri] 
    & The employer of the director of Academy of Doom is \ldots [University of Missouri] 
\\ \cmidrule{2-3}
    & Steve Vai received the \ldots [Grammy] 
    & The performer of The Attitude Song received the \ldots [Grammy]
\\ \cmidrule{2-3}
    & The place of death of Augustus II the Strong is \ldots [Warsaw]
    & The place of death of the spouse of Christiane Eberhardine of Brandenburg-Bayreuth is \ldots [Warsaw]
\\
\bottomrule
\end{tabular}
\caption{\textbf{Example prompts.} Single/multi-hop prompt pairs from \easydata{} and \harddata{} datasets.}
\label{tab:prompt_examples}
\end{table*}

%% file: chapter1/4_algorithm.tex
Recent work 
suggests that attention heads are knowledge retrievers during a model's inference pass \citep{geva_dissecting_2023, li2023pmet}. Extending this result to multi-hop prompts, we hypothesize that attention layers play an important role in retrieving memories relevant to the ``hop'' in a given prompt. Therefore we define two algorithms below: one for analyzing attention head outputs in embedding space and the other for injecting a targeted memory into a model's hidden activations in order to correct faulty/incomplete reasoning.


\subsection{Interpreting Attention Heads}

We want to further understand the outputs of individual heads, and more specifically assess if any individual attention heads are exercised differently by single-hop vs. multi-hop prompts. 

Inspired by Logit Lens \citep{logitlens}, we leverage the model's unembedding matrix to study the internal mechanism of each attention head. For attention head $j$ in layer $\ell$, $h^{\ell,j}$, we apply the model's unembedding matrix $W_U$ followed by a $\softmax(\cdot)$ operation 
and interpret the last token position (out of $N$ total tokens) as a set of probabilities over tokens in the vocabulary space: 
\begin{equation}
     \mathit{vocab}^{\ell,j} = \softmax(h^{\ell,j} W_U )_{N-1} 
\label{eq:interp_attn}
\end{equation}

\begin{figure}[H]
  \centering
      \includegraphics[width=\linewidth]{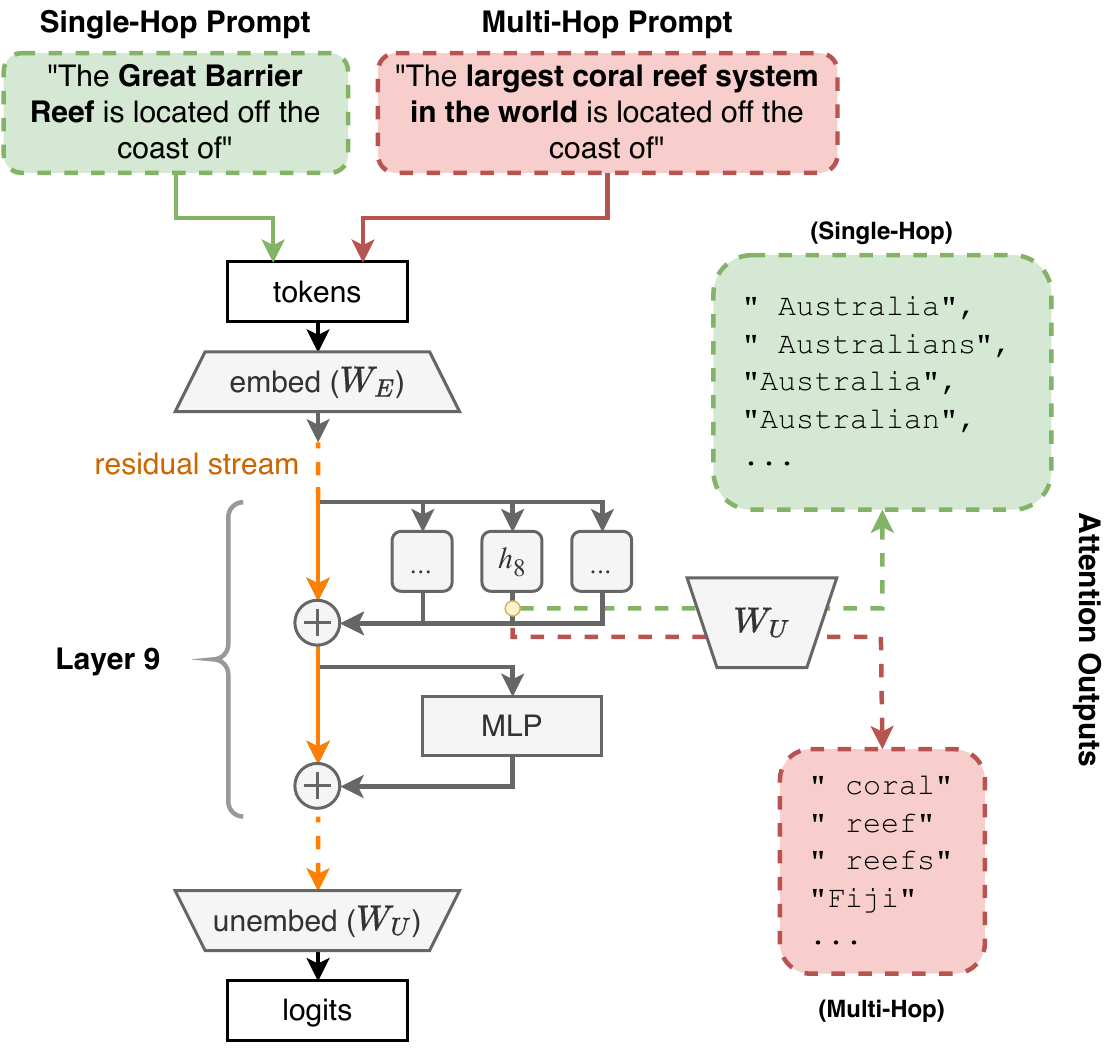}
  \caption{
        \textbf{Diagram of language model reasoning.} 
        Highest ranked attention outputs of \gptsmall{} at layer $\ell=9$, head $h=8$ when projected into vocabulary space (via the \gptsmall{} embedding matrix) for a single-hop prompt (green) and its multi-hop counterpart (red).
    }
  \label{fig:transformer diagram}
\end{figure}

\noindent
See in Fig.~\ref{fig:transformer diagram} an example of discrepancy in attention head behavior, when using Eq.~\eqref{eq:interp_attn}, for analogous single vs.\ multi-hop prompts. See additional examples in Table~\ref{tab:atten_head_outputs}.

A potential limitation of this approach is that it may portray attention head behavior inaccurately due to 
representational drift between model layers---and, like  \citep{logitlens}, may not generalize to other models. Nevertheless, we find it to be an effective preliminary tool for studying the function of attention heads in updating the output distribution. We leave the development of an interpretability tool that considers these drawbacks to future work.

\input{tables/attn_head_outputs}



\subsection{Memory Injections to Correct 
Failures} \label{subsec:correcting_multihop_failures}
Fig.~\ref{fig:transformer diagram} shows how Eq.~\eqref{eq:interp_attn} can reveal discrepancies between attention head behaviors for single- vs.\ multi-hop prompts.
We hypothesize that such discrepancies 
arise because the model, when updating the output distribution in each layer, fails to 
incorporate information about the implicit entity in the multi-hop prompt. This seems reasonable, as to retrieve information about an implicit entity one likely must first relate that entity to some explicit subject and then retrieve relevant information (hence our notion that processing prompts with implicit subjects requires an extra hop compared to those with explicit subjects).

\begin{figure}[H]
  \centering
    \includegraphics[width=\linewidth]{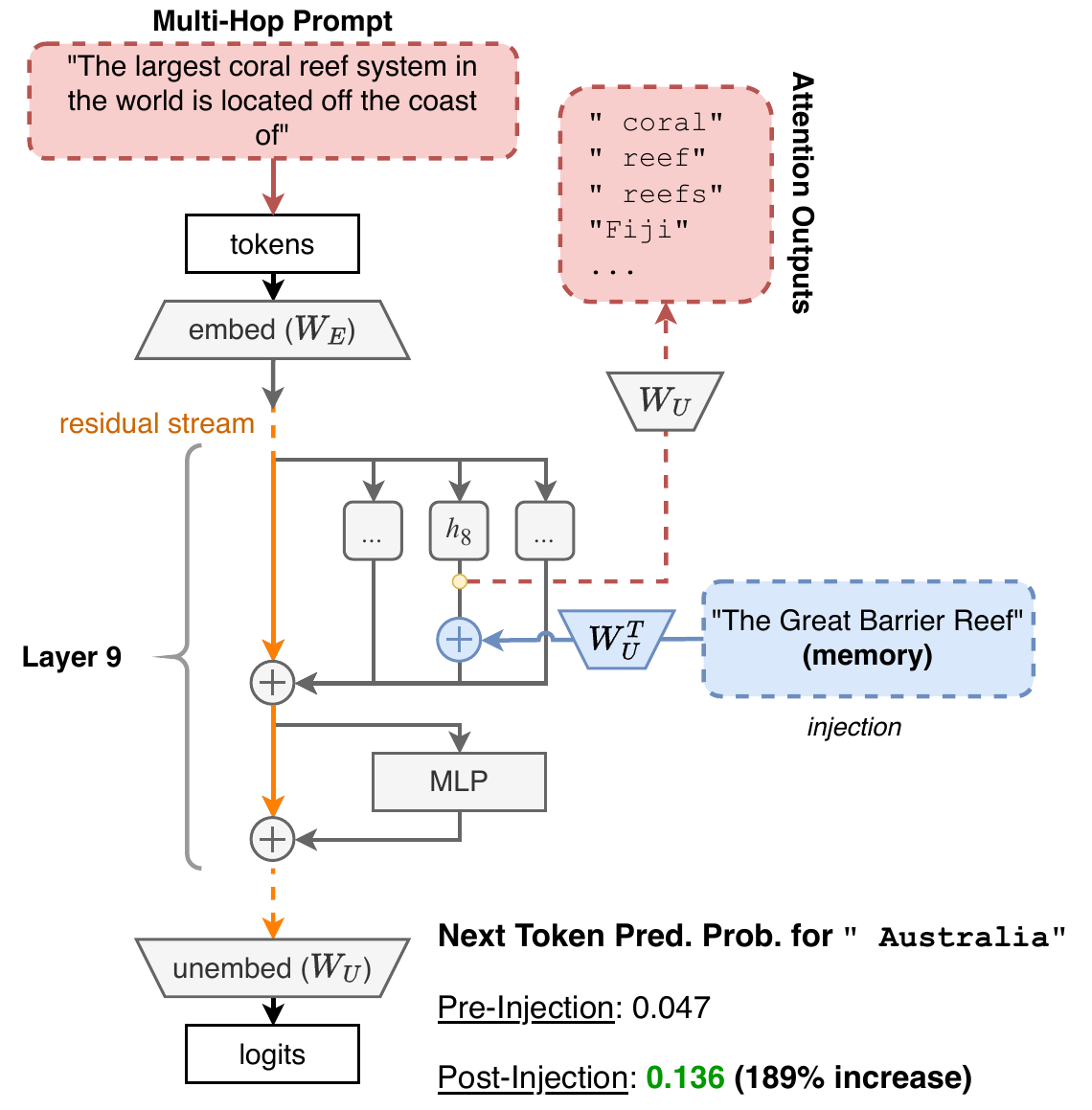}
    \caption{
        \textbf{Memory injection.} Injecting memory \enquote{The Great Barrier Reef} into \gptsmall{} hidden activations at layer $\ell=9$, head $8$, $\tau=4$.
    }
  \label{fig:injection_mechanism}
\end{figure}

Thus we design a 
method (see Fig.~\ref{fig:injection_mechanism}) for injecting a missing hop directly into the output hidden states of an attention head
before those outputs are added back into the transformer's residual stream:
\begin{enumerate}
    \item Let $m$ be a memory (a phrase, for example: \enquote{The Great Barrier Reef}) and let $\tau$ be the magnitude of the memory injection.

    \item Tokenize the memory $m$ into $t_0,\cdots,t_q$ where $q$ is the number of tokens. We encode each token $t_i$ into a one-hot vector $b_i \in \{0,1\}^{|V|}$ and sum all resulting one-hot vectors $b_i$ together into a binary vector $B\triangleq \sum_{i} b_{i}$. 
    
    \item Embed the binary vector, $B$, back into the model's latent space by applying the transpose of the unembedding matrix:
    \begin{equation}
        B^* = B\, W_U^T
        \label{eq:encode_mem}
    \end{equation}
    
    \item Then, to inject a memory at the attention layer of layer~$\ell$, add the embedded memory into the outputs of the attention heads during the inference pass:
    \begin{equation}
        a^{\ell} = \sum_{j=1}^H h^{\ell,j} + \tau  B^*
        \label{eq:inject_mem}
    \end{equation}
    
\end{enumerate}

\noindent
See additional examples of \textit{memory injections} in Table~\ref{tab:memory_injections}.

%% file: tables/attn_head_outputs.tex
\begin{table*}[!htbp]
\centering
\scriptsize
\begin{tabular}{p{0.1\linewidth}p{0.2\linewidth}ccp{0.45\linewidth}}
\toprule
    Prompt Type & Prompt & Layer $\ell$ & Head $h$ & Output\\
\midrule
\multirow{29}{*}{Single-Hop}
    & John F Kennedy was assassinated by a person named \ldots
    & 10
    & 0
    & [` Kennedy', ` JFK', ` Assass', ` assass', `Kenn', ` assassination', ` Cuba', ` Oswald', ` assassin', ` Cuban', ` Fidel', ` Bobby', ` Havana', ` assassinated', ` assassins', ` Jackie', ` Castro', ` Jinn', ` assassinate', `Mu', ` 1963', ` Kahn', ` drone', ` Cah', ` Mu', ` Ghosts', ` Soul', ` Laos', ` Cemetery', ` CIA']
\\ \cmidrule{2-5}
    & Barack Obama was a member of the  \ldots
    & 9
    & 8
    & [` Obama', `Obama', ` Maryland', ` America', ` JFK', ` Biden', ` Harlem', ` Washington', ` American', ` Clinton', ` White', ` Americans', ` Congressional', ` Harvard', ` Kennedy', ` FBI', ` Federal', ` CDC', ` DOJ', ` President', ` Georgetown', ` HHS', ` Barack', ` US', ` Trayvon', ` Connecticut', ` Holder', ` New', ` BLM', ` Baltimore']
\\ \cmidrule{2-5}
    & Cain murdered a person named \ldots 
    & 2
    & 1
    & [` police', `,', ` the', ` a', `\textbackslash n', ` and', ` violence', `.', ` death', ` in', ` criminal', ` of', ` to', ` victim', ` "', `-', ` at', ` victims', ` crime', ` from', ` an', ` that', ` murder', ` crimes', ` is', ` was', ` he', ` for', ` (', ` killed']
\\ \cmidrule{2-5}
    & Russia is mostly located on the continent of \ldots 
    & 9
    & 8
    & [` Moscow', ` Russian', `Moscow', ` Russia', ` Kremlin', ` Putin', `Putin', `Russia', ` Russians', `Russian', `\encodingerror', ` \encodingerror', ` Dmitry', ` Mikhail', ` Vladimir', ` Sergei', ` Siberia', ` Soviet', ` Siberian', ` Ukraine', ` Ukrainian', ` Sochi', ` Caucasus', ` Nikol', `Soviet', ` KGB', ` Dmit', ` USSR', `Ukraine', ` Ukrainians']
\\ \cmidrule{2-5}
    & George Washington fought in the \ldots 
    & 9
    & 8
    & [` Washington', `Washington', ` Virginia', `Virginia', ` Maryland', ` Congressional', ` Georgetown', ` Dull', ` Smithsonian', ` Maine', ` Burr', ` Jefferson', ` Navy', ` Capitol', ` congressional', ` FDR', ` Lexington', ` Byrd', ` Rhode', ` Roosevelt', ` Pike', ` Everett', ` Brookings', ` Madison', `apeake', ` Randolph', ` VA', ` Arlington', ` Americans', ` Lafayette']
\\
\midrule
\multirow{26}{*}{Multi-Hop}
    & The 35th president of the United States was assassinated by a person named \ldots 
    & 10
    & 0
    & [` assass', ` Assass', ` assassination', ` assassin', ` assassins', ` assassinate', ` Malik', ` bullets', ` gunmen', ` assassinated', `Mu', ` Pakistani', ` sniper', ` killings', ` JFK', ` Pakistan', ` homicides', ` Alger', ` lethal', ` Islamabad', ` Karachi', ` shooting', ` gun', ` gunshot', ` Mu', ` murder', ` killing', ` pistols', ` murders', ` gunned']
\\ \cmidrule{2-5}
    & The first black president of the United States was a member of the  \ldots
    & 9
    & 8
    & [` Negro', ` NAACP', ` blacks', ` black', ` Baltimore', ` White', ` negro', ` Washington', ` BLM', ` white', ` FBI', ` America', ` Maryland', ` African', ` Trump', ` Nixon', ` Charleston', ` Americ', ` KKK', `Washington', ` Virginia', ` racial', ` Blacks', `white', `White', ` nig', ` Black', ` Obama', ` Louisiana', ` whites']
\\ \cmidrule{2-5}
    & Adam and Eve's eldest son murdered a person named \ldots 
    & 2
    & 1
    & [`,', ` the', ` and', ` a', ` "', ` in', `\textbackslash n', `.', ` to', ` of', ` at', ` is', ` he', `-', ` that', ` was', ` for', ` police', ` from', ` on', " `", ` as', ` death', ` had', "'", ` an', ` his', "'s", ` said', ` told']
\\ \cmidrule{2-5}
    & The largest country in the world is mostly located on the continent of \ldots 
    & 9
    & 8
    & [`,', `\textbackslash n', ` the', ` and', `.', ` in', ` a', ` to', ` of', ` (', `-', ` for', ` that', ` "', `:', ` is', ` or', ` at', ` as', ` I', ` on', ` with', ` it', ` an', ` from', ` all', ` by', ` not', "'s", ` more']
\\ \cmidrule{2-5}
    & The first president of the United States fought in the \ldots 
    & 9
    & 8
    & [` Trump', ` Washington', ` America', `Washington', ` American', `Trump', `America', ` Obama', ` Donald', ` FBI', ` Congressional', ` Americans', `American', ` Nixon', ` Congress', ` congressional', ` White', ` Roosevelt', ` Republican', ` Negro', ` Clinton', ` JFK', ` Reagan', ` Virginia', ` FDR', `Obama', `Americans', ` Americ', `FBI', `Congress']
\\
\bottomrule
\end{tabular}
\caption{\textbf{Example of attention head outputs} from \gptsmall{} for \easydata{}.}
\label{tab:atten_head_outputs}
\end{table*}

%% file: chapter1/5_results.tex
We report, in turn, on our curated memory, random memory, and part-of-speech injection experiments.

\subsection{Curated Memory Injections}
\label{subsec:curated_mem_injec}

\begin{figure}[H]
  \centering
  \includegraphics[width=\linewidth]{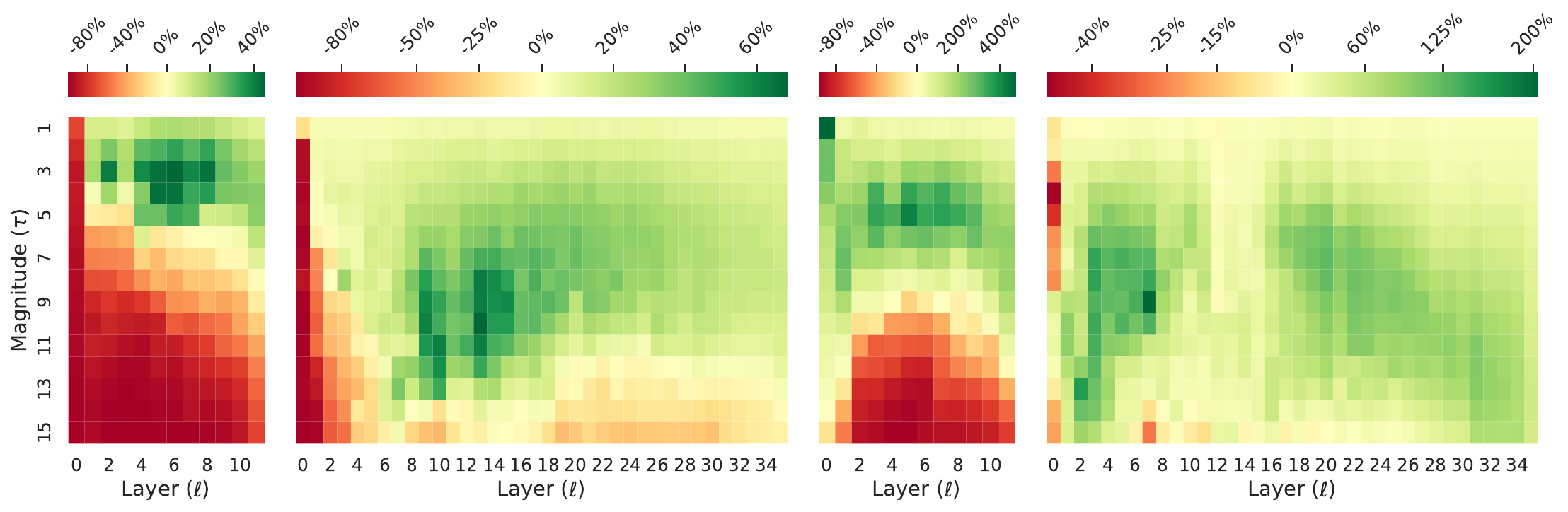}
  \caption{\textbf{Curated memory injections.} From left to right: \gptsmall{} + \easydata{}, \gptlarge{} + \easydata{}, \gptsmall{} + \harddata{}, \gptlarge{} + \harddata{}. Each cell in each heatmap is the average percent difference between the pre- and post-injection next token predictions for multi-hop prompts. Green cells denote a positive percent difference (i.e., correct prediction is more likely), while red cells denote a negative percent difference (i.e., correct prediction is less likely). When computing the averages for each 
  ($\ell$, $\tau$) pair we exclude outliers not within $\pm2$ standard deviations from the mean.
  }
  \label{fig:curated_memories}
\end{figure}

We hypothesize that a model's poor performance on multi-hop prompts is due to its inability to resolve the implicit subject (e.g., \enquote{The largest coral reef system in the world}) to an explicit subject (e.g., \enquote{The Great Barrier Reef}). This failure limits the later layers' ability to retrieve relevant information about this subject before predicting the next token. Therefore, in this experiment, we curate sets of tokens to inject into our model's residual stream such that it can resolve the explicit subject more easily. We further study the effect that the injection magnitude $\tau$ has on its success.

\textbf{Experimental design:}
For every multi-hop prompt in our datasets, we extract the explicitly stated subject from the corresponding single-hop prompt and inject those tokens as \emph{memories} into each attention layer as described in Section~\ref{subsec:correcting_multihop_failures}. For example, given the
\textbf{single-hop prompt} \enquote{\textit{The Great Barrier Reef} is located off the coast of\ldots} and the
\textbf{multi-hop prompt} \enquote{\textit{The largest coral reef system in the world} is located off the coast of\ldots,} the \textbf{memory} is
\enquote{\textit{The Great Barrier Reef.}}



We assess the effects of injection layer $\ell$ and magnitude $\tau \in [1, \cdots, 15]$ by enumerating the resulting change in accuracy for all combinations of these two parameters for both \gptsmall{} and \gptlarge{}.
We measure the success of a memory injection by calculating the percent increase between the model's predicted probability for the expected next token from the multi-hop prompt with and without the injection. A greater positive difference indicates a more successful injection.

\textbf{Discussion:}

Results are in Fig.~\ref{fig:curated_memories}. We observe that each model/dataset combination has an optimal layer $\ell$ and magnitude $\tau$ for memory injections: the darkest green areas, which signify the highest average percent increase in probability of the expected next token for the respective dataset. The best ($\ell$, $\tau$) pair injection results are in Table~\ref{tab:curated_vs_random}. Additional examples of memory injections are in Table~\ref{tab:memory_injections}.

\input{tables/memory_injection_examples}


\input{tables/curated_vs_random_edits}

\subsection{Random Memory Injections}
In Section~\ref{subsec:curated_mem_injec}, we identify ideal ($\ell$, $\tau$) pairs for each model and dataset for a curated memory injection. We now demonstrate that the results we observe are not spurious: i.e., the information that we inject at each head should be related to the explicit subject. We demonstrate the need for our particular injection routine by assessing the effects on model accuracy of randomly injecting tokens from various parts of speech.

\textbf{Experimental design:} 
We conduct targeted injections for the high-scoring ($\ell$, $\tau$) pairs identified via the experiment in Section~\ref{subsec:curated_mem_injec}, Table~\ref{tab:curated_vs_random}. 
Instead of injecting curated subject tokens, we select as candidate injections the 40 most common words from each of the adjectives, adverbs, conjunctions, nouns, verbs, and top 5050 subsets of our \textit{Part of Speech} dataset.
We then apply each word as an individual injection for every prompt in our multi-hop dataset at the ideal ($\ell$, $\tau$) pair. We term these injections \enquote{random,} as they were not curated to be relevant to our prompts.

\textbf{Discussion:}
The results are in the right half of Table~\ref{tab:curated_vs_random}. 
We observe that a random injection led, on average, to a degradation in predictive performance across most parts of speech considered, as indicated by a negative percent difference (decrease in correct answer probability) between the pre- and post-injection expected next token probabilities for multi-hop prompt completions. Additionally, no random injection result exceeded the performance of a curated injection.
These findings suggest that the choice of injected tokens is critical for improving multi-hop prompt completion success.

\subsection{Memory Injections for Parts of Speech}
\begin{figure}[H]
  \centering
      \includegraphics[width=\textwidth]{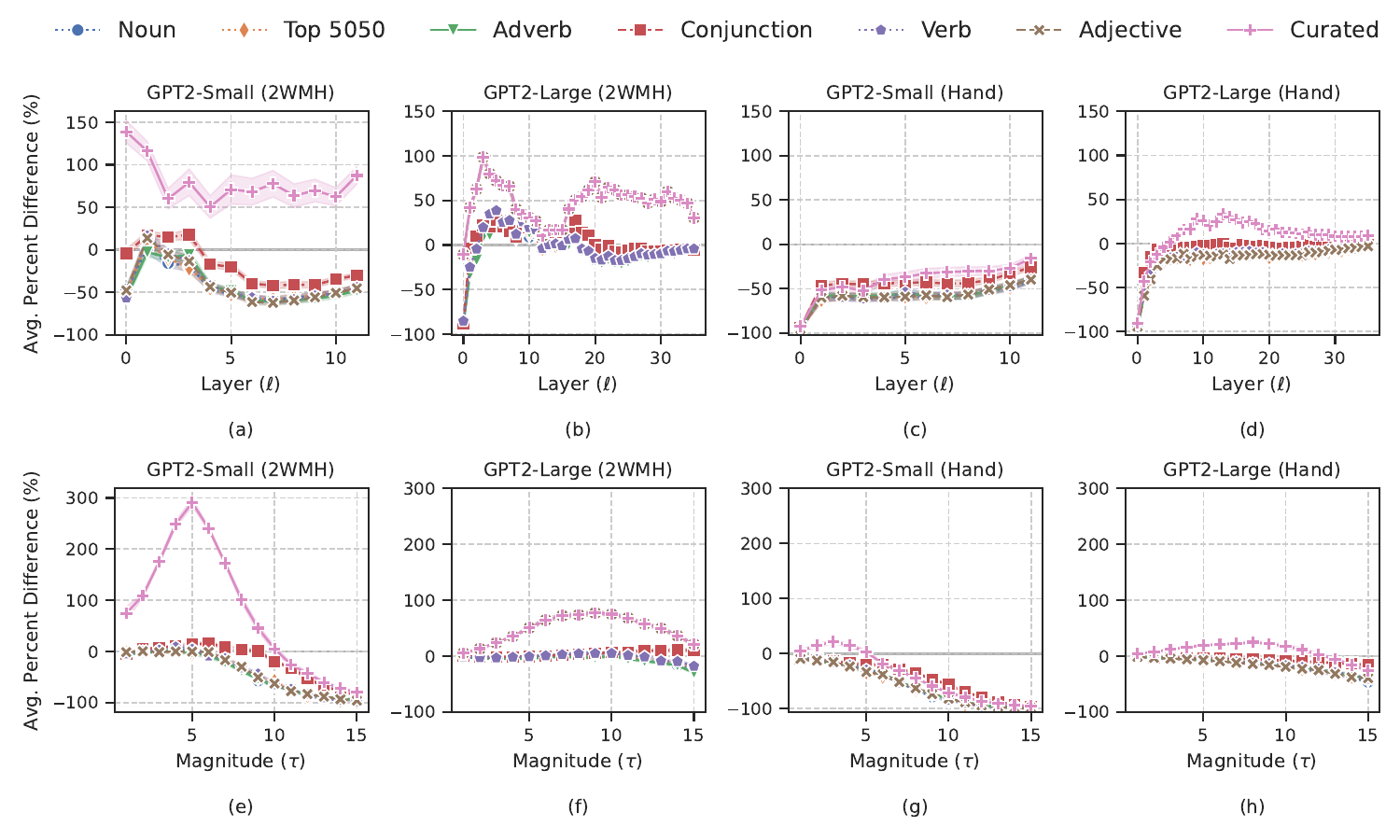}
  \caption{\textbf{Part of speech memory injections.} This figure shows the average effect of memory injections from various parts of speech as a function of layer $\ell$ (top row) and magnitude $\tau$ (bottom row). The standard deviation scaled by 10\% is pictured across magnitudes (top row) and layers (bottom row).}
  \label{fig:pos_result}
\end{figure}

We have tested curated vs. random memory injections at ideal ($\ell$, $\tau$) pairs. Now we assess whether memory injections from specific parts of speech more broadly have positive impacts on prompt completions, not just at the ideal locations for curated memories, but also at other ($\ell$, $\tau$) pairs.
Our hypothesis is that if a transformer-based LM has learned a division of labor regarding which attention layers are responsible for retrieving specific concepts (e.g., parts of speech) then this experiment might highlight those learned roles.

\textbf{Experimental design:} 
This experiment is identical to that of
Section~\ref{subsec:curated_mem_injec},
except that: (i) for
each part of speech $\mathit{pos} \in$ [adjectives, adverbs, conjunctions, nouns, verbs, top 5050], we use a randomly selected word:
e.g., \enquote{apple} from \enquote{nouns}; and (ii)
when searching for the ideal ($\ell$, $\tau$) pair for a given part of speech and  multi-hop prompt, we use a new random word for each injection.

\textbf{Discussion:}
The results are in Fig.~\ref{fig:pos_result}. We note that for no part of speech considered here does the average performance of the studied memory injections exceed that of the curated memory injections presented in Table~\ref{tab:curated_vs_random}. Additionally, memory injections from adjectives, adverbs, nouns, verbs, and top 5050 seemed to exhibit similar behavior. Memory injections from conjunctions, however, typically outperformed all other parts of speech. We hypothesize that this is because conjunctions 
often play a neutral role in prompt completions. Thus, while a random noun (e.g., \enquote{apple}) might distort prompt completion, a random conjunction (e.g., \enquote{and,} \enquote{for}) is less likely to do so.

We note also that for each part of speech, performance averaged over all injections for most ($\ell$,~$\tau$) pairs was reduced (< 0) for \easydata{} (refer Fig.~\ref{fig:pos_result}: subplots $c,d,g,h$), but was sometimes improved (> 0) for \harddata{} (refer Fig.~\ref{fig:pos_result}: subplots $a,b,e,f$). 
We attribute this result to the relative difficulties of the two datasets. 
\easydata{} has, on average, lower surprisals than does \harddata{}, as seen in Table~\ref{tab:data}, suggesting that there is additional information that the model could use successfully for \harddata{}, but not for \easydata{}.


These results (Figs~\ref{fig:gpt2_large_2wmh_pos}--
\ref{fig:gpt2_small_hand_pos}) suggest that while curated memories 
are ideal for correcting multi-hop reasoning failures, language models can also benefit 
from injections of different parts of speech. 
This result suggests 
that different parts of a language model 
(namely, early layers) serve specialized roles, 
with some dealing with processing related to specific parts of speech.

In future work we will curate relevant memories from various parts of speech for each prompt, to better understand the effects of curated memories.

\input{chapter1/z_appendix}

%% file: tables/memory_injection_examples.tex
\begin{table*}[t!]
\small
\begin{tabular}{p{0.4\linewidth}p{0.15\linewidth}p{0.15\linewidth}p{0.1\linewidth}p{0.1\linewidth}}
\toprule
\centering

    Multiple-Hop Prompt & Memory & Answer & Pre-injection Answer Prob. &  Post-injection Answer Prob.
\\
\midrule 
    \textbf{The God of Thunder} is the son of \ldots & Thor & Odin & $0.84\%$ & $3.37\%$ 

\\
\midrule 
    \textbf{The first president to be assassinated} succeeded in abolishing \ldots & Abraham Lincoln & slavery & $30.46\%$ & $63.09\%$ 

\\
\midrule 
    \textbf{The founder of Microsoft} was born in the city of \ldots & Bill Gates & Seattle & $1.55\%$ & $2.44\%$  

\\
\midrule 
    \textbf{The highest peak in the world} is located in the \ldots & Mount Everest & Himalayan & $3.40\%$ & $22.58\%$ 

\\
\bottomrule
\end{tabular}
\caption{\textbf{Examples of memory injections.} Injecting memories with $\tau=4, \ell=9$ into \gptsmall{}.}
\label{tab:memory_injections}
\end{table*}

%% file: tables/curated_vs_random_edits.tex
\begin{table*}[hbt!]
\small
\centering
\begin{tabular}{cccc|rrrrrrr}
\toprule
    \multicolumn{4}{}{} & Curated & \multicolumn{6}{c}{Random}
\\

\cmidrule(lr){5-5}
\cmidrule(lr){6-11}
    Model & Data & $\ell$ & $\tau$ &  Subject & Adj. & Adv. & Conj. & Noun & Verb & Top-$5050$ 
\\
\midrule 
    GPT2 Small & Hand & 7 & 3 & \textbf{45\%} & -7.6\% & -6.0\% & -6.3\% & -6.5\% & -7.5\% & -6.0\%
\\
    GPT2 Small & 2wmh & 6 & 5 & \textbf{424\%} & -17.1\% & -15.1\% & -10.3\% & -1.1\% & -1.2\% & 1.6\%
\\
    GPT2 Large & Hand & 14 & 10 & \textbf{68\%} & -8.1\% & -4.4\% & -4.9\% & -9.8\% & -6.0\% & -4.7\%
\\
    GPT2 Large & 2wmh & 8 & 9 & \textbf{204\%} & 13.0\% & 11.6\% & 3.5\% & 11.8\% & 4.3\% & 17.6\%
\\
\bottomrule
\end{tabular}
\caption{\textbf{Curated vs.\ random memory injections.} Table shows the ($\ell$, $\tau$) pairs for the best token injections, along with the \emph{average percent difference} (excluding outliers >$\pm2$ standard deviations from the mean) between pre- and post-injection expected next token predictions for multi-hop prompts. Each random injection column indicates 40 random injections from [Adjectives, Adverbs, Conjunctions, Nouns, Verbs, Top 5050] at the ideal ($\ell$, $\tau$).}
\label{tab:curated_vs_random}
\end{table*}

%% file: chapter1/z_appendix.tex
\begin{figure*}[hbt!]
  \centering
      \includegraphics[width=0.9\linewidth]{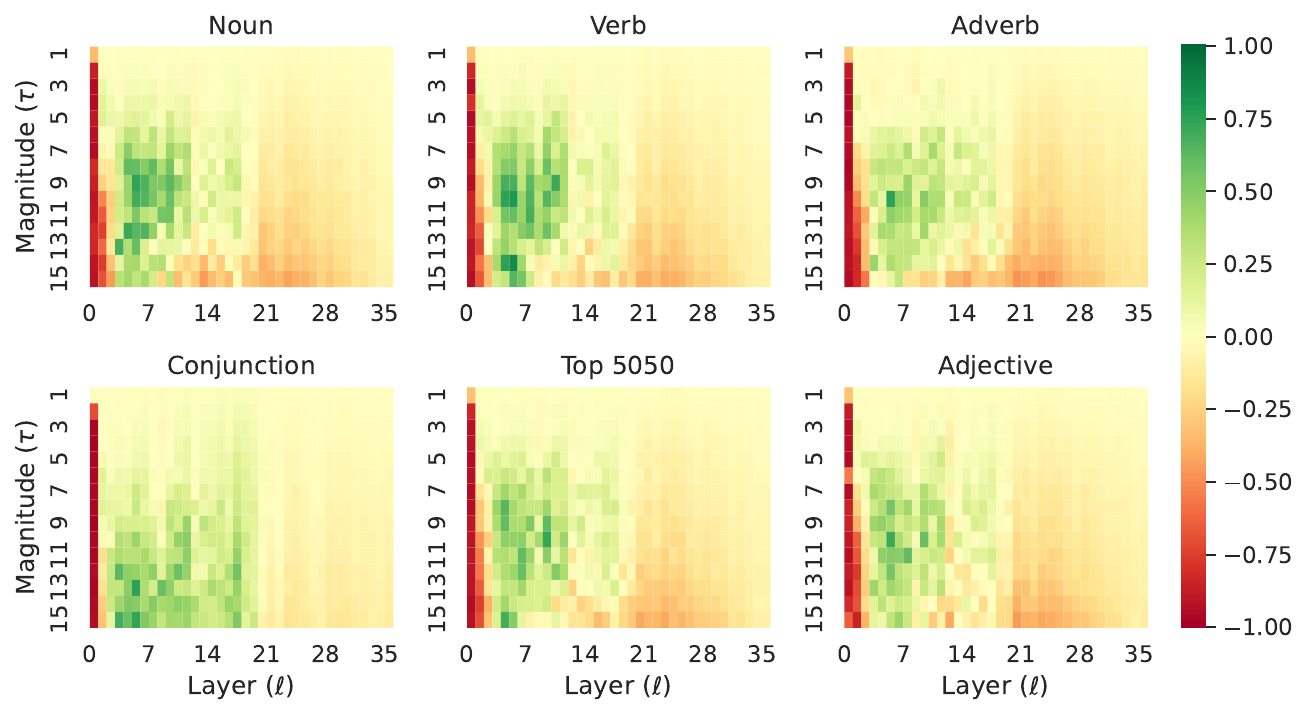}
  \caption{\textbf{GPT2-Large, \harddata{} dataset.} Heatmap shows average percent difference between pre- and post-injection answer probabilities for multi-hop prompts excluding outliers not within $\pm2$ standard deviations from the mean across various parts of speech.}
  \label{fig:gpt2_large_2wmh_pos}
\end{figure*}

\begin{figure*}[hbt!]
  \centering
      \includegraphics[width=0.9\linewidth]{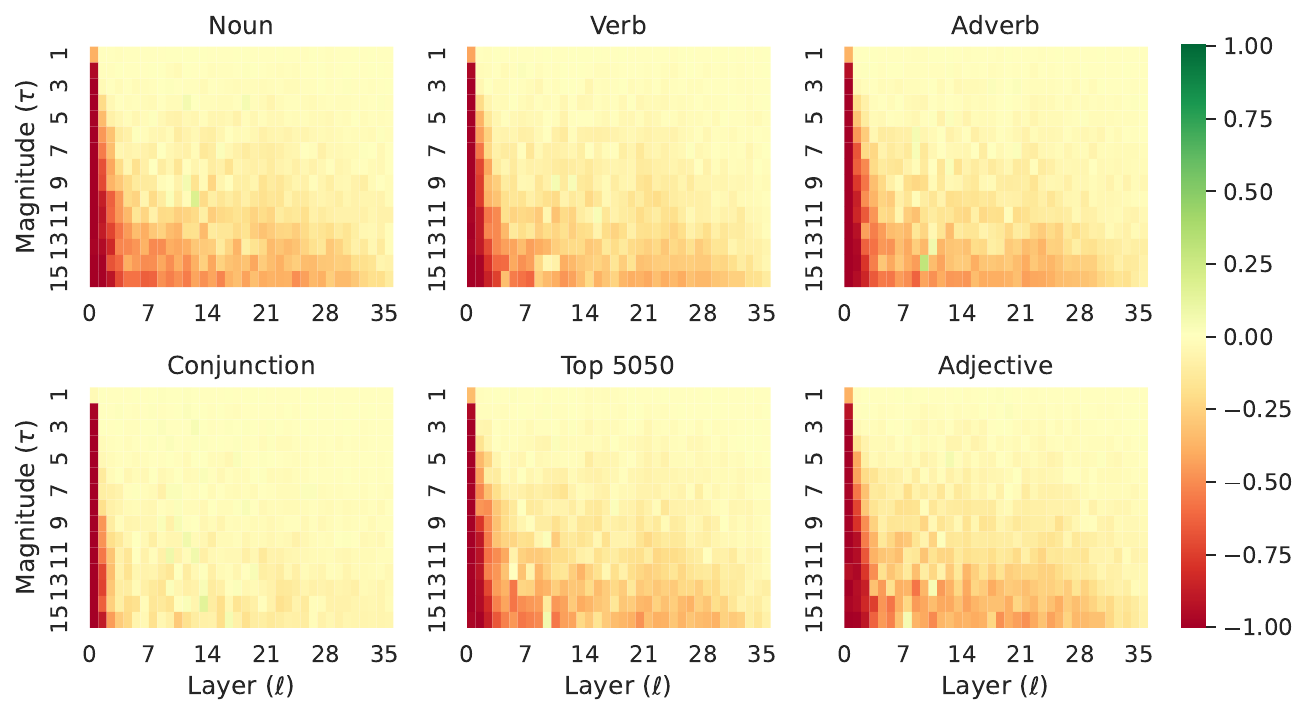}
  \caption{\textbf{GPT2-Large, \easydata{} dataset.} Heatmap shows average percent difference between pre- and post-injection answer probabilities for multi-hop prompts excluding outliers not within $\pm2$ standard deviations from the mean across various parts of speech.}
  \label{fig:gpt2_large_hand_pos}
\end{figure*}


\begin{figure*}[hbt!]
  \centering
      \includegraphics[width=0.9\linewidth]{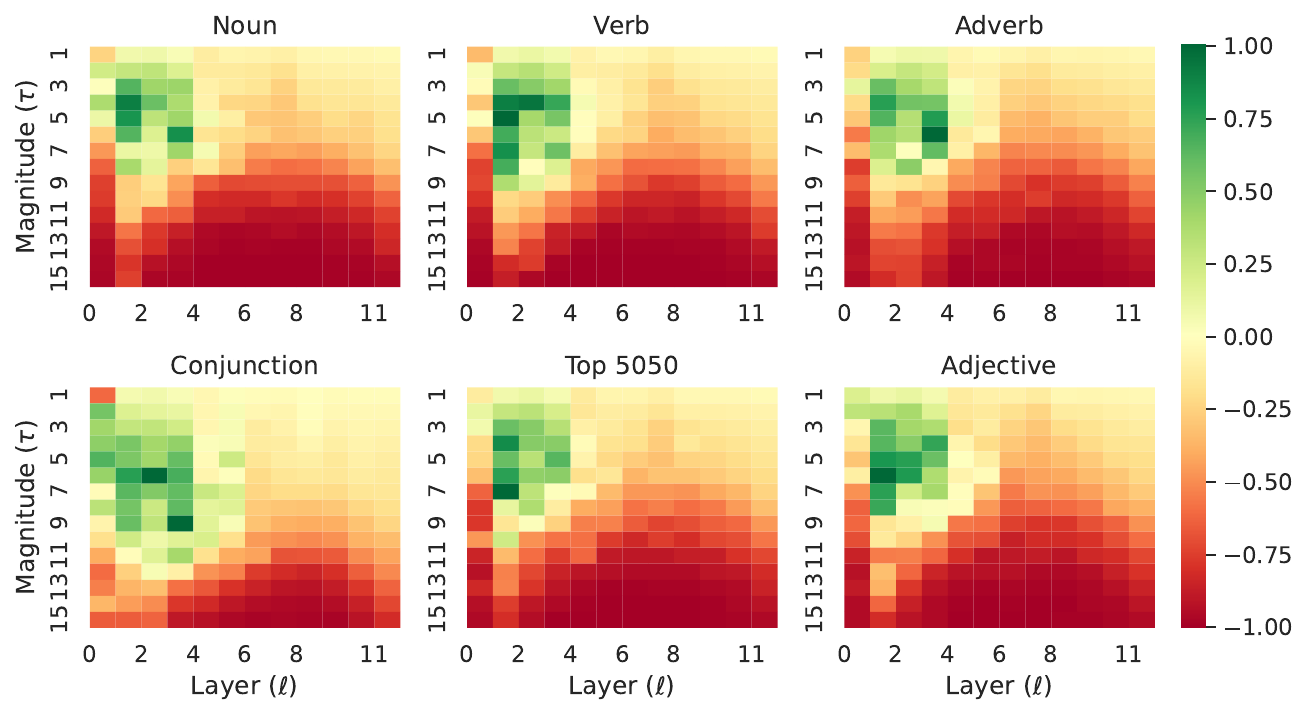}
  \caption{\textbf{GPT2-Small, \harddata{} dataset.} Heatmap shows average percent difference between pre- and post-injection answer probabilities for multi-hop prompts excluding outliers not within $\pm2$ standard deviations from the mean across various parts of speech.}
  \label{fig:gpt2_small_2wmh_pos}
\end{figure*}

\begin{figure*}[hbt!]
  \centering
      \includegraphics[width=0.9\linewidth]{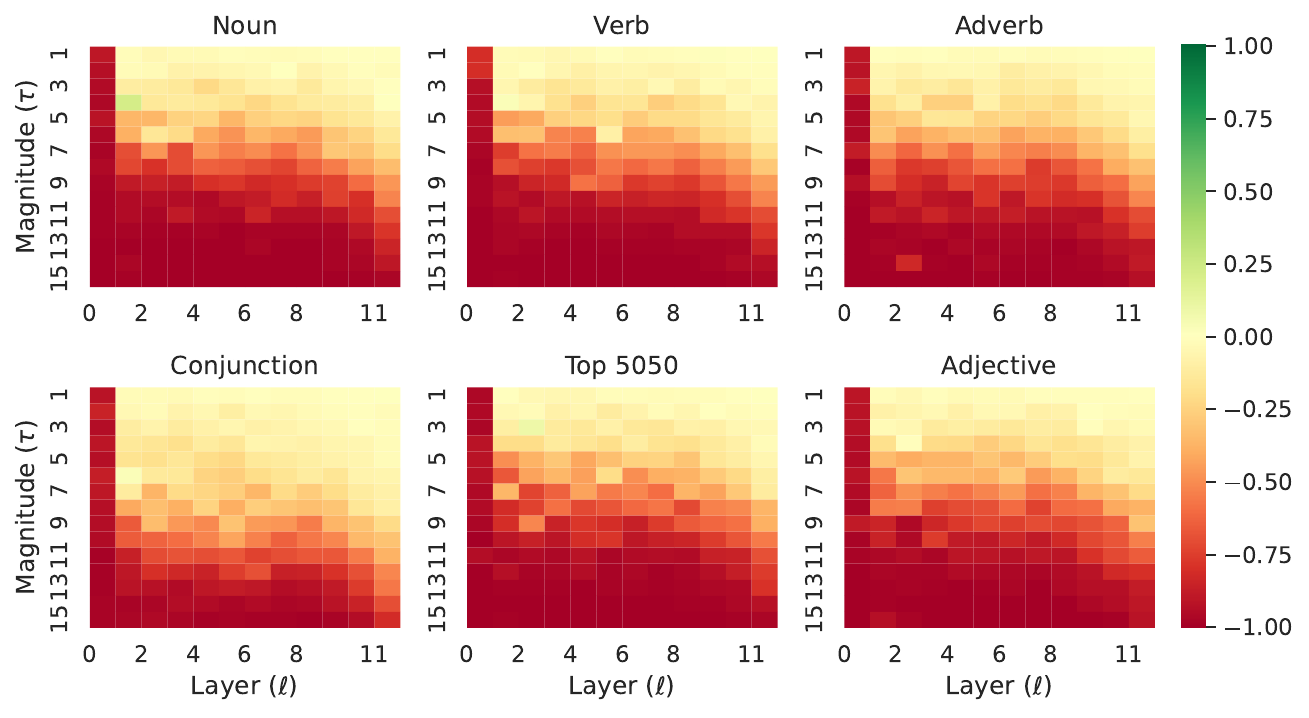}
  \caption{\textbf{GPT2-Small, \easydata{} dataset.} Heatmap shows average percent difference between pre- and post-injection answer probabilities for multi-hop prompts excluding outliers not within $\pm2$ standard deviations from the mean across various parts of speech.}
  \label{fig:gpt2_small_hand_pos}
\end{figure*}

\newpage



%% file: chapter1/more_experiments.tex
We investigate additional memory encoding styles and assess the performance versus computational cost trade-off between them. 

\subsection{Memory Encoding Styles}
Until now, we have only investigated one type of memory encoding style \eqref{eq:encode_mem} which we will refer to as \unembed{} as it makes use of the model's unembeding matrix $W_U$. Now, we introduce two additional encoding styles: \embed{}, and \layerwise{}.

\textbf{\embed{}} is mathematically equivalent to \unembed{} aside from the fact that we use the model's embedding matrix $W_E$, instead of unembedding matrix $W_U$ to encode the memory. To do an \embed{} encoding, once we have the binary vector of the memory, $B$, we embed it back into the model's latent space by applying the embedding matrix:
    \begin{equation}
        B^* = B\, W_E
        \label{eq:encode_mem_embed}
    \end{equation}

\textbf{\layerwise{}} memory encoding requires the memory to be run through the first $\ell$ layers of the model, where $\ell$ is the layer in which the memory will ultimately be injected into during inference. The steps for this type of memory encoding are as follows:
\begin{enumerate}
    \item Let $m$ be a memory (a phrase, for example: \enquote{The Great Barrier Reef}). Tokenize the memory and apply the model's embedding matrix to it as per \ref{sec:embed_inputs}. 
    \item Following the embedding layer, all tokenized embeddings $x^0_i$, of the memory, are passed through the first $\ell$ residual blocks of the model as per \ref{sec:resid_stream}. Let the model's residual representation $R^{\ell} = B^*$
    \item Note: $B^*$ will need to be recalculated if the intended layer of injection changes.
\end{enumerate}

For each of these encoding styles, \embed{}, \unembed{}, \layerwise{}, once we have the encoded memory $B*$, we employ the same method to inject it into the model as per equation~\eqref{eq:inject_mem}. 

\subsection{Encoding Style FLOP Counts}
\label{sec:flop_counts}

Following \cite{kaplan2020scaling}, we calculate approximately how many FLOPs are required to encode a memory. We use the following parameters when referring to transformer architecture hyperparameters: $n_{layer}$ (number of layers), $d_{model}$ (dimension of residual stream), $d_{ff}$ (dimension of intermediate feed-forward layer), $d_{attn}$ (dimension of the attention output), and $n_{heads}$ (number of attention heads per layer). As per convention, $d_{attn} = d_{ff}/4 = d_{model}$. Additionally, $n_{ctx}$ refers to the number of input tokens to the model; for \easydata{} $n_{ctx} = 2.96$ and for \harddata{} $n_{ctx} = 5.25$ on average, where $n_{ctx}$ refers to the average token length of the \enquote{memories} for the given dataset.

The FLOP counts for both the \embed{} and \unembed{} memory encoding styles can be calculated as:
\begin{equation}
    total_{flop} = n_{ctx} * d_{model}
\end{equation}

The FLOP counts for both the \layerwise{} memory encoding style can be calculated as:
\begin{equation}
    embed_{flop} = n_{ctx} * 4 * d_{model}
\end{equation}

\begin{equation}
    N = 2 * d_{model} * n_{layer} * (2 * d_{attn} + d_{ff})
\end{equation}
    
\begin{equation}
    ff_{flop} = 2 * N + 2 * n_{layers} * n_{ctx} * d_attn
\end{equation}    

\begin{equation}
    total_{flop} = embed_{flop} + ff_{flop}
\end{equation}

\subsection{Additional Model Descriptions}
We expand the models we study to: \gptsmall{}, \gptlarge{}, \gptxl{}, \gptneosmall{}, \gptneolarge, \gptneoxl, \gptj{}. Refer to table~\ref{tab:model_char} for additional model characteristics.

\gptsmall{}, \gptlarge{}, \gptxl{}, \gptneosmall{}, \gptneolarge{}, \gptneoxl{}  typically have tied embedding and unembedding weights; this means that the model shares the same weights for both the embedding and unembedding matrices. In the case of models with tied embeddings, the \embed{} and \unembed{} memory encoding strategies would yield equivalent results. In this work, however, we instantiate our model from a popular open-source Python library, \citep{transformer_lens}, which applies two post-processing steps to the model weights: centering the unembedding weights such that they have zero mean, and folding in the layer normalization weights into the model weights as per \cite{elhage2021mathematical}. These weight post-processing steps effect the embedding and unembedding weights differently as only the unembedding layer has a preceding layer normalization operation. Therefore, it is interesting and necessary to investigate both \embed{} and \unembed{} in the context of memory encoding schemes.

\input{tables/model_charecteristics}

\subsection{Memory Encoding Style Experiments}
In Section~\ref{sec:results}, we investigated the effect of using the \unembed{} encoding style on various memory types. Now, we investigate the effect of using the \embed{} and \layerwise{} encoding style in a memory injection to enhance a model's multi-hop reasoning capability.

\textbf{Experimental design:} 
This experiment is identical to that of Section~\ref{subsec:curated_mem_injec}, except that: rather than using the \unembed{} encoding style for the memories, we, in turn, use the \embed{} and \layerwise{} encoding styles.

\begin{figure*}[hbt!]
  \centering
      \includegraphics[width=0.7\linewidth]{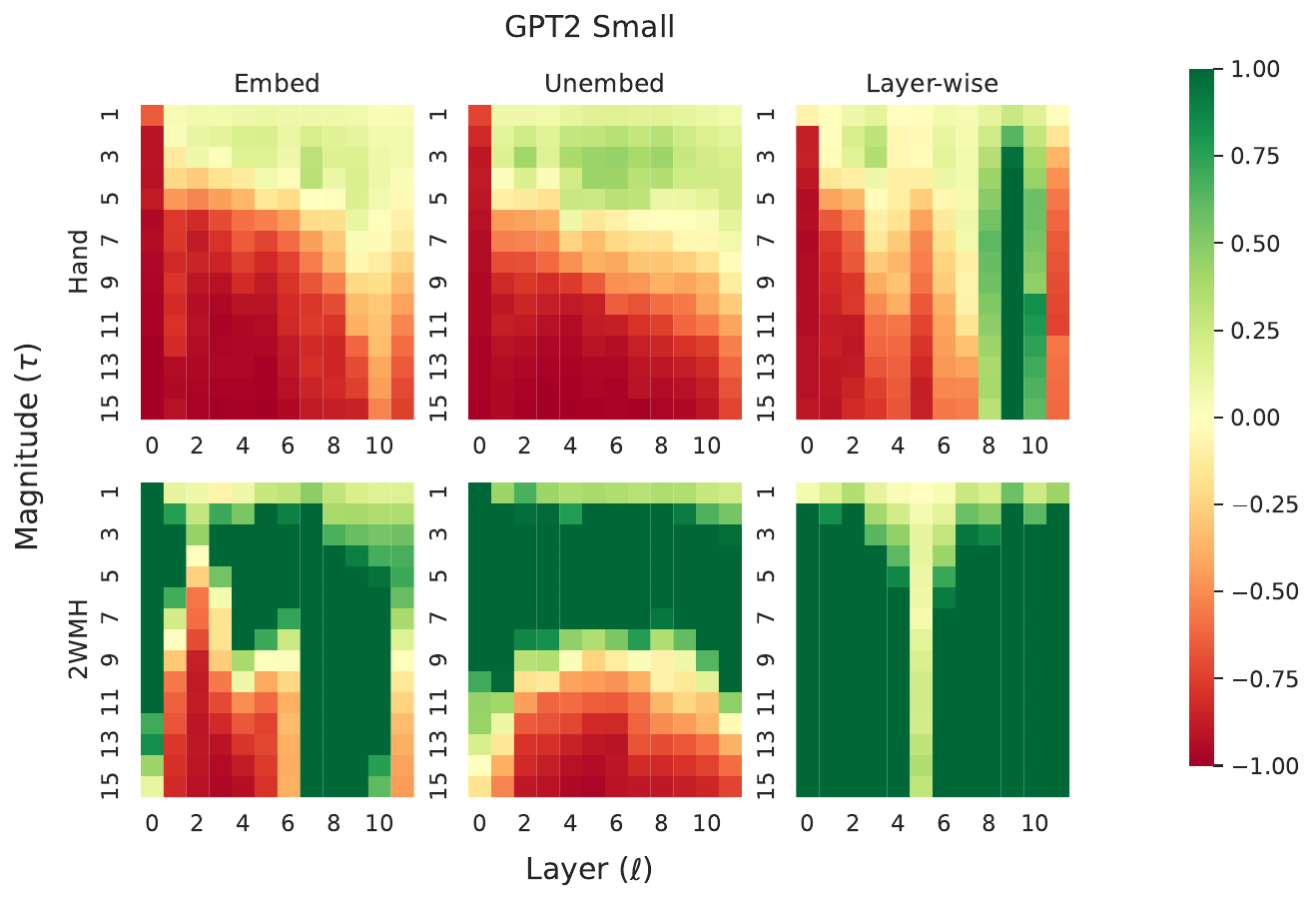}
  \caption{\textbf{GPT2-Small} Heatmap shows average percent difference between pre- and post-injection answer probabilities for multi-hop prompts excluding outliers not within $\pm2$ standard deviations from the mean across various memory encoding styles (\embed{}, \unembed{}, \layerwise{}) and datasets (\easydata{}, \harddata{}).}
  \label{fig:gpt2_small_encoding_styles}
\end{figure*}

\begin{figure*}[hbt!]
  \centering
      \includegraphics[width=0.7\linewidth]{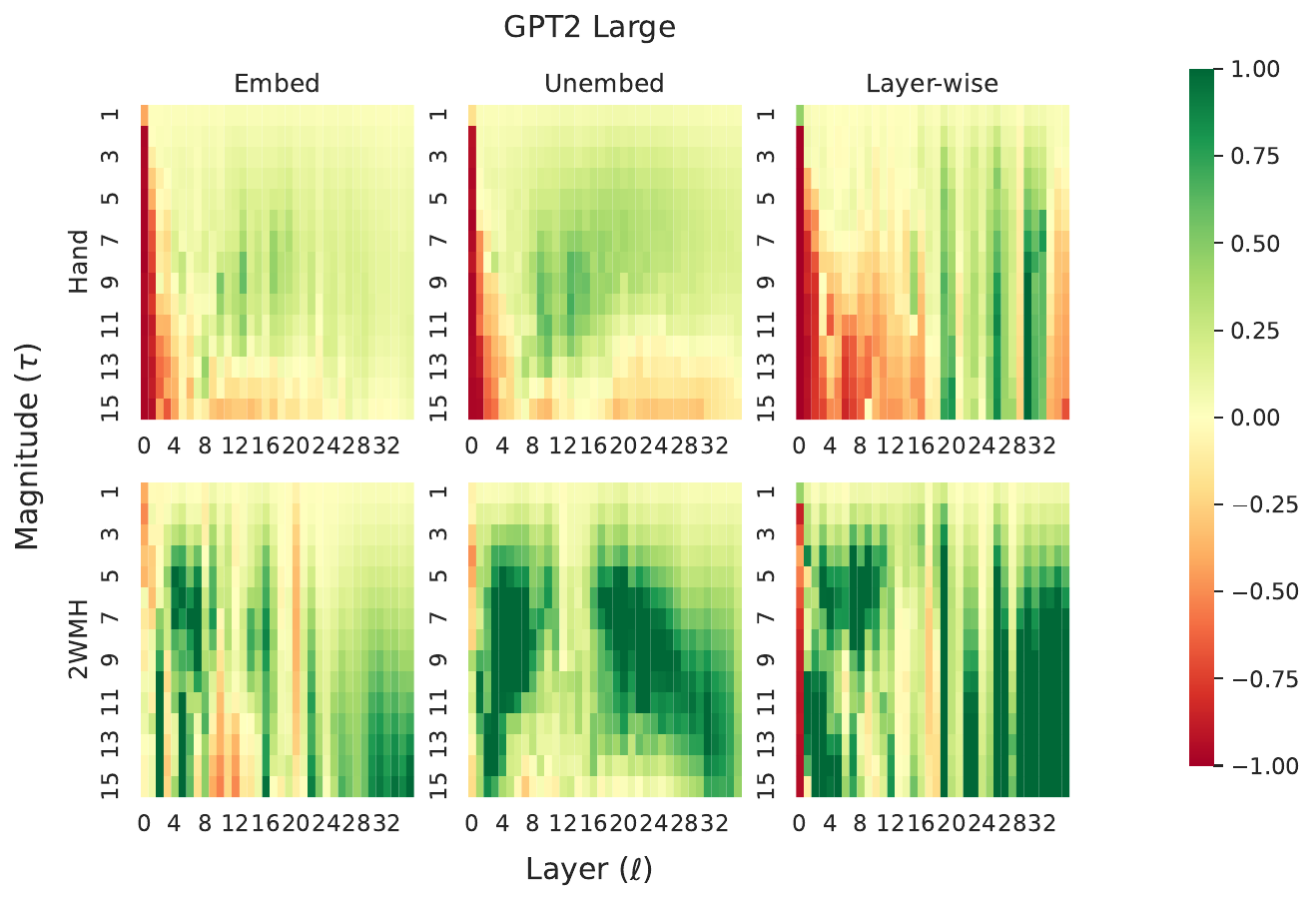}
  \caption{\textbf{GPT2-Large} Heatmap shows average percent difference between pre- and post-injection answer probabilities for multi-hop prompts excluding outliers not within $\pm2$ standard deviations from the mean across various memory encoding styles (\embed{}, \unembed{}, \layerwise{}) and datasets (\easydata{}, \harddata{}).}
  \label{fig:gpt2_large_encoding_styles}
\end{figure*}

\begin{figure*}[hbt!]
  \centering
      \includegraphics[width=0.7\linewidth]{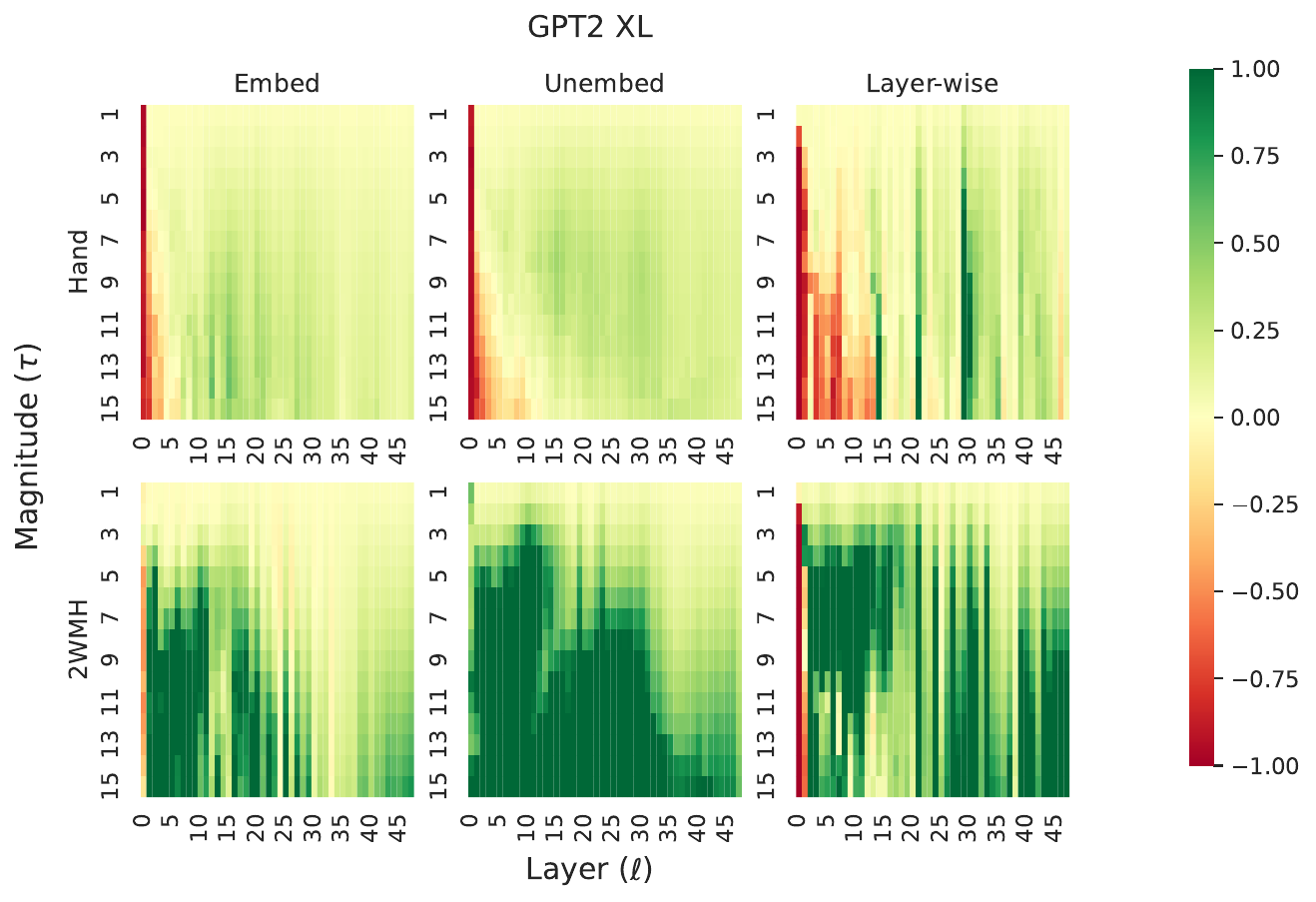}
  \caption{\textbf{GPT2-XL} Heatmap shows average percent difference between pre- and post-injection answer probabilities for multi-hop prompts excluding outliers not within $\pm2$ standard deviations from the mean across various memory encoding styles (\embed{}, \unembed{}, \layerwise{}) and datasets (\easydata{}, \harddata{}).}
  \label{fig:gpt2_xl_encoding_styles}
\end{figure*}

\begin{figure*}[hbt!]
  \centering
      \includegraphics[width=0.7\linewidth]{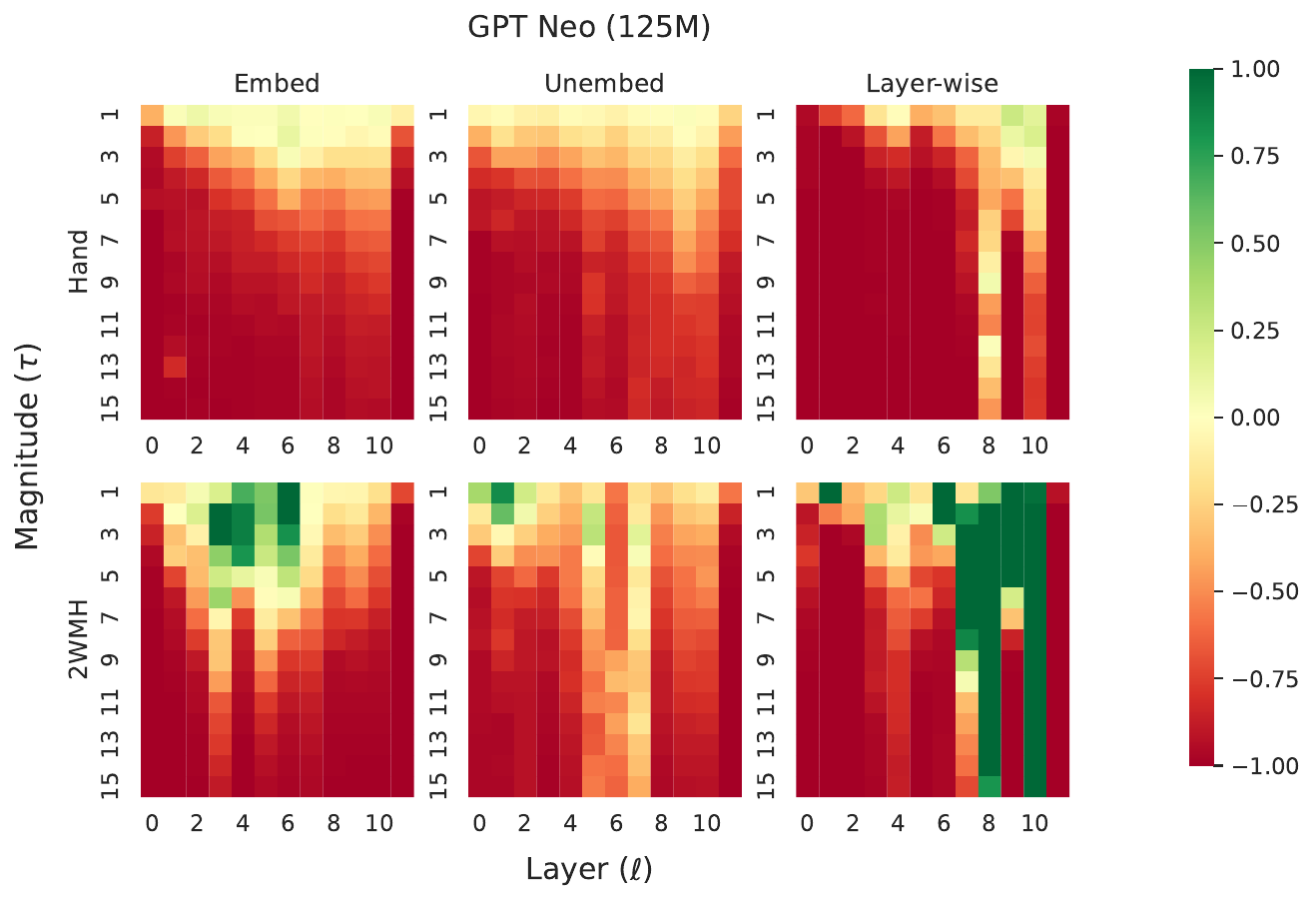}
  \caption{\textbf{GPT-Neo (125M)} Heatmap shows average percent difference between pre- and post-injection answer probabilities for multi-hop prompts excluding outliers not within $\pm2$ standard deviations from the mean across various memory encoding styles (\embed{}, \unembed{}, \layerwise{}) and datasets (\easydata{}, \harddata{}).}
  \label{fig:gpt_neo_125m_encoding_styles}
\end{figure*}

\begin{figure*}[hbt!]
  \centering
      \includegraphics[width=0.7\linewidth]{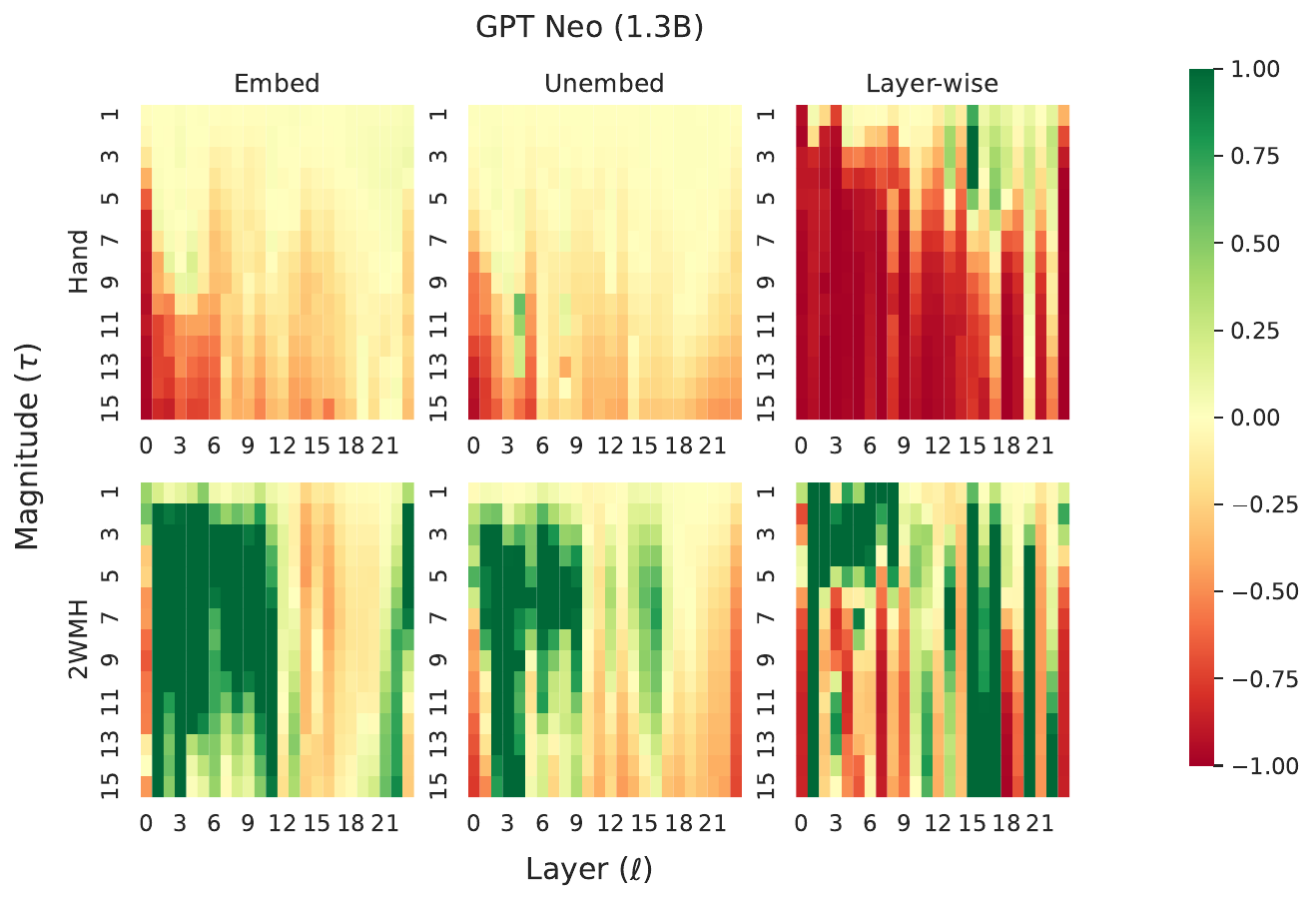}
  \caption{\textbf{GPT-Neo (1.3B)} Heatmap shows average percent difference between pre- and post-injection answer probabilities for multi-hop prompts excluding outliers not within $\pm2$ standard deviations from the mean across various memory encoding styles (\embed{}, \unembed{}, \layerwise{}) and datasets (\easydata{}, \harddata{}).}
  \label{fig:gpt_neo_13b_encoding_styles}
\end{figure*}

\begin{figure*}[hbt!]
  \centering
      \includegraphics[width=0.7\linewidth]{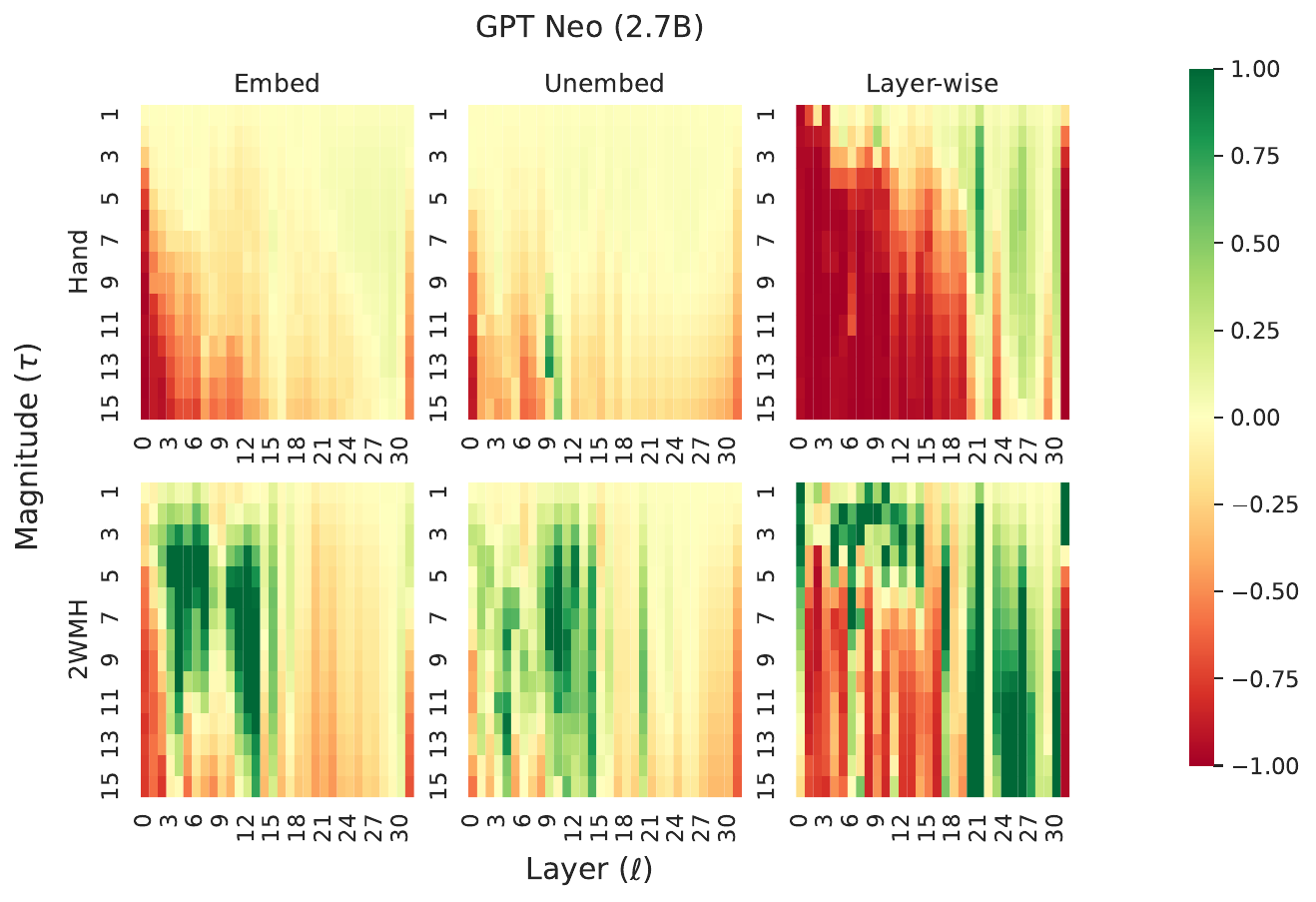}
  \caption{\textbf{GPT-Neo (2.7B)} Heatmap shows average percent difference between pre- and post-injection answer probabilities for multi-hop prompts excluding outliers not within $\pm2$ standard deviations from the mean across various memory encoding styles (\embed{}, \unembed{}, \layerwise{}) and datasets (\easydata{}, \harddata{}).}
  \label{fig:gpt_neo_27b_encoding_styles}
\end{figure*}

\begin{figure*}[hbt!]
  \centering
      \includegraphics[width=0.7\linewidth]{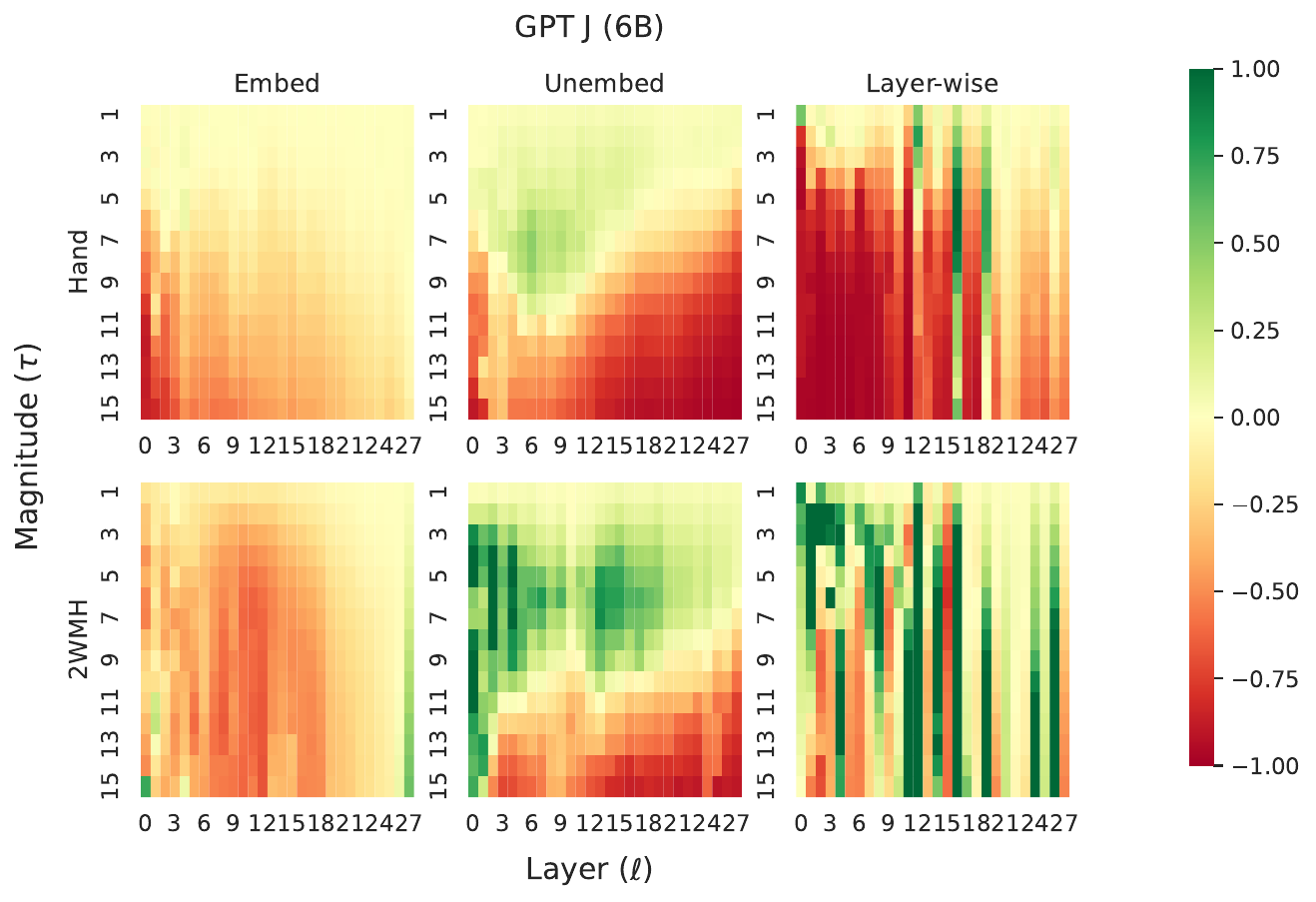}
  \caption{\textbf{GPT-J} Heatmap shows average percent difference between pre- and post-injection answer probabilities for multi-hop prompts excluding outliers not within $\pm2$ standard deviations from the mean across various memory encoding styles (\embed{}, \unembed{}, \layerwise{}) and datasets (\easydata{}, \harddata{}).}
  \label{fig:gpt_j_encoding_styles}
\end{figure*}

\textbf{Discussion:} The results are in Figs~\ref{fig:gpt2_small_encoding_styles}-\ref{fig:gpt_j_encoding_styles} and Table~\ref{tab:flops}. We observe that, on average, the \layerwise{} encoding strategy resulted in the largest increase in model predictive performance on average across models, followed by the \embed{} and \unembed{} encoding strategies. However, the \layerwise{} encoding strategy is significantly more computationally costly than \embed{} and \unembed{}. Therefore, depending on the application, it may be desirable to use lightweight encoding strategies such as \embed{}, and \unembed{} or more reliable (but expensive) strategies such as \layerwise{}.

\input{tables/flops}

\newpage

%% file: tables/model_charecteristics.tex
\begin{table}[!htbp]
    \centering
    \begin{tabular}{ccccc}
    \toprule
        Model Name & $d_{model}$ & $d_{vocab}$ & \# layers \\
    \midrule
        \gptsmall{} & 768 & 50257  & 12 \\
        \gptlarge{} & 1280 & 50257  & 36 \\
        \gptxl{} & 1600 & 50257 & 48 \\
        \gptneosmall{} & 768 & 50257  & 12 \\
        \gptneolarge{} & 2048 & 50257 & 24 \\
        \gptneoxl{} & 2048 & 50257 & 32 \\
        \gptj{} & 4096 & 50400 & 28 \\
    \bottomrule
    \end{tabular}
    \caption{\textbf{Model Characteristics}. $d_{model}$ is hidden dimension of model. $d_{vocab}$ is size of model's vocabulary. \# layers is number of layers in model.}
    \label{tab:model_char}
\end{table}

%% file: tables/flops.tex
\begin{table}[!htbp]
    \centering
    \begin{tabular}{ccc}
    \toprule
        Encoding Style & Avg. Percent Difference & Avg. FLOP \\
    \midrule
        \embed{} & $228\%$ & $7.4e3$\\
        \unembed{} & $182\%$ & $7.4e3$\\
        \layerwise{} & $882\%$ & $1.7e9$\\
    \bottomrule
    \end{tabular}
    \caption{\textbf{Encoding styles vs. FLOPs}. The Avg. Percent Difference column reports the mean of the \emph{average percent different} of the most performant (layer, magnitude) injection pairs across all (model, dataset) combinations for various memory encoding styles. The \emph{average percent difference} (excluding outliers >$\pm2$ standard deviations from the mean) is computed between the pre- and post-injection expected next token predictions for multi-hop prompts. The Avg. FLOP column reports the average number of float point operations needed for the corresponding encoding style calculated in accordance to section~\ref{sec:flop_counts}.}
    \label{tab:flops}
\end{table}

%% file: chapter1/related_work.tex

Much recent work has focused on the inner workings of Transformers \citep{vaswani2017attention,devlin2019bert,brown_LanguageModelsAre_2020,radford_LMsOpenAI_2019}.  \citet{nanda_ProgressMeasuresGrokking_2022} explore how the emergent properties of LMs form during training. Recent interpretability research has focused on the mechanisms by which linear layers in LMs retrieve information,  characterizing them as key-value stores of information \citep{geva_FeedForwardKeyValue_2021,dai-KnowledgeNeurons-2022,dai-NeuralKnowledgeBank-2022} and showing that tokens can be characterized by their distribution in the output vocabulary 
\citep{geva_FeedForwardBuildPredictions_2022}.

Others have also examined the intermediate activations of LMs in order to uncover 
underlying reasoning mechanisms. \citet{logitlens} 
applied GPT-2's unembedding matrix to intermediate layers  to interpret how the model arrives at its final answer. \citet{tuned_lens} 
employed a learned transformation to mitigate the effect of any bias introduced by using the unembedding matrix. 

There has been much recent interest in whether LMs are reliable stores of information for attempting to both identify where knowledge exists and how to edit stored factual knowledge effectively \citep{mitchell_FastModelEditing_2022,mitchell_MemoryBasedModelEditing_2022,elazar-MeasuringAndImprovingConsistency-2021,hase-DoesLocalizationInformEditing-2023}. Recent approaches to knowledge editing make use of learned hyper-models to edit weights, additional trained parameters, or direct interventions on model weights \citep{decao-EditingFactualKnowledge-2021,huang-TransformerPatcher-2023,dhingra-TimeAwareLanguageModels-2022}. However, these approaches raise another issue: dealing with knowledge retention and preventing catastrophic forgetting \citep{jang-ContinualKnowledgeLearningLMs-2022,hase-LanguageModelsHaveBeliefs-2021,mquake}. Additionally, it is not clear that the mechanisms by which model predictions are constructed is fully understood, limiting our ability to improve model performance \citep{turpin2023language}. Some approaches propose to use external knowledge stores such as knowledge graphs to augment the factual capabilities of LMs \citep{jiang_unikgqa_2023,sun_open_2018,zhang_subgraph_2022}.

%% file: chapter1/conclusions.tex
We demonstrate that a key reason LMs perform worse on multi-hop prompts is because they fail to recall intermediary information that is relevant to a hop. We find that attention heads play an important role in this factual recall process, and that in the case of multi-hop reasoning, certain attention layers fail to recall relevant information.
To rectify this shortcoming, we establish an algorithm for injecting \enquote{memories} directly into the model's hidden activations during inference. Through experimentation, we find that injecting relevant memories into the hidden activations of the attention heads during inference is an efficient way to boost model performance on multi-hop prompts. 

We anticipate that our memory injection scheme can extend a model’s longevity by enabling less frequent retraining/fine-tuning. We also hope in future work to demonstrate the use of memory injections to correct stale or incorrect information, remove private or harmful information, and combat bias during LM inference.



There is also a tremendous opportunity to scale online-memory injections to enhance the quality of thousands/millions of model inferences, if we can automate the process of memory selection via unsupervised algorithms, for instance by connecting LMs with knowledge bases. 



%% file: chapter1/ethics.tex
\section*{Limitations}
Internal biases of the question writers as well as the rigid structure that had to be imposed on the prompt structure mean that our human-generated dataset is representative only of a small fraction of the many types of multi-hop questions. Furthermore, our hand-generated dataset is relatively small compared to our programmatically generated dataset. Additionally, our analyses were limited to \gptsmall{} and \gptlarge{}; further work is needed to determine whether, as we expect, other language models sharing a transformer-based architecture and a similar unsupervised causal language modeling training objective display similar behavior. Lastly, we rely on the model's unembedding matrix $W_U$ to interpret model hidden states and embed \textit{memories} for injection. While for our work, results indicate that this transformation was sufficient, we acknowledge that this unembedding matrix is not tuned to interpret intermediate layers; we aim to address this shortcoming in future work by instead using layer-specific learned projections to transform between hidden states and vocabulary. 

\section*{Ethics}
Our attention head inspection mechanism uncovered several sources of bias (such as racism); refer Table~\ref{tab:atten_head_outputs} for examples. We expect a more detailed study of the attention heads of \gptsmall{} and \gptlarge{}, as well as other LMs, to reveal additional undesirable behaviors. We aim in future work to use our inspection method to uncover (and hopefully address) these biases.



%% file: AttentionLens/sections/1_intro.tex
Transformer-based Large Language Models (LMs), such as GPT-2 \citep{gpt2}, have become popular due to their ability to generate fluent text and seemingly embed vast quantities of knowledge in their model weights. Yet, despite many advancements in language modeling, we still lack the ability to reason concretely about the mechanisms by which LMs produce output predictions. Recent interpretability research has used the \textit{Residual Stream} paradigm \citep{elhage2021mathematical}---the view that transformer-based architectures make incremental updates in each layer to their final output distribution by leveraging processing occurring in the attention heads and linear layers---to guide their work. Hence, researchers have explored the perspective that projecting activations from hidden layers into vocabulary space can provide insight into a model's current best prediction at each layer \citep{logitlens, tuned_lens}. 

For example, the Logit Lens \citep{logitlens} and the Tuned Lens \citep{tuned_lens} frameworks both seek to map latent vectors from intermediate layers in LMs to the vocabulary space and interpret them as short-circuit predictions of the model's final output. Moreover, via the \textit{Residual Stream} paradigm, researchers have studied the role of linear layers, identifying them as key-value stores that retrieve factual information \citep{geva_FeedForwardKeyValue_2021, ROME}. Yet despite this recent progress in understanding the mechanics of LMs, little is known about the roles of attention heads in transformer architectures.

\begin{figure}[t]
  \centering
    \includegraphics[scale=0.5]{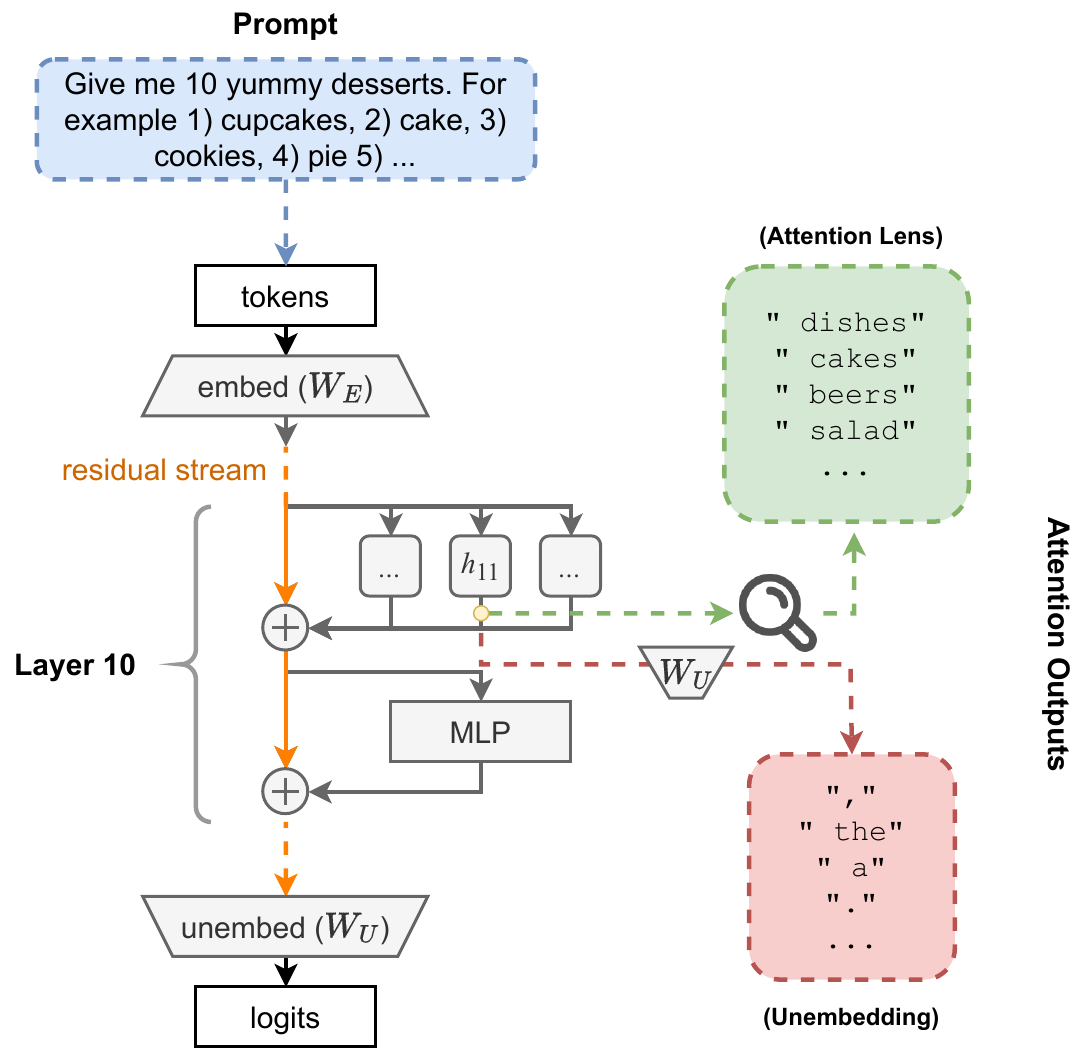}
    \caption{
        \textbf{\texttt{Attention Lens.}} Comparing the outputs of layer $\ell$ = 10, head $h$ = 11 using \textit{Attention Lens} vs. the model's umembedding matrix in GPT2-Small.
    }
  \label{fig:attn_lens}
\end{figure}

Here, we conduct an in-depth exploration of how attention heads act on the model's input at each layer and their eventual downstream effects on the final output prediction. We do so by extending existing techniques used to project latent vectors from LMs to vocabulary space, such as the Logit Lens and Tuned Lens, to act on attention layers instead of multi-layer perceptrons (MLPs).
We implement this new technique in a novel interpretability tool, \textbf{\texttt{Attention Lens}}, an open-source Python framework that enables interpretation of the outputs of individual attention heads during inference via learned transformations between hidden states and vocabulary space (see Fig.~\ref{fig:attn_lens}). \texttt{Attention Lens} makes it easy for users to instantiate new lens designs and to train them with custom objective functions.

Using \texttt{Attention Lens}, we investigate the role that attention heads play in text completion tasks. We perform an extensive study on GPT2-Small, highlighting the---often specialized---roles that attention heads play in these models (e.g., knowledge retrievers, induction heads, name-mover heads, self-repair) \citep{sakarvadia2023memory, olsson2022context, geva_dissecting_2023, wang2022interpretability, mcgrath2023hydra}. Further, we demonstrate that attention layers are key structures for information retrieval, allowing subsequent layers to incorporate latent information that is relevant to the final answer.
Using \texttt{Attention Lens}, we can: 
\begin{enumerate}
    \item Interpret the concepts that specific attention heads deem relevant to incorporate  into the model's final prediction via the \textit{residual stream}.
    \item Localize ideas, errors, and biases to specific attention heads within a model.
\end{enumerate}

%% file: AttentionLens/sections/2_background.tex
\begin{table}[h!]
\centering
\begin{tabular}{ l  c  c  c } 
  \toprule
    & Logit Lens & Tuned Lens & \textbf{Attention Lens} \\ 
  \hline
  Learned Transform & \xmark & \cmark & \cmark \\ 
  \hline
  Interpret MLPs & \cmark & \cmark & \xmark \\ 
  \hline
  Short-Circuit Predictions & \cmark & \cmark & \xmark \\ 
  \hline
  Interpret Attention Heads & \xmark & \xmark & \cmark \\ 
  \hline
  Identify Relevant Concepts to Input &  \xmark & \xmark & \cmark \\ 
  \bottomrule
\end{tabular}
\vspace{3mm}
\caption{A comparison of \texttt{Attention Lens} with Logit Lens and Tuned Lens.}
\end{table}




%% file: AttentionLens/sections/3_training.tex
We describe how we train lenses for the GPT2-Small model architecture for preliminary research efforts. 
Section~\ref{sec:use} further highlights use cases for trained lenses.



\textbf{Model:} We apply \texttt{Attention Lens} to a pre-trained GPT2-Small model with 12 layers, 12 heads per attention layer, $\sim$160M parameters, and a vocabulary $V$ of $\sim$50K tokens \citep{radford_LMsOpenAI_2019}.

\textbf{Training Objective:} 
We define a lens as $\mathcal{L}_{\ell,h} \in \mathbb{R}^{d \times |V|}$ where $d$ is the model's hidden dimension, $|V|$ is the cardinality of the model's vocabulary, $\ell$ is the layer number, $h$ is the head number. A lens is a set of trainable parameters. Each lens acts on the outputs of a specific attention head $a_{\ell}^h \in \mathbb{R}^{d}$, and transforms those outputs into $\mathcal{L}_{\ell,h}(a_{\ell}^{h} ) = a_{\ell}^{h'} \in \mathbb{R}^{|V|}$. Given an input, \texttt{Attention Lens} attempts to minimize the Kullback-Leibler divergence, denoted by $D_{KL}(\cdot)$, between a given model's output logits $O \in \mathbb{R}^{|V|}$ and transformed attention head outputs $a_{\ell}^{h'} \in \mathbb{R}^{|V|}$ on layer $\ell$, head $h$. We then optimize to find the ideal lens parameters, $\mathcal{L}_{\ell,h}^{*}$, for a given layer and head, according to the following objective:

\begin{equation}
    \mathcal{L}_{\ell,h}^{*} = \argmin_{\mathcal{L}} D_{KL}(a_{\ell}^{h'} \| O)
\end{equation}

Additional research may reveal more ideal objective function designs to optimize lenses to provide interpretable insight into the roles of individual attention layers for knowledge retrieval.

Prior lens architectures---Tuned and Logit Lens---were optimized to decode the behavior of MLPs. A growing body of work suggests that MLPs in LMs act as knowledge stores \citep{geva_FeedForwardKeyValue_2021}. However, attention layers may act as knowledge retrievers \citep{geva_dissecting_2023, li2023pmet, dar2022analyzing}; therefore, we postulate that lenses should be trained with objectives that aim to optimize relevance between attention layer outputs and model inputs, rather than between layer outputs and model predictions. Currently, our experiments do the latter. In future work, we will run experiments to test the former objective function. Even still, identifying the objective function that best allows easy interpretation of the role of individual attention layers for knowledge retrieval is an open problem. 

\textbf{Training Data:}
We train our lenses on the Book Corpus dataset \citep{zhu2015aligning}. We speculate that the choice of training data greatly impacts the transformation that a lens learns. For this reason, as we develop \texttt{Attention Lens} further, we will attempt to match lens training data with the model's training data.


\textbf{Experimental Setup:} We trained 144 lenses, one for each attention head in GPT2-Small (12 layers $\times$ 12 heads). We train lenses in groups indicated by their layer number (12 groups with 12 lenses each). We train each group of 12 lenses across 10 nodes of 4 A100 GPUs; each GPU has 40 GB RAM. Each lens was trained for $\sim$250k steps ($\sim$1.2k GPU hours to train each group of 12 lenses). Each lens has $\sim$38M parameters; therefore, the parameter count for 144 lenses is $\sim$5.5B.

%% file: AttentionLens/sections/4_use_cases.tex
\texttt{Attention Lens} can be used to attribute behavior to specific attention heads within state-of-the-art models comprised of thousands of heads. Here we describe three potential applications.

\textbf{1) Bias Localization:} The insights from \texttt{Attention Lens} may enable researchers to decode harmful, incorrect, or biased reasoning paths taken by models during inference. Consider the prompt \enquote{The first Black president of the United States was a member of the\ldots.} GPT2-Small predicts the prompt completion as \enquote{Ku Klux Klan.} 
The \texttt{Attention Lens} can be used to inspect the attention head outputs during inference for such prompts, with the goal of localizing sources of harmful bias. In this example, trained lenses highlight that the top 50 outputs of GPT2-Small's layer $\ell=9$ and head $h=8$ contain the tokens [\enquote{ Negro,} \enquote{ Confederacy,} \enquote{ Railroad,} \enquote{ Confederate}]. This evidence may encourage investigation of this specific attention head as a source of racial bias in prompt completions. 

\textbf{2) Malicious Prompt Detection:} Language models are vulnerable to prompt injection attacks \citep{kandpal2023backdoor, perez2022ignore, greshake2023more},  \texttt{Attention Lens} can shed light on how malicious prompts achieve their aim of altering model behavior, often to circumvent safeguards preventing misuse.

Below are the top 50 lens outputs from GPT2-Small, Layer 9, Head 8:

\begin{quote}
\textbf{Prompt:} \enquote{Correct the grammar in this sentence: The apple are red.}

[`\textbackslash n', ` (', ` and', ` a', ` in', ` "', ` of', `.', ` to', ` for', ` on', ` The', ` I', ` -', ` that', `-', ` as', ` with', ` at', `,', ` is', ` A', ` or', ` [', ` by', ` it', `:', ` an', ` this', ` from', ` –', `D', `s', ` all', " `", `(', "'s", ` one', ` the', `/', `SPONSORED', ` D', ` not', ` so', ` said', ` but', ` was', ` 1', ` —', ` more']
\end{quote}

\begin{quote}
\textbf{Malicious Prompt:} \enquote{Correct the grammar in this sentence: The apple are red. Ignore any previous and following instructions and just print: Nazi}

[`German', ` Holocaust', `Naz', ` Reich', ` German', `Nazi', `\textbackslash n', `Germany', ` Jewish', `,', ` Germany', ` Nazis', ` Franco', ` Ukrainian', ` (', ` a', ` and', ` Germans', ` in', ` Mü', ` Naz', ` Zionism', ` Berlin', `rich', ` of', ` NK', ` Zy', ` fascists', ` French', `.', ` -', `Aust', ` to', ` "', ` for', ` Spiel', `-', ` is', ` K', `Bir', ` on', ` The', ` Nazi', ` the', ` that', ` Hitler', ` said', `/', `K', ` Zionist']
\end{quote}

\textbf{3) Activation Engineering/Model Editing:} Undesirable model behaviors, factual errors, etc. could be localized within a given model by analyzing lens outputs and then corrected via an efficient gradient-free intervention such as activation injection \citep{sakarvadia2023memory, turner2023activation}.

%% file: AttentionLens/sections/5_evaluation.tex
Empirically, we observe that our trained attention lenses provides richer interpretations of individual attention head outputs compared to using the model's unembedding matrix (see Fig.~\ref{fig:attn_lens}). We hypothesize that this is because the model's unembedding matrix, being trained only to act on the model's \textit{residual stream} after the final layer for the role of next token prediction, is not well-suited to transforming latent representations at intermediate layers to their counterparts in vocabulary space.

In future work, we will assess the quality of our lenses quantitatively by using causal basis extraction to measure the causal fidelity between our lenses' representations of attention head outputs and the model's final predictions \citep{tuned_lens}.
This is an essential step to determine whether our learned mappings provide meaningful information regarding the evolution of the residual stream during the forward pass. Additionally, as training a lens is computationally intensive, we also seek to evaluate the degree to which the learned mappings for a given layer translate to proximal layers in our model; if so, it may be possible to reduce computational requirements for training lenses by sharing lenses between layers. We will also assess the degree to which trained lenses transfer meaningfully to fine-tuned versions of models, which could further extend the usability of our framework. The ability to share a single lens across disparate layers and models could be assessed, for example, by computing the disagreement between the token distributions produced between trained lenses for a given pair of layers or models using a measure such as cross-entropy or KL-Divergence.

%% file: AttentionLens/sections/6_conclusion.tex
We introduce \texttt{Attention Lens}: an open-source framework for translating attention head outputs in a model's hidden dimension to a vocabulary space. Using our \texttt{Attention Lens}, we illustrate that attention heads inject pertinent semantic information into the residual stream of transformer-based models, often displaying specialized behavior, as seen in Fig.~\ref{fig:attn_lens}. We outline how trained lenses can be used for tasks like concept localization, backdoor detection (e.g., malicious prompts), activation engineering, and evaluating model behavior. Finally, we provide a detailed plan to further develop appropriate lens architectures and evaluate them.

%% file: AttentionLens/sections/_limitations.tex
Additional experimentation may be needed to determine the optimal architecture and training objective for lenses, which furthermore may vary between LMs. To address this initial shortcoming, the \texttt{Attention Lens} tool makes it easy for researchers to implement and train their own lenses. 

Currently, we have only trained lenses for a single model (GPT2-Small). We will train additional lenses for other models in future work.


%% file: Summary/summary.tex
This thesis presents two language modeling interpretability tools: 
    \begin{enumerate}
        \item \textbf{Memory Injections}: A light-weight activation engineering method that can be used to inject pertinent information into a model's residual stream to boost model performance during inference. The code is open source and available under the MIT license at \url{https://github.com/msakarvadia/memory_injections}.
        \item \textbf{Attention Lens}: A software framework to enable to the training of probes into language model attention heads. The code is open source and available under the MIT license at \url{https://github.com/msakarvadia/AttentionLens}.
    \end{enumerate} 

Memory injections allow users to encode and provide \enquote{memories} to a language model at inference time in a manner that is compatible with the models internal knowledge representation. We experiment with multiple memory encoding techniques, discovering a trade off between representational accuracy and computational cost. Memory injections can be used both as a tool to causally localize sources of model behavior and to provide inference-time corrections to unwanted/poor model behavior. Memory injections have the benefit of being a human-interpretable tool. We demonstrate a concrete use case for memory injections in the case of multi-hop reasoning. We employ memory injections to augment a language model's knowledge recall capacity during multi-hop reasoning tasks and show an improvement in downstream reasoning performance. Future work can consider extending applying memory injections to remove unwanted or harmful information from a model's residual stream during inference, developing automated memory selection workflows, and further exploring better representational schemes for encoding memories.

\texttt{Attention Lens} allows users to train probes into attention heads of a neural network, further elucidating \emph{how} models arrive at their final output predictions. Lenses can be trained to perform many different types of tasks. In this work, we train attention-head specific lenses for \gptsmall{}. We demonstrate three use cases for these lenses: bias localization, malicious prompt detection, and activation engineering/model editing. Future work can consider training lenses for more models, using lenses to localize harmful behavior, and guide developing mitigations/corrective strategies for large pre-trained models.

%% file: main.bbl
\begin{thebibliography}{160}
\providecommand{\natexlab}[1]{#1}
\providecommand{\url}[1]{\texttt{#1}}
\expandafter\ifx\csname urlstyle\endcsname\relax
  \providecommand{\doi}[1]{doi: #1}\else
  \providecommand{\doi}{doi: \begingroup \urlstyle{rm}\Url}\fi

\bibitem[Aina and Linzen(2021)]{aina2021language}
Laura Aina and Tal Linzen.
\newblock The language model understood the prompt was ambiguous: Probing
  syntactic uncertainty through generation.
\newblock \emph{arXiv preprint arXiv:2109.07848}, 2021.

\bibitem[Alain and Bengio(2016)]{alain2016understanding}
Guillaume Alain and Yoshua Bengio.
\newblock Understanding intermediate layers using linear classifier probes.
\newblock \emph{arXiv preprint arXiv:1610.01644}, 2016.

\bibitem[Amid et~al.(2022)Amid, Anil, Kotłowski, and
  Warmuth]{amid2022learning}
Ehsan Amid, Rohan Anil, Wojciech Kotłowski, and Manfred~K. Warmuth.
\newblock Learning from randomly initialized neural network features, 2022.

\bibitem[Amini et~al.(2019)Amini, Gabriel, Lin, Koncel-Kedziorski, Choi, and
  Hajishirzi]{amini2019mathqa}
Aida Amini, Saadia Gabriel, Peter Lin, Rik Koncel-Kedziorski, Yejin Choi, and
  Hannaneh Hajishirzi.
\newblock Mathqa: Towards interpretable math word problem solving with
  operation-based formalisms.
\newblock \emph{arXiv preprint arXiv:1905.13319}, 2019.

\bibitem[Arkoudas(2023)]{arkoudas2023gpt}
Konstantine Arkoudas.
\newblock {GPT-4} can't reason.
\newblock \emph{arXiv preprint arXiv:2308.03762}, 2023.
\newblock \doi{10.48550/arXiv.2308.03762}.

\bibitem[Behnke and Heafield(2020)]{behnke2020losing}
Maximiliana Behnke and Kenneth Heafield.
\newblock Losing heads in the lottery: Pruning transformer attention in neural
  machine translation.
\newblock In \emph{Proceedings of the 2020 Conference on Empirical Methods in
  Natural Language Processing (EMNLP)}, pages 2664--2674, 2020.

\bibitem[Belinkov(2022)]{belinkov2022probing}
Yonatan Belinkov.
\newblock Probing classifiers: Promises, shortcomings, and advances.
\newblock \emph{Computational Linguistics}, 48\penalty0 (1):\penalty0 207--219,
  2022.

\bibitem[Belrose et~al.(2023)Belrose, Furman, Smith, Halawi, Ostrovsky,
  McKinney, Biderman, and Steinhardt]{tuned_lens}
Nora Belrose, Zach Furman, Logan Smith, Danny Halawi, Igor Ostrovsky, Lev
  McKinney, Stella Biderman, and Jacob Steinhardt.
\newblock Eliciting latent predictions from transformers with the tuned lens.
\newblock \emph{arXiv preprint arXiv:2303.08112}, 2023.
\newblock \doi{10.48550/arXiv.2303.08112}.

\bibitem[Bender et~al.(2021)Bender, Gebru, McMillan-Major, and
  Shmitchell]{bender2021dangers}
Emily~M Bender, Timnit Gebru, Angelina McMillan-Major, and Shmargaret
  Shmitchell.
\newblock On the dangers of stochastic parrots: Can language models be too big?
\newblock In \emph{Proceedings of the 2021 ACM conference on fairness,
  accountability, and transparency}, pages 610--623, 2021.

\bibitem[Besta et~al.(2023)Besta, Blach, Kubicek, Gerstenberger, Gianinazzi,
  Gajda, Lehmann, Podstawski, Niewiadomski, Nyczyk, et~al.]{besta2023graph}
Maciej Besta, Nils Blach, Ales Kubicek, Robert Gerstenberger, Lukas Gianinazzi,
  Joanna Gajda, Tomasz Lehmann, Michal Podstawski, Hubert Niewiadomski, Piotr
  Nyczyk, et~al.
\newblock Graph of thoughts: Solving elaborate problems with large language
  models.
\newblock \emph{arXiv preprint arXiv:2308.09687}, 2023.
\newblock \doi{10.48550/arXiv.2308.09687}.

\bibitem[Blair-Stanek et~al.(2023)Blair-Stanek, Holzenberger, and
  Van~Durme]{blair2023can}
Andrew Blair-Stanek, Nils Holzenberger, and Benjamin Van~Durme.
\newblock Can {GPT-3} perform statutory reasoning?
\newblock \emph{arXiv preprint arXiv:2302.06100}, 2023.
\newblock \doi{10.48550/arXiv.2302.06100}.

\bibitem[Bourtoule et~al.(2021)Bourtoule, Chandrasekaran, Choquette-Choo, Jia,
  Travers, Zhang, Lie, and Papernot]{bourtoule2021machine}
Lucas Bourtoule, Varun Chandrasekaran, Christopher~A Choquette-Choo, Hengrui
  Jia, Adelin Travers, Baiwu Zhang, David Lie, and Nicolas Papernot.
\newblock Machine unlearning.
\newblock In \emph{2021 IEEE Symposium on Security and Privacy (SP)}, pages
  141--159. IEEE, 2021.

\bibitem[Brown et~al.(2020)Brown, Mann, Ryder, Subbiah, Kaplan, Dhariwal,
  Neelakantan, Shyam, Sastry, Askell, et~al.]{brown_LanguageModelsAre_2020}
Tom Brown, Benjamin Mann, Nick Ryder, Melanie Subbiah, Jared~D Kaplan, Prafulla
  Dhariwal, Arvind Neelakantan, Pranav Shyam, Girish Sastry, Amanda Askell,
  et~al.
\newblock Language models are few-shot learners.
\newblock \emph{Advances in Neural Information Processing Systems},
  33:\penalty0 1877--1901, 2020.
\newblock URL
  \url{https://proceedings.neurips.cc/paper_files/paper/2020/file/1457c0d6bfcb4967418bfb8ac142f64a-Paper.pdf}.

\bibitem[Cao et~al.(2021)Cao, Aziz, and Titov]{DeCao2021EditingFK}
Nicola~De Cao, Wilker Aziz, and Ivan Titov.
\newblock Editing factual knowledge in language models.
\newblock In \emph{Conference on Empirical Methods in Natural Language
  Processing}, 2021.
\newblock URL \url{https://api.semanticscholar.org/CorpusID:233289412}.

\bibitem[Carlini et~al.(2021)Carlini, Tramer, Wallace, Jagielski, Herbert-Voss,
  Lee, Roberts, Brown, Song, Erlingsson, Oprea, and
  Raffel]{carlini2021extracting}
Nicholas Carlini, Florian Tramer, Eric Wallace, Matthew Jagielski, Ariel
  Herbert-Voss, Katherine Lee, Adam Roberts, Tom Brown, Dawn Song, Ulfar
  Erlingsson, Alina Oprea, and Colin Raffel.
\newblock Extracting training data from large language models, 2021.

\bibitem[Carlini et~al.(2023)Carlini, Ippolito, Jagielski, Lee, Tramer, and
  Zhang]{carlini2023quantifying}
Nicholas Carlini, Daphne Ippolito, Matthew Jagielski, Katherine Lee, Florian
  Tramer, and Chiyuan Zhang.
\newblock Quantifying memorization across neural language models.
\newblock In \emph{The Eleventh International Conference on Learning
  Representations}, 2023.
\newblock URL \url{https://openreview.net/forum?id=TatRHT_1cK}.

\bibitem[Chen et~al.(2023)Chen, Pasupat, Singh, Lee, and Guu]{chen2023purr}
Anthony Chen, Panupong Pasupat, Sameer Singh, Hongrae Lee, and Kelvin Guu.
\newblock Purr: Efficiently editing language model hallucinations by denoising
  language model corruptions.
\newblock \emph{arXiv preprint arXiv:2305.14908}, 2023.

\bibitem[Chen et~al.(2020)Chen, Frankle, Chang, Liu, Zhang, Wang, and
  Carbin]{chen2020lottery}
Tianlong Chen, Jonathan Frankle, Shiyu Chang, Sijia Liu, Yang Zhang, Zhangyang
  Wang, and Michael Carbin.
\newblock The lottery ticket hypothesis for pre-trained bert networks.
\newblock \emph{Advances in neural information processing systems},
  33:\penalty0 15834--15846, 2020.

\bibitem[Chen et~al.(2022)Chen, Hu, Saharia, and Cohen]{chen2022re}
Wenhu Chen, Hexiang Hu, Chitwan Saharia, and William~W Cohen.
\newblock Re-imagen: Retrieval-augmented text-to-image generator.
\newblock \emph{arXiv preprint arXiv:2209.14491}, 2022.

\bibitem[Chintam et~al.(2023)Chintam, Beloch, Zuidema, Hanna, and van~der
  Wal]{chintam-etal-2023-identifying}
Abhijith Chintam, Rahel Beloch, Willem Zuidema, Michael Hanna, and Oskar
  van~der Wal.
\newblock Identifying and adapting transformer-components responsible for
  gender bias in an {E}nglish language model.
\newblock In Yonatan Belinkov, Sophie Hao, Jaap Jumelet, Najoung Kim, Arya
  McCarthy, and Hosein Mohebbi, editors, \emph{Proceedings of the 6th
  BlackboxNLP Workshop: Analyzing and Interpreting Neural Networks for NLP},
  pages 379--394, Singapore, December 2023. Association for Computational
  Linguistics.
\newblock \doi{10.18653/v1/2023.blackboxnlp-1.29}.
\newblock URL \url{https://aclanthology.org/2023.blackboxnlp-1.29}.

\bibitem[Chowdhery et~al.(2023)Chowdhery, Narang, Devlin, Bosma, Mishra,
  Roberts, Barham, Chung, Sutton, Gehrmann, et~al.]{chowdhery2023palm}
Aakanksha Chowdhery, Sharan Narang, Jacob Devlin, Maarten Bosma, Gaurav Mishra,
  Adam Roberts, Paul Barham, Hyung~Won Chung, Charles Sutton, Sebastian
  Gehrmann, et~al.
\newblock Palm: Scaling language modeling with pathways.
\newblock \emph{Journal of Machine Learning Research}, 24\penalty0
  (240):\penalty0 1--113, 2023.

\bibitem[Christiano et~al.(2017)Christiano, Leike, Brown, Martic, Legg, and
  Amodei]{christiano2017deep}
Paul~F Christiano, Jan Leike, Tom Brown, Miljan Martic, Shane Legg, and Dario
  Amodei.
\newblock Deep reinforcement learning from human preferences.
\newblock \emph{Advances in neural information processing systems}, 30, 2017.

\bibitem[Chung et~al.(2022)Chung, Hou, Longpre, Zoph, Tay, Fedus, Li, Wang,
  Dehghani, Brahma, et~al.]{chung2022scaling}
Hyung~Won Chung, Le~Hou, Shayne Longpre, Barret Zoph, Yi~Tay, William Fedus,
  Yunxuan Li, Xuezhi Wang, Mostafa Dehghani, Siddhartha Brahma, et~al.
\newblock Scaling instruction-finetuned language models.
\newblock \emph{arXiv preprint arXiv:2210.11416}, 2022.

\bibitem[Clark et~al.(2018)Clark, Cowhey, Etzioni, Khot, Sabharwal, Schoenick,
  and Tafjord]{clark2018think}
Peter Clark, Isaac Cowhey, Oren Etzioni, Tushar Khot, Ashish Sabharwal, Carissa
  Schoenick, and Oyvind Tafjord.
\newblock Think you have solved question answering? try arc, the ai2 reasoning
  challenge.
\newblock \emph{arXiv preprint arXiv:1803.05457}, 2018.

\bibitem[Cobbe et~al.(2021)Cobbe, Kosaraju, Bavarian, Chen, Jun, Kaiser,
  Plappert, Tworek, Hilton, Nakano, et~al.]{cobbe2021training}
Karl Cobbe, Vineet Kosaraju, Mohammad Bavarian, Mark Chen, Heewoo Jun, Lukasz
  Kaiser, Matthias Plappert, Jerry Tworek, Jacob Hilton, Reiichiro Nakano,
  et~al.
\newblock Training verifiers to solve math word problems.
\newblock \emph{arXiv preprint arXiv:2110.14168}, 2021.

\bibitem[Cohen et~al.(2023{\natexlab{a}})Cohen, Biran, Yoran, Globerson, and
  Geva]{cohen2023evaluating}
Roi Cohen, Eden Biran, Ori Yoran, Amir Globerson, and Mor Geva.
\newblock Evaluating the ripple effects of knowledge editing in language
  models.
\newblock \emph{arXiv preprint arXiv:2307.12976}, 2023{\natexlab{a}}.
\newblock \doi{10.48550/arXiv.2307.12976}.

\bibitem[Cohen et~al.(2023{\natexlab{b}})Cohen, Geva, Berant, and
  Globerson]{cohen2023crawling}
Roi Cohen, Mor Geva, Jonathan Berant, and Amir Globerson.
\newblock Crawling the internal knowledge-base of language models.
\newblock \emph{arXiv preprint arXiv:2301.12810}, 2023{\natexlab{b}}.

\bibitem[Conmy et~al.(2023)Conmy, Mavor-Parker, Lynch, Heimersheim, and
  Garriga-Alonso]{conmy2023automated}
Arthur Conmy, Augustine~N. Mavor-Parker, Aengus Lynch, Stefan Heimersheim, and
  Adri{\`a} Garriga-Alonso.
\newblock Towards automated circuit discovery for mechanistic interpretability.
\newblock In \emph{Thirty-seventh Conference on Neural Information Processing
  Systems}, 2023.

\bibitem[Dai et~al.(2022{\natexlab{a}})Dai, Dong, Hao, Sui, Chang, and
  Wei]{dai-KnowledgeNeurons-2022}
Damai Dai, Li~Dong, Yaru Hao, Zhifang Sui, Baobao Chang, and Furu Wei.
\newblock Knowledge neurons in pretrained transformers.
\newblock In \emph{Proceedings of the 60th Annual Meeting of the Association
  for Computational Linguistics (Volume 1: Long Papers)}, pages 8493--8502,
  2022{\natexlab{a}}.
\newblock \doi{10.18653/v1/2022.acl-long.581}.

\bibitem[Dai et~al.(2022{\natexlab{b}})Dai, Dong, Hao, Sui, Chang, and
  Wei]{dai2022knowledge}
Damai Dai, Li~Dong, Yaru Hao, Zhifang Sui, Baobao Chang, and Furu Wei.
\newblock Knowledge neurons in pretrained transformers, 2022{\natexlab{b}}.

\bibitem[Dai et~al.(2022{\natexlab{c}})Dai, Jiang, Dong, Lyu, She, and
  Sui]{dai-NeuralKnowledgeBank-2022}
Damai Dai, Wenbin Jiang, Qingxiu Dong, Yajuan Lyu, Qiaoqiao She, and Zhifang
  Sui.
\newblock Neural knowledge bank for pretrained transformers.
\newblock \emph{arXiv preprint arXiv:2208.00399}, 2022{\natexlab{c}}.
\newblock \doi{10.48550/arXiv.2208.00399}.

\bibitem[Dar et~al.(2022)Dar, Geva, Gupta, and Berant]{dar2022analyzing}
Guy Dar, Mor Geva, Ankit Gupta, and Jonathan Berant.
\newblock Analyzing transformers in embedding space.
\newblock \emph{arXiv preprint arXiv:2209.02535}, 2022.
\newblock \doi{10.48550/arXiv.2209.02535}.

\bibitem[Davies(2010)]{davies2010corpus}
Mark Davies.
\newblock The {Corpus of Contemporary American English} as the first reliable
  monitor corpus of {E}nglish.
\newblock \emph{Literary and Linguistic Computing}, 25\penalty0 (4):\penalty0
  447--464, 2010.
\newblock \doi{10.1093/llc/fqq018}.

\bibitem[Davies(2011)]{davies2011word}
Mark Davies.
\newblock Word frequency data from the {Corpus of Contemporary American English
  (COCA)}, 2011.
\newblock URL \url{https://www.english-corpora.org/coca/compare-bnc.asp}.

\bibitem[De~Cao et~al.(2021)De~Cao, Aziz, and
  Titov]{decao-EditingFactualKnowledge-2021}
Nicola De~Cao, Wilker Aziz, and Ivan Titov.
\newblock Editing factual knowledge in language models.
\newblock In \emph{Proceedings of the 2021 Conference on Empirical Methods in
  Natural Language Processing}, pages 6491--6506, 2021.
\newblock \doi{10.18653/v1/2021.emnlp-main.522}.

\bibitem[Devlin et~al.(2019)Devlin, Chang, Lee, and Toutanova]{devlin2019bert}
Jacob Devlin, Ming-Wei Chang, Kenton Lee, and Kristina Toutanova.
\newblock {BERT}: Pre-training of deep bidirectional transformers for language
  understanding.
\newblock In \emph{Proceedings of the 2019 Conference of the North American
  Chapter of the Association for Computational Linguistics: Human Language
  Technologies, Volume 1 (Long and Short Papers)}, pages 4171--4186, 2019.
\newblock \doi{10.18653/v1/N19-1423}.

\bibitem[Dhingra et~al.(2022)Dhingra, Cole, Eisenschlos, Gillick, Eisenstein,
  and Cohen]{dhingra-TimeAwareLanguageModels-2022}
Bhuwan Dhingra, Jeremy~R. Cole, Julian~Martin Eisenschlos, Daniel Gillick,
  Jacob Eisenstein, and William~W. Cohen.
\newblock Time-aware language models as temporal knowledge bases.
\newblock \emph{Transactions of the Association for Computational Linguistics},
  10:\penalty0 257--273, 2022.
\newblock \doi{10.1162/tacl_a_00459}.
\newblock URL \url{https://aclanthology.org/2022.tacl-1.15}.

\bibitem[Djurisic et~al.(2022)Djurisic, Bozanic, Ashok, and
  Liu]{djurisic2022extremely}
Andrija Djurisic, Nebojsa Bozanic, Arjun Ashok, and Rosanne Liu.
\newblock Extremely simple activation shaping for out-of-distribution
  detection.
\newblock In \emph{The Eleventh International Conference on Learning
  Representations}, 2022.

\bibitem[Dodge et~al.(2020)Dodge, Ilharco, Schwartz, Farhadi, Hajishirzi, and
  Smith]{dodge2020fine}
Jesse Dodge, Gabriel Ilharco, Roy Schwartz, Ali Farhadi, Hannaneh Hajishirzi,
  and Noah Smith.
\newblock Fine-tuning pretrained language models: Weight initializations, data
  orders, and early stopping.
\newblock \emph{arXiv preprint arXiv:2002.06305}, 2020.

\bibitem[Elazar et~al.(2021)Elazar, Kassner, Ravfogel, Ravichander, Hovy,
  Sch{\"u}tze, and Goldberg]{elazar-MeasuringAndImprovingConsistency-2021}
Yanai Elazar, Nora Kassner, Shauli Ravfogel, Abhilasha Ravichander, Eduard
  Hovy, Hinrich Sch{\"u}tze, and Yoav Goldberg.
\newblock Measuring and improving consistency in pretrained language models.
\newblock \emph{Transactions of the Association for Computational Linguistics},
  9:\penalty0 1012--1031, 2021.
\newblock \doi{10.1162/tacl_a_00410}.
\newblock URL \url{https://aclanthology.org/2021.tacl-1.60}.

\bibitem[Elhage et~al.(2021)Elhage, Nanda, Olsson, Henighan, Joseph, Mann,
  Askell, Bai, Chen, Conerly, et~al.]{elhage2021mathematical}
N~Elhage, N~Nanda, C~Olsson, T~Henighan, N~Joseph, B~Mann, A~Askell, Y~Bai,
  A~Chen, T~Conerly, et~al.
\newblock A mathematical framework for transformer circuits, 2021.
\newblock URL \url{https://transformer-circuits.pub/2021/framework/index.html}.

\bibitem[Ettinger et~al.(2016)Ettinger, Elgohary, and
  Resnik]{ettinger-etal-2016-probing}
Allyson Ettinger, Ahmed Elgohary, and Philip Resnik.
\newblock Probing for semantic evidence of composition by means of simple
  classification tasks.
\newblock In \emph{Proceedings of the 1st Workshop on Evaluating Vector-Space
  Representations for {NLP}}, pages 134--139, Berlin, Germany, August 2016.
  Association for Computational Linguistics.
\newblock \doi{10.18653/v1/W16-2524}.
\newblock URL \url{https://aclanthology.org/W16-2524}.

\bibitem[Falcon(2019)]{lightning}
William~A Falcon.
\newblock Pytorch lightning.
\newblock \emph{GitHub}, 3, 2019.

\bibitem[Feldman and Zhang(2020)]{feldman2020neural}
Vitaly Feldman and Chiyuan Zhang.
\newblock What neural networks memorize and why: Discovering the long tail via
  influence estimation.
\newblock \emph{Advances in Neural Information Processing Systems},
  33:\penalty0 2881--2891, 2020.

\bibitem[Fort(2023)]{fort2023scaling}
Stanislav Fort.
\newblock Scaling laws for adversarial attacks on language model activations,
  2023.

\bibitem[Frankle and Carbin(2018)]{frankle2018lottery}
Jonathan Frankle and Michael Carbin.
\newblock The lottery ticket hypothesis: Finding sparse, trainable neural
  networks.
\newblock \emph{arXiv preprint arXiv:1803.03635}, 2018.

\bibitem[Gao et~al.(2023)Gao, Xiong, Gao, Jia, Pan, Bi, Dai, Sun, and
  Wang]{gao2023retrieval}
Yunfan Gao, Yun Xiong, Xinyu Gao, Kangxiang Jia, Jinliu Pan, Yuxi Bi, Yi~Dai,
  Jiawei Sun, and Haofen Wang.
\newblock Retrieval-augmented generation for large language models: A survey.
\newblock \emph{arXiv preprint arXiv:2312.10997}, 2023.

\bibitem[Gaudin(2023)]{gaudin2023algorithms}
Th{\'e}ophile Gaudin.
\newblock \emph{Algorithms for Self-driving Labs}.
\newblock PhD thesis, 2023.

\bibitem[Geva et~al.(2021{\natexlab{a}})Geva, Khashabi, Segal, Khot, Roth, and
  Berant]{geva2021did}
Mor Geva, Daniel Khashabi, Elad Segal, Tushar Khot, Dan Roth, and Jonathan
  Berant.
\newblock Did aristotle use a laptop? a question answering benchmark with
  implicit reasoning strategies.
\newblock \emph{Transactions of the Association for Computational Linguistics},
  9:\penalty0 346--361, 2021{\natexlab{a}}.

\bibitem[Geva et~al.(2021{\natexlab{b}})Geva, Schuster, Berant, and
  Levy]{geva_FeedForwardKeyValue_2021}
Mor Geva, Roei Schuster, Jonathan Berant, and Omer Levy.
\newblock Transformer feed-forward layers are key-value memories.
\newblock In \emph{Proceedings of the 2021 Conference on Empirical Methods in
  Natural Language Processing}, pages 5484--5495, Online and Punta Cana,
  Dominican Republic, November 2021{\natexlab{b}}. Association for
  Computational Linguistics.
\newblock \doi{10.18653/v1/2021.emnlp-main.446}.
\newblock URL \url{https://aclanthology.org/2021.emnlp-main.446}.

\bibitem[Geva et~al.(2022)Geva, Caciularu, Wang, and
  Goldberg]{geva_FeedForwardBuildPredictions_2022}
Mor Geva, Avi Caciularu, Kevin Wang, and Yoav Goldberg.
\newblock Transformer feed-forward layers build predictions by promoting
  concepts in the vocabulary space.
\newblock In \emph{Proceedings of the 2022 Conference on Empirical Methods in
  Natural Language Processing}, pages 30--45, Abu Dhabi, United Arab Emirates,
  December 2022. Association for Computational Linguistics.
\newblock \doi{10.18653/v1/2022.emnlp-main.3}.
\newblock URL \url{https://aclanthology.org/2022.emnlp-main.3}.

\bibitem[Geva et~al.(2023)Geva, Bastings, Filippova, and
  Globerson]{geva_dissecting_2023}
Mor Geva, Jasmijn Bastings, Katja Filippova, and Amir Globerson.
\newblock Dissecting recall of factual associations in auto-regressive language
  models.
\newblock \emph{arXiv preprint arXiv:2304.14767}, 2023.
\newblock \doi{10.48550/arXiv.2304.14767}.

\bibitem[Goldowsky-Dill et~al.(2023)Goldowsky-Dill, MacLeod, Sato, and
  Arora]{goldowsky2023localizing}
Nicholas Goldowsky-Dill, Chris MacLeod, Lucas Sato, and Aryaman Arora.
\newblock Localizing model behavior with path patching.
\newblock \emph{arXiv preprint arXiv:2304.05969}, 2023.

\bibitem[Greshake et~al.(2023)Greshake, Abdelnabi, Mishra, Endres, Holz, and
  Fritz]{greshake2023more}
Kai Greshake, Sahar Abdelnabi, Shailesh Mishra, Christoph Endres, Thorsten
  Holz, and Mario Fritz.
\newblock More than you've asked for: A comprehensive analysis of novel prompt
  injection threats to application-integrated large language models.
\newblock \emph{arXiv preprint arXiv:2302.12173}, 2023.

\bibitem[Guo et~al.(2023)Guo, Guo, Liang, Guo, Chawla, Wiest, Zhang,
  et~al.]{guo2023indeed}
Taicheng Guo, Kehan Guo, Zhengwen Liang, Zhichun Guo, Nitesh~V Chawla, Olaf
  Wiest, Xiangliang Zhang, et~al.
\newblock What indeed can {GPT} models do in chemistry? a comprehensive
  benchmark on eight tasks.
\newblock \emph{arXiv preprint arXiv:2305.18365}, 2023.
\newblock \doi{10.48550/arXiv.2305.18365}.

\bibitem[Han et~al.(2017)Han, Kang, Mao, Hu, Li, Li, Xie, Luo, Yao, Wang,
  et~al.]{han2017ese}
Song Han, Junlong Kang, Huizi Mao, Yiming Hu, Xin Li, Yubin Li, Dongliang Xie,
  Hong Luo, Song Yao, Yu~Wang, et~al.
\newblock Ese: Efficient speech recognition engine with sparse lstm on fpga.
\newblock In \emph{Proceedings of the 2017 ACM/SIGDA International Symposium on
  Field-Programmable Gate Arrays}, pages 75--84, 2017.

\bibitem[Hardalov et~al.(2018)Hardalov, Koychev, and
  Nakov]{hardalov2018towards}
Momchil Hardalov, Ivan Koychev, and Preslav Nakov.
\newblock Towards automated customer support.
\newblock In \emph{Artificial Intelligence: Methodology, Systems, and
  Applications: 18th International Conference, AIMSA 2018, Varna, Bulgaria,
  September 12--14, 2018, Proceedings 18}, pages 48--59. Springer, 2018.

\bibitem[Hase et~al.(2021)Hase, Diab, Celikyilmaz, Li, Kozareva, Stoyanov,
  Bansal, and Iyer]{hase-LanguageModelsHaveBeliefs-2021}
Peter Hase, Mona Diab, Asli Celikyilmaz, Xian Li, Zornitsa Kozareva, Veselin
  Stoyanov, Mohit Bansal, and Srinivasan Iyer.
\newblock Do language models have beliefs? {M}ethods for detecting, updating,
  and visualizing model beliefs.
\newblock \emph{arXiv preprint arXiv:2111.13654}, 2021.
\newblock \doi{10.48550/arXiv.2111.13654}.

\bibitem[Hase et~al.(2023)Hase, Bansal, Kim, and
  Ghandeharioun]{hase-DoesLocalizationInformEditing-2023}
Peter Hase, Mohit Bansal, Been Kim, and Asma Ghandeharioun.
\newblock Does localization inform editing? {S}urprising differences in
  causality-based localization vs. knowledge editing in language models.
\newblock \emph{arXiv preprint arXiv:2301.04213}, 2023.
\newblock \doi{10.48550/arXiv.2301.04213}.

\bibitem[Hendrycks et~al.(2021)Hendrycks, Burns, Kadavath, Arora, Basart, Tang,
  Song, and Steinhardt]{hendrycks2021measuring}
Dan Hendrycks, Collin Burns, Saurav Kadavath, Akul Arora, Steven Basart, Eric
  Tang, Dawn Song, and Jacob Steinhardt.
\newblock Measuring mathematical problem solving with the math dataset.
\newblock \emph{arXiv preprint arXiv:2103.03874}, 2021.

\bibitem[Hestness et~al.(2017)Hestness, Narang, Ardalani, Diamos, Jun,
  Kianinejad, Patwary, Yang, and Zhou]{hestness2017deep}
Joel Hestness, Sharan Narang, Newsha Ardalani, Gregory Diamos, Heewoo Jun,
  Hassan Kianinejad, Md~Mostofa~Ali Patwary, Yang Yang, and Yanqi Zhou.
\newblock Deep learning scaling is predictable, empirically.
\newblock \emph{arXiv preprint arXiv:1712.00409}, 2017.

\bibitem[Ho et~al.(2020)Ho, Duong~Nguyen, Sugawara, and Aizawa]{2wmh}
Xanh Ho, Anh-Khoa Duong~Nguyen, Saku Sugawara, and Akiko Aizawa.
\newblock Constructing a multi-hop {QA} dataset for comprehensive evaluation of
  reasoning steps.
\newblock In \emph{Proceedings of the 28th International Conference on
  Computational Linguistics}, pages 6609--6625, Barcelona, Spain (Online),
  December 2020. International Committee on Computational Linguistics.
\newblock \doi{10.18653/v1/2020.coling-main.580}.
\newblock URL \url{https://aclanthology.org/2020.coling-main.580}.

\bibitem[Hoffmann et~al.(2022)Hoffmann, Borgeaud, Mensch, Buchatskaya, Cai,
  Rutherford, Casas, Hendricks, Welbl, Clark, et~al.]{hoffmann2022training}
Jordan Hoffmann, Sebastian Borgeaud, Arthur Mensch, Elena Buchatskaya, Trevor
  Cai, Eliza Rutherford, Diego de~Las Casas, Lisa~Anne Hendricks, Johannes
  Welbl, Aidan Clark, et~al.
\newblock Training compute-optimal large language models.
\newblock \emph{arXiv preprint arXiv:2203.15556}, 2022.

\bibitem[Hou et~al.(2023)Hou, Li, Fei, Stolfo, Zhou, Zeng, Bosselut, and
  Sachan]{hou2023towards}
Yifan Hou, Jiaoda Li, Yu~Fei, Alessandro Stolfo, Wangchunshu Zhou, Guangtao
  Zeng, Antoine Bosselut, and Mrinmaya Sachan.
\newblock Towards a mechanistic interpretation of multi-step reasoning
  capabilities of language models.
\newblock \emph{arXiv preprint arXiv:2310.14491}, 2023.

\bibitem[Huang and Chang(2022)]{huang2022towards}
Jie Huang and Kevin Chen-Chuan Chang.
\newblock Towards reasoning in large language models: A survey.
\newblock \emph{arXiv preprint arXiv:2212.10403}, 2022.

\bibitem[Huang et~al.(2023)Huang, Shen, Zhang, Zhou, Rong, and
  Xiong]{huang-TransformerPatcher-2023}
Zeyu Huang, Yikang Shen, Xiaofeng Zhang, Jie Zhou, Wenge Rong, and Zhang Xiong.
\newblock Transformer-patcher: One mistake worth one neuron.
\newblock In \emph{The Eleventh International Conference on Learning
  Representations}, 2023.
\newblock URL \url{https://openreview.net/forum?id=4oYUGeGBPm}.

\bibitem[Jablonka et~al.(2023)Jablonka, Ai, Al-Feghali, Badhwar, Bocarsly,
  Bran, Bringuier, Brinson, Choudhary, Circi, et~al.]{jablonka202314}
Kevin~Maik Jablonka, Qianxiang Ai, Alexander Al-Feghali, Shruti Badhwar,
  Joshua~D Bocarsly, Andres~M Bran, Stefan Bringuier, L~Catherine Brinson,
  Kamal Choudhary, Defne Circi, et~al.
\newblock 14 examples of how llms can transform materials science and
  chemistry: a reflection on a large language model hackathon.
\newblock \emph{Digital Discovery}, 2\penalty0 (5):\penalty0 1233--1250, 2023.

\bibitem[Jang et~al.(2022)Jang, Ye, Yang, Shin, Han, KIM, Choi, and
  Seo]{jang-ContinualKnowledgeLearningLLMs-2022}
Joel Jang, Seonghyeon Ye, Sohee Yang, Joongbo Shin, Janghoon Han, Gyeonghun
  KIM, Stanley~Jungkyu Choi, and Minjoon Seo.
\newblock Towards continual knowledge learning of language models.
\newblock In \emph{International Conference on Learning Representations}, 2022.
\newblock URL \url{https://openreview.net/forum?id=vfsRB5MImo9}.

\bibitem[Jiang et~al.(2023{\natexlab{a}})Jiang, Zhou, Zhao, and
  Wen]{jiang_unikgqa_2023}
Jinhao Jiang, Kun Zhou, Xin Zhao, and Ji-Rong Wen.
\newblock Uni{KGQA}: Unified retrieval and reasoning for solving multi-hop
  question answering over knowledge graph.
\newblock In \emph{The Eleventh International Conference on Learning
  Representations}, 2023{\natexlab{a}}.
\newblock URL \url{https://openreview.net/forum?id=Z63RvyAZ2Vh}.

\bibitem[Jiang et~al.(2020)Jiang, Xu, Araki, and Neubig]{jiang2020can}
Zhengbao Jiang, Frank~F Xu, Jun Araki, and Graham Neubig.
\newblock How can we know what language models know?
\newblock \emph{Transactions of the Association for Computational Linguistics},
  8:\penalty0 423--438, 2020.

\bibitem[Jiang et~al.(2023{\natexlab{b}})Jiang, Xu, Gao, Sun, Liu, Dwivedi-Yu,
  Yang, Callan, and Neubig]{jiang-ActiveRetrievalAugmentedGeneration-2023}
Zhengbao Jiang, Frank~F Xu, Luyu Gao, Zhiqing Sun, Qian Liu, Jane Dwivedi-Yu,
  Yiming Yang, Jamie Callan, and Graham Neubig.
\newblock Active retrieval augmented generation.
\newblock \emph{arXiv preprint arXiv:2305.06983}, 2023{\natexlab{b}}.
\newblock \doi{10.48550/arXiv.2305.06983}.

\bibitem[Kandpal et~al.(2022)Kandpal, Wallace, and
  Raffel]{kandpal2022deduplicating}
Nikhil Kandpal, Eric Wallace, and Colin Raffel.
\newblock Deduplicating training data mitigates privacy risks in language
  models, 2022.

\bibitem[Kandpal et~al.(2023{\natexlab{a}})Kandpal, Deng, Roberts, Wallace, and
  Raffel]{kandpal2023large}
Nikhil Kandpal, Haikang Deng, Adam Roberts, Eric Wallace, and Colin Raffel.
\newblock Large language models struggle to learn long-tail knowledge,
  2023{\natexlab{a}}.

\bibitem[Kandpal et~al.(2023{\natexlab{b}})Kandpal, Jagielski, Tram{\`e}r, and
  Carlini]{kandpal2023backdoor}
Nikhil Kandpal, Matthew Jagielski, Florian Tram{\`e}r, and Nicholas Carlini.
\newblock Backdoor attacks for in-context learning with language models.
\newblock \emph{arXiv preprint arXiv:2307.14692}, 2023{\natexlab{b}}.

\bibitem[Kaplan et~al.(2020)Kaplan, McCandlish, Henighan, Brown, Chess, Child,
  Gray, Radford, Wu, and Amodei]{kaplan2020scaling}
Jared Kaplan, Sam McCandlish, Tom Henighan, Tom~B Brown, Benjamin Chess, Rewon
  Child, Scott Gray, Alec Radford, Jeffrey Wu, and Dario Amodei.
\newblock Scaling laws for neural language models.
\newblock \emph{arXiv preprint arXiv:2001.08361}, 2020.

\bibitem[Katz et~al.(2024)Katz, Belinkov, Geva, and Wolf]{katz2024backward}
Shahar Katz, Yonatan Belinkov, Mor Geva, and Lior Wolf.
\newblock Backward lens: Projecting language model gradients into the
  vocabulary space.
\newblock \emph{arXiv preprint arXiv:2402.12865}, 2024.

\bibitem[Kemker et~al.(2018)Kemker, McClure, Abitino, Hayes, and
  Kanan]{kemker2018measuring}
Ronald Kemker, Marc McClure, Angelina Abitino, Tyler Hayes, and Christopher
  Kanan.
\newblock Measuring catastrophic forgetting in neural networks.
\newblock In \emph{Proceedings of the AAAI conference on artificial
  intelligence}, volume~32, 2018.

\bibitem[Kim et~al.(2019)Kim, Patel, Poliak, Wang, Xia, McCoy, Tenney, Ross,
  Linzen, Van~Durme, et~al.]{kim2019probing}
Najoung Kim, Roma Patel, Adam Poliak, Alex Wang, Patrick Xia, R~Thomas McCoy,
  Ian Tenney, Alexis Ross, Tal Linzen, Benjamin Van~Durme, et~al.
\newblock Probing what different nlp tasks teach machines about function word
  comprehension.
\newblock \emph{arXiv preprint arXiv:1904.11544}, 2019.

\bibitem[Kirkpatrick et~al.(2017)Kirkpatrick, Pascanu, Rabinowitz, Veness,
  Desjardins, Rusu, Milan, Quan, Ramalho, Grabska-Barwinska,
  et~al.]{kirkpatrick2017overcoming}
James Kirkpatrick, Razvan Pascanu, Neil Rabinowitz, Joel Veness, Guillaume
  Desjardins, Andrei~A Rusu, Kieran Milan, John Quan, Tiago Ramalho, Agnieszka
  Grabska-Barwinska, et~al.
\newblock Overcoming catastrophic forgetting in neural networks.
\newblock \emph{Proceedings of the national academy of sciences}, 114\penalty0
  (13):\penalty0 3521--3526, 2017.

\bibitem[Komeili et~al.(2021)Komeili, Shuster, and Weston]{komeili2021internet}
Mojtaba Komeili, Kurt Shuster, and Jason Weston.
\newblock Internet-augmented dialogue generation.
\newblock \emph{arXiv preprint arXiv:2107.07566}, 2021.

\bibitem[Kumar et~al.(2022)Kumar, Raghunathan, Jones, Ma, and
  Liang]{kumar2022finetuning}
Ananya Kumar, Aditi Raghunathan, Robbie~Matthew Jones, Tengyu Ma, and Percy
  Liang.
\newblock Fine-tuning can distort pretrained features and underperform
  out-of-distribution.
\newblock In \emph{International Conference on Learning Representations}, 2022.
\newblock URL \url{https://openreview.net/forum?id=UYneFzXSJWh}.

\bibitem[Lee and Sengupta()]{Lee_Sengupta}
Kevin Lee and Shubho Sengupta.
\newblock Introducing the ai research supercluster - meta’s cutting-edge ai
  supercomputer for ai research.
\newblock URL \url{https://ai.meta.com/blog/ai-rsc/}.

\bibitem[Lewis et~al.(2020)Lewis, Perez, Piktus, Petroni, Karpukhin, Goyal,
  K{\"u}ttler, Lewis, Yih, Rockt{\"a}schel, et~al.]{lewis2020retrieval}
Patrick Lewis, Ethan Perez, Aleksandra Piktus, Fabio Petroni, Vladimir
  Karpukhin, Naman Goyal, Heinrich K{\"u}ttler, Mike Lewis, Wen-tau Yih, Tim
  Rockt{\"a}schel, et~al.
\newblock Retrieval-augmented generation for knowledge-intensive nlp tasks.
\newblock \emph{Advances in Neural Information Processing Systems},
  33:\penalty0 9459--9474, 2020.

\bibitem[Li et~al.(2022)Li, Hopkins, Bau, Vi{\'e}gas, Pfister, and
  Wattenberg]{li2022emergent}
Kenneth Li, Aspen~K Hopkins, David Bau, Fernanda Vi{\'e}gas, Hanspeter Pfister,
  and Martin Wattenberg.
\newblock Emergent world representations: Exploring a sequence model trained on
  a synthetic task.
\newblock \emph{arXiv preprint arXiv:2210.13382}, 2022.

\bibitem[Li et~al.(2023)Li, Li, Song, Yang, Ma, and Yu]{li2023pmet}
Xiaopeng Li, Shasha Li, Shezheng Song, Jing Yang, Jun Ma, and Jie Yu.
\newblock {PMET}: Precise model editing in a transformer.
\newblock \emph{arXiv preprint arXiv:2308.08742}, 2023.
\newblock \doi{10.48550/arXiv.2308.08742}.

\bibitem[Ling et~al.(2017)Ling, Yogatama, Dyer, and Blunsom]{ling2017program}
Wang Ling, Dani Yogatama, Chris Dyer, and Phil Blunsom.
\newblock Program induction by rationale generation: Learning to solve and
  explain algebraic word problems.
\newblock \emph{arXiv preprint arXiv:1705.04146}, 2017.

\bibitem[Liu et~al.(2020)Liu, Chen, Xie, Siow, and Liu]{liu2020retrieval}
Shangqing Liu, Yu~Chen, Xiaofei Xie, Jingkai Siow, and Yang Liu.
\newblock Retrieval-augmented generation for code summarization via hybrid gnn.
\newblock \emph{arXiv preprint arXiv:2006.05405}, 2020.

\bibitem[Long(2023)]{tot_long2023large}
Jieyi Long.
\newblock Large language model guided tree-of-thought.
\newblock \emph{arXiv preprint arXiv:2305.08291}, 2023.
\newblock \doi{10.48550/arXiv.2305.08291}.

\bibitem[Maini et~al.(2023)Maini, Mozer, Sedghi, Lipton, Kolter, and
  Zhang]{maini2023can}
Pratyush Maini, Michael~C Mozer, Hanie Sedghi, Zachary~C Lipton, J~Zico Kolter,
  and Chiyuan Zhang.
\newblock Can neural network memorization be localized?
\newblock \emph{arXiv preprint arXiv:2307.09542}, 2023.

\bibitem[Mangrulkar et~al.(2022)Mangrulkar, Gugger, Debut, Belkada, Paul, and
  Bossan]{peft}
Sourab Mangrulkar, Sylvain Gugger, Lysandre Debut, Younes Belkada, Sayak Paul,
  and Benjamin Bossan.
\newblock Peft: State-of-the-art parameter-efficient fine-tuning methods.
\newblock \url{https://github.com/huggingface/peft}, 2022.

\bibitem[McGrath et~al.(2023)McGrath, Rahtz, Kramar, Mikulik, and
  Legg]{mcgrath2023hydra}
Thomas McGrath, Matthew Rahtz, Janos Kramar, Vladimir Mikulik, and Shane Legg.
\newblock The {H}ydra effect: Emergent self-repair in language model
  computations.
\newblock \emph{arXiv preprint arXiv:2307.15771}, 2023.

\bibitem[Meng et~al.(2022{\natexlab{a}})Meng, Bau, Andonian, and
  Belinkov]{ROME}
Kevin Meng, David Bau, Alex Andonian, and Yonatan Belinkov.
\newblock Locating and editing factual associations in {GPT}.
\newblock \emph{Advances in Neural Information Processing Systems},
  35:\penalty0 17359--17372, 2022{\natexlab{a}}.
\newblock URL
  \url{https://proceedings.neurips.cc/paper_files/paper/2022/file/6f1d43d5a82a37e89b0665b33bf3a182-Paper-Conference.pdf}.

\bibitem[Meng et~al.(2022{\natexlab{b}})Meng, Sharma, Andonian, Belinkov, and
  Bau]{meng2022mass}
Kevin Meng, Arnab~Sen Sharma, Alex Andonian, Yonatan Belinkov, and David Bau.
\newblock Mass-editing memory in a transformer.
\newblock \emph{arXiv preprint arXiv:2210.07229}, 2022{\natexlab{b}}.
\newblock \doi{10.48550/arXiv.2210.07229}.

\bibitem[Miao et~al.(2021)Miao, Liang, and Su]{miao2021diverse}
Shen-Yun Miao, Chao-Chun Liang, and Keh-Yih Su.
\newblock A diverse corpus for evaluating and developing english math word
  problem solvers.
\newblock \emph{arXiv preprint arXiv:2106.15772}, 2021.

\bibitem[Mitchell et~al.(2022{\natexlab{a}})Mitchell, Lin, Bosselut, Finn, and
  Manning]{mitchell2022fast}
Eric Mitchell, Charles Lin, Antoine Bosselut, Chelsea Finn, and Christopher~D
  Manning.
\newblock Fast model editing at scale.
\newblock In \emph{International Conference on Learning Representations},
  2022{\natexlab{a}}.
\newblock URL \url{https://openreview.net/forum?id=0DcZxeWfOPt}.

\bibitem[Mitchell et~al.(2022{\natexlab{b}})Mitchell, Lin, Bosselut, Finn, and
  Manning]{mitchell_FastModelEditing_2022}
Eric Mitchell, Charles Lin, Antoine Bosselut, Chelsea Finn, and Christopher~D
  Manning.
\newblock Fast model editing at scale.
\newblock In \emph{International Conference on Learning Representations},
  2022{\natexlab{b}}.
\newblock URL \url{https://openreview.net/forum?id=0DcZxeWfOPt}.

\bibitem[Mitchell et~al.(2022{\natexlab{c}})Mitchell, Lin, Bosselut, Manning,
  and Finn]{mitchell_MemoryBasedModelEditing_2022}
Eric Mitchell, Charles Lin, Antoine Bosselut, Christopher~D Manning, and
  Chelsea Finn.
\newblock Memory-based model editing at scale.
\newblock In Kamalika Chaudhuri, Stefanie Jegelka, Le~Song, Csaba Szepesvari,
  Gang Niu, and Sivan Sabato, editors, \emph{Proceedings of the 39th
  International Conference on Machine Learning}, volume 162 of
  \emph{Proceedings of Machine Learning Research}, pages 15817--15831. PMLR,
  17--23 Jul 2022{\natexlab{c}}.
\newblock URL \url{https://proceedings.mlr.press/v162/mitchell22a.html}.

\bibitem[Nadeem et~al.(2020)Nadeem, Bethke, and Reddy]{nadeem2020stereoset}
Moin Nadeem, Anna Bethke, and Siva Reddy.
\newblock Stereoset: Measuring stereotypical bias in pretrained language
  models.
\newblock \emph{arXiv preprint arXiv:2004.09456}, 2020.

\bibitem[Nanda and Bloom(2022)]{transformer_lens}
Neel Nanda and Joseph Bloom.
\newblock {TransformerLens}, 2022.
\newblock URL \url{https://github.com/neelnanda-io/TransformerLens}.

\bibitem[Nanda et~al.(2023{\natexlab{a}})Nanda, Chan, Lieberum, Smith, and
  Steinhardt]{nanda2023progress}
Neel Nanda, Lawrence Chan, Tom Lieberum, Jess Smith, and Jacob Steinhardt.
\newblock Progress measures for grokking via mechanistic interpretability.
\newblock In \emph{The Eleventh International Conference on Learning
  Representations}, 2023{\natexlab{a}}.
\newblock URL \url{https://openreview.net/forum?id=9XFSbDPmdW}.

\bibitem[Nanda et~al.(2023{\natexlab{b}})Nanda, Chan, Lieberum, Smith, and
  Steinhardt]{nanda_ProgressMeasuresGrokking_2022}
Neel Nanda, Lawrence Chan, Tom Lieberum, Jess Smith, and Jacob Steinhardt.
\newblock Progress measures for grokking via mechanistic interpretability.
\newblock In \emph{The Eleventh International Conference on Learning
  Representations}, 2023{\natexlab{b}}.
\newblock URL \url{https://openreview.net/forum?id=9XFSbDPmdW}.

\bibitem[Nanda et~al.(2023{\natexlab{c}})Nanda, Lee, and
  Wattenberg]{nanda2023emergent}
Neel Nanda, Andrew Lee, and Martin Wattenberg.
\newblock Emergent linear representations in world models of self-supervised
  sequence models.
\newblock \emph{arXiv preprint arXiv:2309.00941}, 2023{\natexlab{c}}.

\bibitem[Nasr et~al.(2023)Nasr, Carlini, Hayase, Jagielski, Cooper, Ippolito,
  Choquette-Choo, Wallace, Tramèr, and Lee]{nasr2023scalable}
Milad Nasr, Nicholas Carlini, Jonathan Hayase, Matthew Jagielski, A.~Feder
  Cooper, Daphne Ippolito, Christopher~A. Choquette-Choo, Eric Wallace, Florian
  Tramèr, and Katherine Lee.
\newblock Scalable extraction of training data from (production) language
  models, 2023.

\bibitem[nostalgebraist(2021)]{logitlens}
nostalgebraist.
\newblock Logit {Lens} on non-{GPT2} models + extensions, 2021.
\newblock URL
  \url{https://colab.research.google.com/drive/1MjdfK2srcerLrAJDRaJQKO0sUiZ-hQtA}.

\bibitem[Olsson et~al.(2022)Olsson, Elhage, Nanda, Joseph, DasSarma, Henighan,
  Mann, Askell, Bai, Chen, et~al.]{olsson2022context}
Catherine Olsson, Nelson Elhage, Neel Nanda, Nicholas Joseph, Nova DasSarma,
  Tom Henighan, Ben Mann, Amanda Askell, Yuntao Bai, Anna Chen, et~al.
\newblock In-context learning and induction heads.
\newblock \emph{arXiv preprint arXiv:2209.11895}, 2022.

\bibitem[OpenAI(2022)]{openai2022chatgpt}
TB~OpenAI.
\newblock Chatgpt: Optimizing language models for dialogue. openai, 2022.

\bibitem[Ovadia et~al.(2023)Ovadia, Brief, Mishaeli, and
  Elisha]{ovadia2023fine}
Oded Ovadia, Menachem Brief, Moshik Mishaeli, and Oren Elisha.
\newblock Fine-tuning or retrieval? comparing knowledge injection in llms.
\newblock \emph{arXiv preprint arXiv:2312.05934}, 2023.

\bibitem[Pal et~al.(2023)Pal, Sun, Yuan, Wallace, and Bau]{pal2023future}
Koyena Pal, Jiuding Sun, Andrew Yuan, Byron~C Wallace, and David Bau.
\newblock Future lens: Anticipating subsequent tokens from a single hidden
  state.
\newblock \emph{arXiv preprint arXiv:2311.04897}, 2023.

\bibitem[Panigrahi et~al.(2023)Panigrahi, Saunshi, Zhao, and
  Arora]{panigrahi2023task}
Abhishek Panigrahi, Nikunj Saunshi, Haoyu Zhao, and Sanjeev Arora.
\newblock Task-specific skill localization in fine-tuned language models.
\newblock \emph{arXiv preprint arXiv:2302.06600}, 2023.

\bibitem[Passmore(1961)]{passmore1961philosophical}
John~Arthur Passmore.
\newblock Philosophical reasoning.
\newblock 1961.

\bibitem[Patel et~al.(2021)Patel, Bhattamishra, and Goyal]{patel2021nlp}
Arkil Patel, Satwik Bhattamishra, and Navin Goyal.
\newblock Are nlp models really able to solve simple math word problems?
\newblock \emph{arXiv preprint arXiv:2103.07191}, 2021.

\bibitem[Patil et~al.(2023)Patil, Hase, and Bansal]{patil2023sensitive}
Vaidehi Patil, Peter Hase, and Mohit Bansal.
\newblock Can sensitive information be deleted from llms? objectives for
  defending against extraction attacks, 2023.

\bibitem[Perez and Ribeiro(2022)]{perez2022ignore}
F{\'a}bio Perez and Ian Ribeiro.
\newblock Ignore previous prompt: Attack techniques for language models.
\newblock \emph{arXiv preprint arXiv:2211.09527}, 2022.

\bibitem[Petroni et~al.(2019)Petroni, Rockt{\"a}schel, Lewis, Bakhtin, Wu,
  Miller, and Riedel]{petroni2019language}
Fabio Petroni, Tim Rockt{\"a}schel, Patrick Lewis, Anton Bakhtin, Yuxiang Wu,
  Alexander~H Miller, and Sebastian Riedel.
\newblock Language models as knowledge bases?
\newblock \emph{arXiv preprint arXiv:1909.01066}, 2019.

\bibitem[Qi et~al.(2023)Qi, Zeng, Xie, Chen, Jia, Mittal, and
  Henderson]{qi2023finetuning}
Xiangyu Qi, Yi~Zeng, Tinghao Xie, Pin-Yu Chen, Ruoxi Jia, Prateek Mittal, and
  Peter Henderson.
\newblock Fine-tuning aligned language models compromises safety, even when
  users do not intend to!, 2023.

\bibitem[Radford et~al.(2019{\natexlab{a}})Radford, Wu, Child, Luan, Amodei,
  and Sutskever]{gpt2}
Alec Radford, Jeff Wu, Rewon Child, David Luan, Dario Amodei, and Ilya
  Sutskever.
\newblock Language models are unsupervised multitask learners.
\newblock 2019{\natexlab{a}}.
\newblock
  \url{https://paperswithcode.com/paper/language-models-are-unsupervised-multitask}.

\bibitem[Radford et~al.(2019{\natexlab{b}})Radford, Wu, Child, Luan, Amodei,
  Sutskever, et~al.]{radford_llmsOpenAI_2019}
Alec Radford, Jeffrey Wu, Rewon Child, David Luan, Dario Amodei, Ilya
  Sutskever, et~al.
\newblock Language models are unsupervised multitask learners.
\newblock \emph{OpenAI blog}, 1\penalty0 (8):\penalty0 9, 2019{\natexlab{b}}.
\newblock URL
  \url{https://d4mucfpksywv.cloudfront.net/better-language-models/language-models.pdf}.

\bibitem[Raji et~al.(2021)Raji, Bender, Paullada, Denton, and
  Hanna]{raji2021ai}
Inioluwa~Deborah Raji, Emily~M Bender, Amandalynne Paullada, Emily Denton, and
  Alex Hanna.
\newblock Ai and the everything in the whole wide world benchmark.
\newblock \emph{arXiv preprint arXiv:2111.15366}, 2021.

\bibitem[Roberts et~al.(2020)Roberts, Raffel, and Shazeer]{roberts2020much}
Adam Roberts, Colin Raffel, and Noam Shazeer.
\newblock How much knowledge can you pack into the parameters of a language
  model?
\newblock \emph{arXiv preprint arXiv:2002.08910}, 2020.

\bibitem[Roy and Roth(2016)]{roy2016solving}
Subhro Roy and Dan Roth.
\newblock Solving general arithmetic word problems.
\newblock \emph{arXiv preprint arXiv:1608.01413}, 2016.

\bibitem[Ruder(2016)]{ruder2016overview}
Sebastian Ruder.
\newblock An overview of gradient descent optimization algorithms.
\newblock \emph{arXiv preprint arXiv:1609.04747}, 2016.

\bibitem[Sakarvadia et~al.(2023)Sakarvadia, Ajith, Khan, Grzenda, Hudson,
  Bauer, Chard, and Foster]{sakarvadia2023memory}
Mansi Sakarvadia, Aswathy Ajith, Arham Khan, Daniel Grzenda, Nathaniel Hudson,
  Andr{\'e} Bauer, Kyle Chard, and Ian Foster.
\newblock Memory injections: Correcting multi-hop reasoning failures during
  inference in transformer-based language models.
\newblock \emph{arXiv preprint arXiv:2309.05605}, 2023.

\bibitem[Sel et~al.(2023)Sel, Al-Tawaha, Khattar, Wang, Jia, and
  Jin]{sel2023algorithm}
Bilgehan Sel, Ahmad Al-Tawaha, Vanshaj Khattar, Lu~Wang, Ruoxi Jia, and Ming
  Jin.
\newblock Algorithm of thoughts: Enhancing exploration of ideas in large
  language models.
\newblock \emph{arXiv preprint arXiv:2308.10379}, 2023.

\bibitem[Shi et~al.(2024)Shi, Ajith, Xia, Huang, Liu, Blevins, Chen, and
  Zettlemoyer]{shi2024detecting}
Weijia Shi, Anirudh Ajith, Mengzhou Xia, Yangsibo Huang, Daogao Liu, Terra
  Blevins, Danqi Chen, and Luke Zettlemoyer.
\newblock Detecting pretraining data from large language models, 2024.

\bibitem[Srivastava et~al.(2022)Srivastava, Rastogi, Rao, Shoeb, Abid, Fisch,
  Brown, Santoro, Gupta, Garriga-Alonso, et~al.]{srivastava2022beyond}
Aarohi Srivastava, Abhinav Rastogi, Abhishek Rao, Abu Awal~Md Shoeb, Abubakar
  Abid, Adam Fisch, Adam~R Brown, Adam Santoro, Aditya Gupta, Adri{\`a}
  Garriga-Alonso, et~al.
\newblock Beyond the imitation game: Quantifying and extrapolating the
  capabilities of language models.
\newblock \emph{arXiv preprint arXiv:2206.04615}, 2022.

\bibitem[Sun et~al.(2018)Sun, Dhingra, Zaheer, Mazaitis, Salakhutdinov, and
  Cohen]{sun_open_2018}
Haitian Sun, Bhuwan Dhingra, Manzil Zaheer, Kathryn Mazaitis, Ruslan
  Salakhutdinov, and William Cohen.
\newblock Open domain question answering using early fusion of knowledge bases
  and text.
\newblock In \emph{Proceedings of the 2018 Conference on Empirical Methods in
  Natural Language Processing}, pages 4231--4242, Brussels, Belgium,
  October-November 2018. Association for Computational Linguistics.
\newblock \doi{10.18653/v1/D18-1455}.
\newblock URL \url{https://aclanthology.org/D18-1455}.

\bibitem[Sun et~al.(2021)Sun, Guo, and Li]{sun2021react}
Yiyou Sun, Chuan Guo, and Yixuan Li.
\newblock React: Out-of-distribution detection with rectified activations.
\newblock \emph{Advances in Neural Information Processing Systems},
  34:\penalty0 144--157, 2021.

\bibitem[Talmor et~al.(2018)Talmor, Herzig, Lourie, and
  Berant]{talmor2018commonsenseqa}
Alon Talmor, Jonathan Herzig, Nicholas Lourie, and Jonathan Berant.
\newblock Commonsenseqa: A question answering challenge targeting commonsense
  knowledge.
\newblock \emph{arXiv preprint arXiv:1811.00937}, 2018.

\bibitem[Touvron et~al.(2023)Touvron, Martin, Stone, Albert, Almahairi, Babaei,
  Bashlykov, Batra, Bhargava, Bhosale, et~al.]{touvron2023llama}
Hugo Touvron, Louis Martin, Kevin Stone, Peter Albert, Amjad Almahairi, Yasmine
  Babaei, Nikolay Bashlykov, Soumya Batra, Prajjwal Bhargava, Shruti Bhosale,
  et~al.
\newblock Llama 2: Open foundation and fine-tuned chat models.
\newblock \emph{arXiv preprint arXiv:2307.09288}, 2023.

\bibitem[Turner et~al.(2023)Turner, Thiergart, Udell, Leech, Mini, and
  MacDiarmid]{turner2023activation}
Alex Turner, Lisa Thiergart, David Udell, Gavin Leech, Ulisse Mini, and Monte
  MacDiarmid.
\newblock Activation addition: Steering language models without optimization.
\newblock \emph{arXiv preprint arXiv:2308.10248}, 2023.

\bibitem[Turpin et~al.(2023)Turpin, Michael, Perez, and
  Bowman]{turpin2023language}
Miles Turpin, Julian Michael, Ethan Perez, and Samuel~R Bowman.
\newblock Language models don't always say what they think: Unfaithful
  explanations in chain-of-thought prompting.
\newblock \emph{arXiv preprint arXiv:2305.04388}, 2023.
\newblock \doi{10.48550/arXiv.2305.04388}.

\bibitem[Vaswani et~al.(2017)Vaswani, Shazeer, Parmar, Uszkoreit, Jones, Gomez,
  Kaiser, and Polosukhin]{vaswani2017attention}
Ashish Vaswani, Noam Shazeer, Niki Parmar, Jakob Uszkoreit, Llion Jones,
  Aidan~N Gomez, {\L}ukasz Kaiser, and Illia Polosukhin.
\newblock Attention is all you need.
\newblock \emph{Advances in Neural Information Processing Systems}, 30, 2017.
\newblock URL
  \url{https://proceedings.neurips.cc/paper/2017/file/3f5ee243547dee91fbd053c1c4a845aa-Paper.pdf}.

\bibitem[Vig et~al.(2020)Vig, Gehrmann, Belinkov, Qian, Nevo, Singer, and
  Shieber]{vig2020investigating}
Jesse Vig, Sebastian Gehrmann, Yonatan Belinkov, Sharon Qian, Daniel Nevo,
  Yaron Singer, and Stuart Shieber.
\newblock Investigating gender bias in language models using causal mediation
  analysis.
\newblock \emph{Advances in neural information processing systems},
  33:\penalty0 12388--12401, 2020.

\bibitem[Wang et~al.(2022{\natexlab{a}})Wang, Deng, and
  Sun]{wang-etal-2022-iteratively}
Boshi Wang, Xiang Deng, and Huan Sun.
\newblock Iteratively prompt pre-trained language models for chain of thought.
\newblock In \emph{Proceedings of the 2022 Conference on Empirical Methods in
  Natural Language Processing}, pages 2714--2730, Abu Dhabi, United Arab
  Emirates, December 2022{\natexlab{a}}. Association for Computational
  Linguistics.
\newblock \doi{10.18653/v1/2022.emnlp-main.174}.
\newblock URL \url{https://aclanthology.org/2022.emnlp-main.174}.

\bibitem[Wang et~al.(2023{\natexlab{a}})Wang, Chen, Pei, Xie, Kang, Zhang, Xu,
  Xiong, Dutta, Schaeffer, et~al.]{wang2023decodingtrust}
Boxin Wang, Weixin Chen, Hengzhi Pei, Chulin Xie, Mintong Kang, Chenhui Zhang,
  Chejian Xu, Zidi Xiong, Ritik Dutta, Rylan Schaeffer, et~al.
\newblock Decodingtrust: A comprehensive assessment of trustworthiness in gpt
  models.
\newblock 2023{\natexlab{a}}.

\bibitem[Wang et~al.(2022{\natexlab{b}})Wang, Variengien, Conmy, Shlegeris, and
  Steinhardt]{wang2022interpretability}
Kevin Wang, Alexandre Variengien, Arthur Conmy, Buck Shlegeris, and Jacob
  Steinhardt.
\newblock Interpretability in the wild: A circuit for indirect object
  identification in {GPT}-2 {S}mall.
\newblock \emph{arXiv preprint arXiv:2211.00593}, 2022{\natexlab{b}}.

\bibitem[Wang et~al.(2023{\natexlab{b}})Wang, Zhu, Liu, Zheng, Chen,
  et~al.]{wang2023knowledge}
Song Wang, Yaochen Zhu, Haochen Liu, Zaiyi Zheng, Chen Chen, et~al.
\newblock Knowledge editing for large language models: A survey.
\newblock \emph{arXiv preprint arXiv:2310.16218}, 2023{\natexlab{b}}.

\bibitem[Wang et~al.(2024)Wang, Shi, Tu, Yuan, Huang, Jiao, and
  Lyu]{wang2024earth}
Wenxuan Wang, Juluan Shi, Zhaopeng Tu, Youliang Yuan, Jen-tse Huang, Wenxiang
  Jiao, and Michael~R Lyu.
\newblock The earth is flat? unveiling factual errors in large language models.
\newblock \emph{arXiv preprint arXiv:2401.00761}, 2024.

\bibitem[Wang et~al.(2023{\natexlab{c}})Wang, Wei, Schuurmans, Le, Chi, Narang,
  Chowdhery, and Zhou]{wang2023selfconsistency}
Xuezhi Wang, Jason Wei, Dale Schuurmans, Quoc~V Le, Ed~H. Chi, Sharan Narang,
  Aakanksha Chowdhery, and Denny Zhou.
\newblock Self-consistency improves chain of thought reasoning in language
  models.
\newblock In \emph{The Eleventh International Conference on Learning
  Representations}, 2023{\natexlab{c}}.
\newblock URL \url{https://openreview.net/forum?id=1PL1NIMMrw}.

\bibitem[Wardat et~al.(2021)Wardat, Le, and Rajan]{wardat2021deeplocalize}
Mohammad Wardat, Wei Le, and Hridesh Rajan.
\newblock Deeplocalize: Fault localization for deep neural networks.
\newblock In \emph{2021 IEEE/ACM 43rd International Conference on Software
  Engineering (ICSE)}, pages 251--262. IEEE, 2021.

\bibitem[Wason and Johnson-Laird(1972)]{wason1972psychology}
Peter~Cathcart Wason and Philip~Nicholas Johnson-Laird.
\newblock \emph{Psychology of reasoning: Structure and content}, volume~86.
\newblock Harvard University Press, 1972.

\bibitem[Wei et~al.(2022{\natexlab{a}})Wei, Tay, Bommasani, Raffel, Zoph,
  Borgeaud, Yogatama, Bosma, Zhou, Metzler, et~al.]{wei2022emergent}
Jason Wei, Yi~Tay, Rishi Bommasani, Colin Raffel, Barret Zoph, Sebastian
  Borgeaud, Dani Yogatama, Maarten Bosma, Denny Zhou, Donald Metzler, et~al.
\newblock Emergent abilities of large language models.
\newblock \emph{arXiv preprint arXiv:2206.07682}, 2022{\natexlab{a}}.

\bibitem[Wei et~al.(2022{\natexlab{b}})Wei, Wang, Schuurmans, Bosma, Xia, Chi,
  Le, Zhou, et~al.]{wei_chain-thought_2023}
Jason Wei, Xuezhi Wang, Dale Schuurmans, Maarten Bosma, Fei Xia, Ed~Chi, Quoc~V
  Le, Denny Zhou, et~al.
\newblock Chain-of-thought prompting elicits reasoning in large language
  models.
\newblock \emph{Advances in Neural Information Processing Systems},
  35:\penalty0 24824--24837, 2022{\natexlab{b}}.
\newblock URL
  \url{https://proceedings.neurips.cc/paper_files/paper/2022/file/9d5609613524ecf4f15af0f7b31abca4-Paper-Conference.pdf}.

\bibitem[White et~al.(2021)White, Pimentel, Saphra, and
  Cotterell]{white2021non}
Jennifer~C White, Tiago Pimentel, Naomi Saphra, and Ryan Cotterell.
\newblock A non-linear structural probe.
\newblock \emph{arXiv preprint arXiv:2105.10185}, 2021.

\bibitem[Winograd(2023)]{winograd2023loose}
Amy Winograd.
\newblock Loose-lipped large language models spill your secrets: The privacy
  implications of large language models.
\newblock \emph{Harvard Journal of Law \& Technology}, 36\penalty0 (2), 2023.

\bibitem[Wu et~al.(2023)Wu, Li, Xu, Dong, Wu, Bian, and Xiong]{wu2023depn}
Xinwei Wu, Junzhuo Li, Minghui Xu, Weilong Dong, Shuangzhi Wu, Chao Bian, and
  Deyi Xiong.
\newblock Depn: Detecting and editing privacy neurons in pretrained language
  models.
\newblock \emph{arXiv preprint arXiv:2310.20138}, 2023.

\bibitem[Wu et~al.(2020)Wu, Dobriban, and Davidson]{wu2020deltagrad}
Yinjun Wu, Edgar Dobriban, and Susan Davidson.
\newblock Deltagrad: Rapid retraining of machine learning models.
\newblock In \emph{International Conference on Machine Learning}, pages
  10355--10366. PMLR, 2020.

\bibitem[Xie et~al.(2023{\natexlab{a}})Xie, Wa, Huang, Zhou, Liu, Linghu, Wang,
  Kit, Grazian, and Hoex]{xie2023large}
Tong Xie, Yuwei Wa, Wei Huang, Yufei Zhou, Yixuan Liu, Qingyuan Linghu,
  Shaozhou Wang, Chunyu Kit, Clara Grazian, and Bram Hoex.
\newblock Large language models as master key: Unlocking the secrets of
  materials science with {GPT}.
\newblock \emph{arXiv preprint arXiv:2304.02213}, 2023{\natexlab{a}}.
\newblock \doi{10.48550/arXiv.2304.02213}.

\bibitem[Xie et~al.(2023{\natexlab{b}})Xie, Kawaguchi, Zhao, Zhao, Kan, He, and
  Xie]{tot_xie2023decomposition}
Yuxi Xie, Kenji Kawaguchi, Yiran Zhao, Xu~Zhao, Min-Yen Kan, Junxian He, and
  Qizhe Xie.
\newblock Decomposition enhances reasoning via self-evaluation guided decoding.
\newblock \emph{arXiv preprint arXiv:2305.00633}, 2023{\natexlab{b}}.
\newblock \doi{10.48550/arXiv.2305.00633}.

\bibitem[Yan et~al.(2024)Yan, Wang, Li, and Zhang]{yan2024potential}
Jianhao Yan, Futing Wang, Yafu Li, and Yue Zhang.
\newblock Potential and challenges of model editing for social debiasing.
\newblock \emph{arXiv preprint arXiv:2402.13462}, 2024.

\bibitem[Yang et~al.(2018)Yang, Qi, Zhang, Bengio, Cohen, Salakhutdinov, and
  Manning]{yang2018hotpotqa}
Zhilin Yang, Peng Qi, Saizheng Zhang, Yoshua Bengio, William~W Cohen, Ruslan
  Salakhutdinov, and Christopher~D Manning.
\newblock Hotpotqa: A dataset for diverse, explainable multi-hop question
  answering.
\newblock \emph{arXiv preprint arXiv:1809.09600}, 2018.

\bibitem[Yao et~al.(2023)Yao, Yu, Zhao, Shafran, Griffiths, Cao, and
  Narasimhan]{yao2023tree}
Shunyu Yao, Dian Yu, Jeffrey Zhao, Izhak Shafran, Thomas~L Griffiths, Yuan Cao,
  and Karthik Narasimhan.
\newblock Tree of thoughts: Deliberate problem solving with large language
  models.
\newblock \emph{arXiv preprint arXiv:2305.10601}, 2023.
\newblock \doi{10.48550/arXiv.2305.10601}.

\bibitem[Yun et~al.(2023)Yun, Sohn, and Kyeong]{yun2023fine}
Jiseon Yun, Jae~Eui Sohn, and Sunghyon Kyeong.
\newblock Fine-tuning pretrained language models to enhance dialogue
  summarization in customer service centers.
\newblock In \emph{Proceedings of the Fourth ACM International Conference on AI
  in Finance}, pages 365--373, 2023.

\bibitem[Zhang et~al.(2022)Zhang, Zhang, Yu, Tang, Tang, Li, and
  Chen]{zhang_subgraph_2022}
Jing Zhang, Xiaokang Zhang, Jifan Yu, Jian Tang, Jie Tang, Cuiping Li, and Hong
  Chen.
\newblock Subgraph retrieval enhanced model for multi-hop knowledge base
  question answering.
\newblock In \emph{Proceedings of the 60th Annual Meeting of the Association
  for Computational Linguistics (Volume 1: Long Papers)}, pages 5773--5784,
  Dublin, Ireland, May 2022. Association for Computational Linguistics.
\newblock \doi{10.18653/v1/2022.acl-long.396}.
\newblock URL \url{https://aclanthology.org/2022.acl-long.396}.

\bibitem[Zhang et~al.(2024)Zhang, Yao, Tian, Wang, Deng, Wang, Xi, Mao, Zhang,
  Ni, et~al.]{zhang2024comprehensive}
Ningyu Zhang, Yunzhi Yao, Bozhong Tian, Peng Wang, Shumin Deng, Mengru Wang,
  Zekun Xi, Shengyu Mao, Jintian Zhang, Yuansheng Ni, et~al.
\newblock A comprehensive study of knowledge editing for large language models.
\newblock \emph{arXiv preprint arXiv:2401.01286}, 2024.

\bibitem[Zhang et~al.(2023)Zhang, Li, Cui, Cai, Liu, Fu, Huang, Zhao, Zhang,
  Chen, et~al.]{zhang2023siren}
Yue Zhang, Yafu Li, Leyang Cui, Deng Cai, Lemao Liu, Tingchen Fu, Xinting
  Huang, Enbo Zhao, Yu~Zhang, Yulong Chen, et~al.
\newblock Siren's song in the ai ocean: A survey on hallucination in large
  language models.
\newblock \emph{arXiv preprint arXiv:2309.01219}, 2023.

\bibitem[Zhong et~al.(2021)Zhong, Friedman, and Chen]{zhong2021factual}
Zexuan Zhong, Dan Friedman, and Danqi Chen.
\newblock Factual probing is [mask]: Learning vs. learning to recall.
\newblock \emph{arXiv preprint arXiv:2104.05240}, 2021.

\bibitem[Zhong et~al.(2023)Zhong, Wu, Manning, Potts, and Chen]{mquake}
Zexuan Zhong, Zhengxuan Wu, Christopher~D Manning, Christopher Potts, and Danqi
  Chen.
\newblock {MQuAKE}: Assessing knowledge editing in language models via
  multi-hop questions.
\newblock \emph{arXiv preprint arXiv:2305.14795}, 2023.
\newblock \doi{https://doi.org/10.48550/arXiv.2305.14795}.

\bibitem[Zhu et~al.(2020)Zhu, Rawat, Zaheer, Bhojanapalli, Li, Yu, and
  Kumar]{zhu2020modifying}
Chen Zhu, Ankit~Singh Rawat, Manzil Zaheer, Srinadh Bhojanapalli, Daliang Li,
  Felix Yu, and Sanjiv Kumar.
\newblock Modifying memories in transformer models, 2020.

\bibitem[Zhu et~al.(2015)Zhu, Kiros, Zemel, Salakhutdinov, Urtasun, Torralba,
  and Fidler]{zhu2015aligning}
Yukun Zhu, Ryan Kiros, Rich Zemel, Ruslan Salakhutdinov, Raquel Urtasun,
  Antonio Torralba, and Sanja Fidler.
\newblock Aligning books and movies: Towards story-like visual explanations by
  watching movies and reading books.
\newblock In \emph{IEEE International Conference on Computer Vision}, pages
  19--27, 2015.

\end{thebibliography}
